\documentclass[twoside,11pt]{article}

\usepackage{jmlr2earxiv,amsmath,amssymb,booktabs, algorithm,natbib}
\usepackage[noend]{algpseudocode}
\usepackage[english]{babel}
\usepackage[utf8]{inputenc}
\usepackage{xcolor}

\def\E{\mathbb E}
\def\P{\mathbb P}

\def\BB{\mathcal B}
\def\CC{\mathcal C}
\def\EE{\mathcal E}
\def\GG{\mathcal G}

\def\MM{\mathcal M}
\def\NN{\mathcal N}

\def\RR{\mathcal R}
\def\SS{\mathcal S}

\usepackage{lastpage}

\ShortHeadings{ODEs for Dynamic Matching in Heterogeneous Networks}{Dai and He}

\firstpageno{1}

\begin{document}

\title{An ODE Model for Dynamic Matching in Heterogeneous Networks}

 \author{\name Xiaowu Dai \email  \\
       \emph{University of California, Los Angeles}
       \AND
       \name Hengzhi He \email  \\
       \emph{Peking University}}

 \maketitle
 
\begin{footnotetext}[1]
{\textit{Address for correspondence:} Xiaowu Dai, Department of Statistics, University of California, Los Angeles, 8125 Math Sciences Bldg \#951554,  Los Angeles, CA 90095, USA.}
\end{footnotetext}

\begin{abstract}
We study the problem of dynamic matching in heterogeneous networks, where agents are subject to compatibility restrictions and stochastic arrival and departure times. In particular, we consider networks with one type of easy-to-match agents and multiple types of hard-to-match agents, each subject to its own compatibility constraints. Such a setting arises in many real-world applications, including kidney exchange programs and carpooling platforms. We introduce a novel approach to modeling dynamic matching by establishing the ordinary differential equation (ODE) model, which offers a new perspective for evaluating various matching algorithms. We study two algorithms, namely the Greedy and Patient Algorithms, where both algorithms prioritize matching compatible hard-to-match agents over easy-to-match agents in heterogeneous networks. Our results demonstrate the trade-off between the conflicting goals of matching agents quickly and optimally, offering insights into the de- sign of real-world dynamic matching systems. We provide simulations and a real-world case study using data from the Organ Procurement and Transplantation Network to validate theoretical predictions.
\end{abstract}


\section{Introduction} 
Dynamic matching is a fundamental statistical problem that is both complex and crucial, with wide-ranging implications for numerous real-world applications. Examples of such applications include paired kidney exchange and ride-sharing in carpooling platforms \citep{Roth2004,roth2007efficient, feng2021we}. 
In dynamic matching problems, agents arrive and depart at random times, and the planner aims to maximize the number of matches by considering multiple factors such as agent compatibility and arrival and departure times. 
Dynamic matching has attracted the attention of researchers in economics, operations research, and computer science and remains an active area of research due to its significant implications for the design and implementation of real-world matching systems \citep[e.g.,][]{roth2005pairwise, unver2010dynamic, akbarpour2020thickness, aouad2020dynamic}.

In this paper, we focus on dynamic matching in heterogeneous networks, where agents are subject to various compatibility constraints and their arrival and departure times are uncertain. These compatibility restrictions can arise from a multitude of sources, such as blood-type compatibility in organ transplantation markets, where a successful match is contingent upon biological criteria such as blood type and genotype \citep{roth2005pairwise,unver2010dynamic}.
To address this challenge, we study a stochastic compatibility model that considers multiple types of agents, including both easy-to-match and hard-to-match agents. The model represents the matching market as a random graph, where nodes denote agents and edges denote the compatibility between them.
However, evaluating dynamic matching algorithms in discrete-time models presents a number of challenges. First, the dynamic and non-stationary arrival and departure of agents complicate the accurate evaluation of social welfare under different algorithms. Second, the incomplete information regarding future agent arrivals and departures creates uncertainty in predicting market states and evaluating algorithm performance. Furthermore, the interdependence between matching and loss functions, resulting from the impact of one agent's matching on others, further adds to the complexity of the analysis.

We propose a novel approach to modeling dynamic matching in heterogeneous networks using ordinary differential equation (ODE) models by taking small step sizes in time. Our approach enables the evaluation and comparison of different matching algorithms, while providing fresh insights into the interplay between sequential decision-making and first-order ODE outcomes. By leveraging this framework, we analyze two algorithms, namely the \emph{Greedy Algorithm} and the \emph{Patient Algorithm}. The Greedy Algorithm matches agents as soon as they arrive in the market, while the Patient Algorithm matches agents when they become critical. In heterogeneous networks, both algorithms prioritize the matching of compatible, hard-to-match agents over easy-to-match agents, thus contributing to the maximization of social welfare by directing limited matching opportunities towards agents with greater difficulty finding compatible partners.

Our results show the trade-off between agents' waiting times and the percentage of matched agents in heterogeneous dynamic markets.
The Greedy Algorithm prioritizes fast matching and reduces waiting times. However, the Greedy Algorithm may also lead to a reduced pool of compatible agents. Our findings reveal that the rate of unmatched agents departing the market in the Greedy Algorithm decreases \emph{linearly} with increasing network density. In contrast, the Patient Algorithm, with its focus on increasing the pool of compatible agents, exhibits no slower than an \emph{exponential} decay rate with increasing network density. Although the Patient Algorithm increases the number of matching, we show that it also significantly extends waiting times in comparison to the Greedy Algorithm. These insights into the trade-off between agents' waiting times and the percentage of matched agents have useful implications for the design of dynamic matching systems in real-world scenarios.

We provide extensive simulations to verify the accuracy of our ODE approximation, perform a thorough sensitivity analysis of key parameters in our dynamic matching models, and validate our theoretical predictions through a real-world case study using data from the Organ Procurement and Transplantation Network (OPTN). Our real data analysis indicates that the proportion of easy-to-match agents in the market has little impact on the loss for easy-to-match agents, but causes a significant change in the loss for hard-to-match agents. Additionally, we observe that the Patient Algorithm yields a smaller loss compared to the Greedy Algorithm. These findings demonstrate the practical applicability of our theory and contribute to a deeper understanding of the kidney exchange market.

\subsection{Contributions and Outline}
In this work, we present a novel approach for modeling dynamic matching markets by establishing an asymptotic equivalence between the discrete-time process of dynamic matching and continuous ODEs.
Our work demonstrates that ODEs provide a conceptually simpler tool for understanding and analyzing various  dynamic matching algorithms. 
We summarize our principal methodological and theoretical contributions as follows.

\begin{itemize}
    \item We employ ODEs to model the behavior of dynamic matching algorithms at the exact limit, by taking infinitesimally small time steps.  To the best of our knowledge, this work is the first to use ODEs for this purpose, offering new perspectives on the relationship between sequential decision-making and first-order ODE outcomes. Our models accommodate heterogeneous networks with multiple types of agents and varying compatibility, providing a flexible tool for analyzing and understanding dynamic matching algorithms.
    \item We study two algorithms for dynamic matching in heterogeneous networks, which address the trade-off between agents' waiting times and the percentage of matched agents in heterogeneous dynamic markets. The first algorithm, the \emph{Greedy Algorithm}, matches agents immediately upon their arrival to the market, giving priority to compatible hard-to-match agents over compatible easy-to-match agents. The second algorithm, the \emph{Patient Algorithm},  matches agents only when they become critical and prioritizes matching with compatible hard-to-match agents over easy-to-match agents. These algorithms build upon prior studies of dynamic matching in homogeneous networks \citep{akbarpour2020thickness} by expanding the scope to heterogeneous networks, presenting novel challenges in terms of analysis and comparison.
    \item We demonstrate the existence of ordinary differential equation (ODE) solutions for the Greedy and Patient Algorithms by leveraging the Poincaré–Bendixson Theorem. Our ODE solutions enable us to derive the steady-state distribution of the number of easy- and hard-to-match agents waiting in the pool, as well as evaluate the social welfare loss functions.
    We find that the loss of the Greedy Algorithm decays in a \emph{linear} rate as the network density parameter $d$ increases, while the loss of the Patient Algorithm  exhibits no slower than an \emph{exponential} decay rate with increasing $d$. These results indicate that the Patient Algorithm outperforms the Greedy Algorithm in terms of social welfare, as evidenced by a reduction in loss functions and an increased number of matched agents. However, this advantage is accompanied by longer waiting times for certain agents, particularly easy-to-match agents, compared to the performance of the Greedy Algorithm.
    \item The ODE models also offer to gain a deeper understanding of the dynamic matching algorithms. For example, our analysis of the Patient Algorithm reveals a significant effect of the proportion of easy-to-match agents and hard-to-match agents on the loss of hard-to-match agents, while its effect on the loss of easy-to-match agents is relatively negligible. Additionally, our results show a \emph{phase transition} in the performance of the Patient Algorithm with respect to the likelihood of encountering a hard-to-match agent, demonstrating the inherent complexity of dynamic matching.
\end{itemize}

\noindent
The rest of the paper is organized as follows. In Section \ref{sec:modelofdynamic}, we describe the problem of dynamic matching agents in heterogeneous networks. In Section \ref{sec:mainresults}, we derive continuous ODE models and provide analytical results for the Greedy and Patient Algorithms. In Section \ref{sec:numericalexamples}, we present numerical simulations and a real data example to validate the analytical results. In Section \ref{sec:relatedworks}, we discuss the related works.  In Section \ref{sec:conclusion}, we conclude the paper with a discussion of future work. All proofs are provided in the Supplementary Appendix.

\section{A Model of Dynamic Matching in Heterogeneous Networks}
\label{sec:modelofdynamic}
In this section, we present a model of a heterogeneous agent matching market on a dynamic network, over a time interval of $[0,T]$. Our focus is on characterizing the evolution of the network and its impact on the matching outcome.

\subsection{Heterogeneous Agents}
\label{sec:agentsinnetwork}
Consider that a new agent's type is randomly selected from $\{A_0,A_1,A_2,\ldots,A_p\}$, a set of $p+1$ types, according to the distribution, 
\begin{equation}
\label{eqn:occuragents}
    \P(\text{type}=A_0)  = 1-p\lambda, \text{ and }
    \P(\text{type}=A_j)  = \lambda, \ \forall j\in\{1,\ldots,n\},
\end{equation}  
where $p\geq 0$, and $0<\lambda< 1/p$. 
We adopt a Poisson process to model the arrival and departure of heterogeneous agents in the matching market \citep{ashlagi2018matching, akbarpour2020thickness}.
The Poisson process is characterized by a rate of arrival, $m\geq 1$.
This choice of model ensures that no two agents arrive simultaneously, almost surely.

We use an independent Poisson process with rate $1$ to
to model the event of an agent becoming \emph{critical}. Upon becoming critical, an unmatched agent immediately departs from the market. If an agent leaves the market without finding a match due to becoming critical, we refer to this as the agent \emph{perishing}. An agent's departure from the market may occur through either being matched with another agent or through perishing.

Let $\SS_t$ be the set of agents in the market at time $t\in[0,T]$, and we are interested in the evolution of $\SS_t$ over $t$. For the sake of clarity, we start our analysis with an empty market, i.e., $\SS_0 = \emptyset$. To capture the heterogeneity of agents, we distinguish between $p+1$ different types of agents, denoted by $A_0, A_1, ..., A_p$. To quantify the number of agents of each type present in the market, we define $|\SS_t(A_j)|$ as the number of $A_j$-type agents in the market at time $t$. Consequently, the total number of agents in the market can be obtained as the sum over all types, i.e., $|\SS_t| = \sum_{k=0}^p|\SS_t(A_k)|$.

\subsection{Compatibility}
\begin{figure}[t!]
\centering
\includegraphics[width=0.35\textwidth]{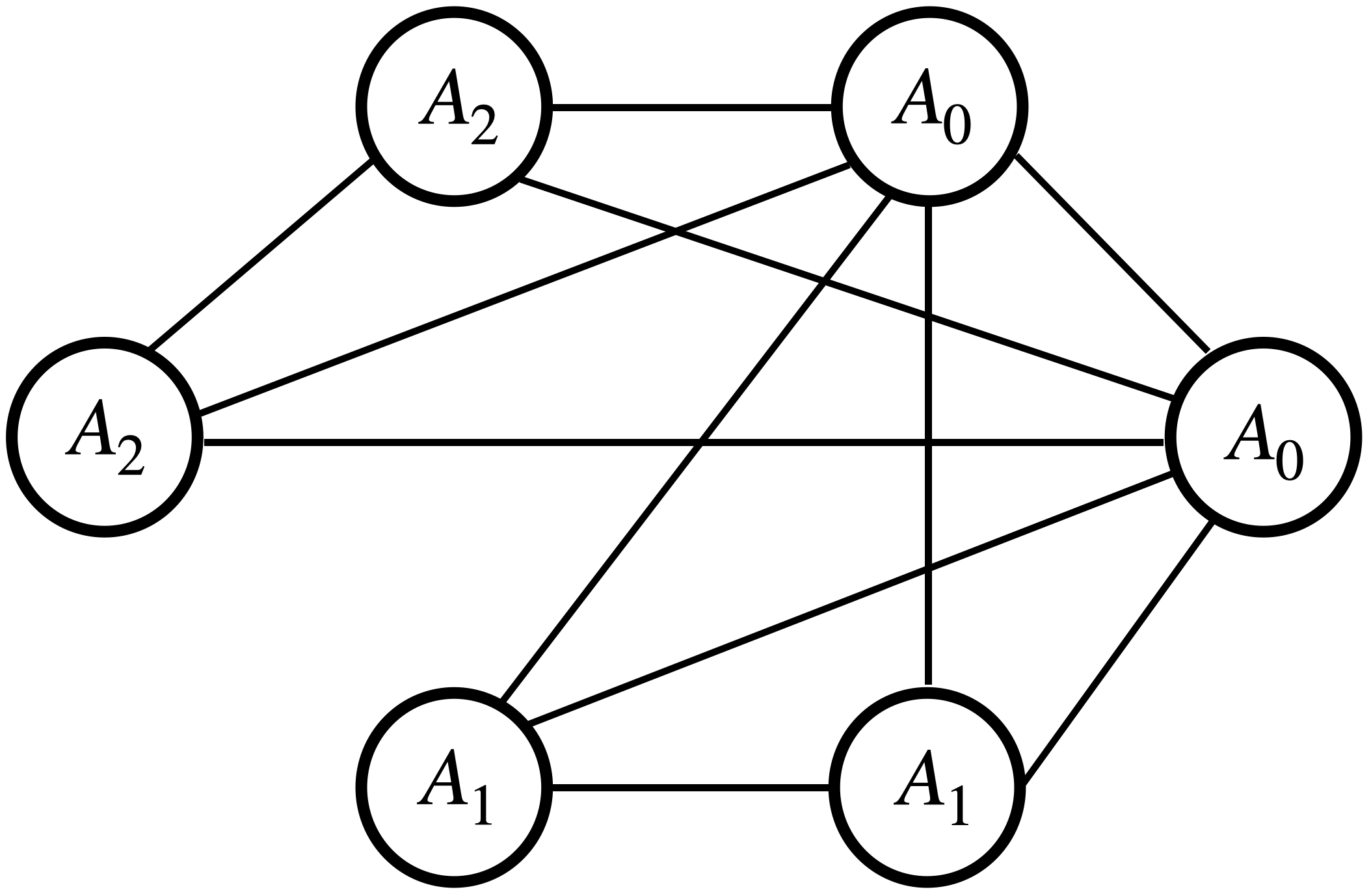}
\caption{An example of a compatibility network with the edges drawn from Eq.~\eqref{eqn:connect}.}
\label{fig:comptexample}
\end{figure}

Let $\alpha\in(0,1)$ be a compatibility parameter of the model.
For a pair of agents in the network, we define that the undirected compatibility depends on agents' types and hence is heterogeneous. For any $j\neq j'\in\{1,\ldots,p\}$,
\begin{equation}
\label{eqn:connect}
\begin{aligned}
    &\P\left(\{A_0,A_0\}\text{ are compatible}\right)=\alpha, \ \  \P\left(\{A_j,A_j\}\text{ are compatible}\right)=\alpha,\\
    & \P\left(\{A_0,A_j\} \text{ are compatible}\right)=\alpha,\ \ \P\left(\{A_j,A_{j'}\} \text{ are compatible}\right)=0.
\end{aligned}
\end{equation}
In model \eqref{eqn:connect}, we assume that the compatibility probabilities between agents are independent across different pairs. 
We also define the parameter,
\begin{equation}
\label{eqn:densityd}
    d \equiv \alpha m,
\end{equation}
and $d$ represents the density in the network. 
An agent with an $A_0$ type is considered \emph{easy-to-match}, as they are compatible with all other agent types with probability $\alpha$. In contrast, an agent with an $A_j$ type, where $j\geq 1$, is considered \emph{hard-to-match}, as they are only compatible with $A_0$- or $A_j$-type agents with probability $\alpha$. This distinct compatibility structure is shown in Figure \ref{fig:comptexample} for an example with $\alpha = 1$.

We note that the models in Eqs.~\eqref{eqn:occuragents} and \eqref{eqn:connect} differ from those in \citet{ashlagi2018matching} in several significant ways. First, our definition of easy-to-match and hard-to-match agents is more versatile and practical, allowing for multiple types of hard-to-match agents, while \citet{ashlagi2018matching} only allows for one type of hard-to-match agent. Second, our compatibility definition accommodates a sparse graph, where the matching probability $\alpha$ approaches zero as the arrival rate $m$ increases, while in \citet{ashlagi2018matching}, the matching probability $\alpha$ remains constant. Third, as a result of the differences in the models, our approach entails constructing a discrete dynamic model and then deriving a corresponding continuous ODE model to analyze the steady-state, as opposed to directly analyzing the discrete model in \citet{ashlagi2018matching}.

Given a time $t\geq 0$, let $\EE_t \subseteq \SS_t \times \SS_t$ be the set of compatible pairs of agents in the market, and $\GG_t=(\SS_t,\EE_t)$ be the resulting network representation. A collection of edges $\MM_t \subseteq \EE_t$ is considered a valid \emph{matching} if for any pair of edges $\{(a,b), (c,d)\} \subseteq \EE_t$, either $\{a,b\}=\{c,d\}$ or $\{a,b\} \cap \{c,d \}=\emptyset$. At each time $t\geq 0$, a \emph{matching algorithm} selects a valid matching $\MM_t$ from the network $\GG_t$, which could be an empty set, and the agents in $\MM_t$ immediately leave the market. It is assumed that the planner, at any time $t\geq 0$, has access to only the information of $\{\GG_{t^{'} },t^{'} \leq t \}$, and is unaware of $\{\GG_{t^{'} },t^{'} > t \}$. For an agent $a \in \SS_t$, the set of its neighbors in $\GG_t$ is denoted as $\NN(a)\subseteq \SS_t$.

\subsection{Goal}

Our objective is to optimize social welfare by maximizing the number of agents who depart the market matched. This translates to finding a matching algorithm that minimizes the number of perished agents, given the constraints and conditions of the dynamic network. We define the \emph{total loss} as follows,
\begin{equation}
\label{eqn:lossratio}
    L^{\text{alg}} \equiv  \frac{\E\left[\left|\cup_{t\leq T}\SS_t-\bar{\SS}_T^{\text{alg}}-\SS_{T}\right|\right]}{\E\left[\left|\cup_{t\leq T}\SS_t\right|\right]}.
\end{equation}
Here 
$\bar{\SS}^{\text{alg}}_T$ denotes the set of matched agents at the end of the time period $T$ as determined by the chosen algorithm,
\begin{equation*}
    \bar{\SS}^{\text{alg}}_T \equiv \left\{a\in \cup_{t\leq T}\SS_t: a \text{ is matched by ``alg" by time $T$}\right\}.
\end{equation*}
Thus, the total loss $L^{\text{alg}}$ in Eq.~\eqref{eqn:lossratio} is calculated as the ratio of the expected number of perished agents to the expected total number of agents present in the market over time $T$. 
For the $A_k$-type agents, $k\in{0,1,\ldots,p}$, we define the set of matched agents by time $T$ as $\bar{\SS}^{\text{alg}}_T(A_k) \equiv \{a\in\SS(A_k): a \text{ is matched by ``alg" by time $T$}\}$. The \emph{individual loss} for each type of agent is computed as follows,
\begin{equation}
\label{eqn:lossforeachagent}
L^{\text{alg}}(A_k)  \equiv  \frac{\E\left[\left|\cup_{t\leq T}\SS_t(A_k)-\bar{\SS}_T^{\text{alg}}(A_k)-\SS_T(A_k)\right|\right]}{\E\left[\left|\cup_{t\leq T}\SS_t(A_k)\right|\right]}.
\end{equation}

The evaluation of matching algorithms in a dynamic market modeled as a discrete-time process, as described in Section \ref{sec:agentsinnetwork}, presents four major challenges. First, the non-stationary nature of the available agent pool, arising from the dynamic arrival and departure of agents, makes it difficult to accurately evaluate the loss function defined in Eqs.~\eqref{eqn:lossratio} and \eqref{eqn:lossforeachagent} under different algorithms. Second, the heterogeneity of agents with varying compatibility and loss functions adds to the complexity of the analysis. Third, the absence of complete information on future agent arrivals and departures creates uncertainty in predicting the market state and evaluating the performance of matching algorithms. Finally, the interdependencies between matchings and loss functions, resulting from the impact of one agent's matching on others, further complicate the analysis. To mitigate these technical difficulties,  we establish a continuous approximation by formulating Ordinary Differential Equation (ODE) models in Section \ref{sec:mainresults}. This allows us to perform a more analytical evaluation of the various matching algorithms in the dynamic market.


\subsection{Greedy and Patient Algorithms}
In the design of matching algorithms for maximizing social welfare, two entangling factors must be considered. The first factor is the \emph{timing} of matchings, as the planner must determine when to match agents based on the current network structure rather than having full knowledge of the future network. The second factor is the \emph{match partner selection}, which influences the future network structure. For instance, in a market with agents $\{a,b,c\}$ at time $t$, where the compatible pairs are $\{a,b\}$ and $\{a,c\}$, but $b$ and $c$ are not compatible. If agent $a$ becomes critical at time $t$, the choice of matching $\{a,b\}$ or $\{a,c\}$ is inconsequential. However, if a new agent $d$ arrives at time $t+1$ and is compatible with $c$ but not with $b$, then the choice of matching $\{a,c\}$ at time $t$ excludes the possibility of matching either $b$ or $d$ at time $t+1$. In this section, we study two algorithms that separate the effects of timing and match partner selection in heterogeneous networks. 
\begin{algorithm}
\caption{ \normalsize{{Greedy Algorithm for heterogeneous dynamic matching}}}
\label{alg:greedyalg}
\begin{algorithmic}[1]
\State \textbf{for} a new agent $a_t$ arrives the market at time $t\geq 0$ \textbf{do}
\State \quad\quad \textbf{if} $\NN_1(a_t) \equiv \{b\in \NN(a_t)~|~b\text{'s type is } A_j, \forall j\geq 1 \}  \neq \emptyset$:
\State \quad\quad\quad\quad randomly choose $b\in \NN_1(a_t)$ at uniform, and match $b$ with $a_t$.
\State \quad\quad \textbf{else if} $\NN_2(a_t) \equiv \{b\in \NN(a_t)~|~b\text{'s type is } A_0, \forall j\geq 1 \}  \neq \emptyset$:
\State \quad\quad\quad\quad randomly choose $b\in \NN_2(a_t)$ at uniform, and match $b$ with $a_t$.
\State \quad\quad \textbf{else}:
\State \quad\quad\quad\quad  agent $a_t$ stays at the pool.
\State \textbf{end for}
\end{algorithmic}
\end{algorithm}

The first algorithm, the \emph{Greedy Algorithm}, aims to match an agent upon their arrival in the market. Under the Poisson model for agent arrival in the network in Section \ref{sec:agentsinnetwork}, there are no two agents arriving at the same time. The algorithm matches the arrival agent with its neighbors in the pool, following the rule that a compatible hard-to-match agent (i.e., $A_1,\ldots, A_p$-type) is matched \emph{before} considering a compatible easy-to-match agent (i.e., $A_0$-type). The procedure is outlined in Algorithm \ref{alg:greedyalg}. In cases where the cost of waiting is negligible, it is weakly better to leave $A_0$-type agents in the pool, as they are more likely to be compatible with future agents. The Greedy Algorithm operates in a \emph{local} fashion, considering only the immediate neighbors of the arrival agents and ignoring the global or future network structure.

The second algorithm, the \emph{Patient Algorithm},  operates by matching agents upon they become critical. Under the Poisson model described in Section \ref{sec:agentsinnetwork}, no two agents would become critical at the same time almost surely. When a critical agent emerges, the algorithm evaluates its neighboring agents in the pool and prioritizes a matching with a compatible hard-to-match agent over an easy-to-match agent. The procedure is described in Algorithm \ref{alg:patientalg}. Similar to the Greedy Algorithm, the Patient Algorithm operates with a local perspective, only considering the neighbors of the critical agents.

\begin{algorithm}
\caption{ \normalsize{{Patient Algorithm for heterogeneous dynamic matching}}}
\label{alg:patientalg}
\begin{algorithmic}[1]
\State \textbf{for} an agent $a_t$ becomes critical at time $t\geq 0$ \textbf{do}
\State \quad\quad \textbf{if} $\NN_1(a_t) \equiv \{b\in \NN(a_t)~|~b\text{'s type is } A_j, \forall j\geq 1 \}  \neq \emptyset$:
\State \quad\quad\quad\quad randomly choose $b\in \NN_1(a_t)$ at uniform, and match $b$ with $a_t$.
\State \quad\quad \textbf{else if} $\NN_2(a_t) \equiv \{b\in \NN(a_t)~|~b\text{'s type is } A_0, \forall j\geq 1 \}  \neq \emptyset$:
\State \quad\quad\quad\quad randomly choose $b\in \NN_2(a_t)$ at uniform, and match $b$ with $a_t$.
\State \quad\quad \textbf{else}:
\State \quad\quad\quad\quad agent $a_t$ perishes.
\State \textbf{end for}
\end{algorithmic}
\end{algorithm}

The key distinction between the Greedy Algorithm and the Patient Algorithm lies in the utilization of different information: the former leverages the information of agents arriving at the market, while the latter requires information of critical agents. These algorithms also differentiate themselves from their counterparts in networks comprising solely single-type agents \citep{akbarpour2020thickness}, as they must consider the compatibility of match partners and prioritize the matching of hard-to-match agents over easy-to-match agents in heterogeneous networks. Proposition \ref{thm:matchwithwho} presents a rigorous analytical evaluation of the probabilities of matching different types of agents with these algorithms. 

\begin{proposition}
\label{thm:matchwithwho}
Consider that at time $t\in[0,T]$, the planner tries to match an $A_k$-type agent of interest, $k\in\{0,1,\ldots,p\}$. This event happens when the agent of interest enters the market at time $t$ while the planner uses Greedy Algorithm, or when the agent of interest becomes critical at time $t$ while the planner uses Patient Algorithm.
Let $\pi^*_t(A_{k},A_{k'})$ be the probability that the planner matches the $A_{k}$-type agent of interest with an $A_{k'}$-type agent who is in the pool, where $k'\in\{0,1,\ldots,p\}$. Then for any $j\geq 1$,
\begin{equation*}
\begin{aligned}
    & \pi^*_t(A_0,A_0) && = \quad \left[1-(1-\alpha)^{|\SS_t(A_0)|}\right](1-\alpha)^{\sum_{k=1}^{p} |\SS_t(A_{k})|},\\
    & \pi^*_t(A_0,A_j) && = \quad \left[1-(1-\alpha)^{|\SS_t(A_{j})|}\right]\sum_{k=0}^{p-1} \frac{1}{k+1} \sum_{\{j_1,\ldots,j_k\}\subseteq [p]\setminus\{j\} } \beta_{j,j_1,\ldots,j_k},\\
    & \pi^*_t(A_j,A_0) && = \quad \left[1-(1-\alpha)^{|\SS_t(A_{0})|}\right](1-\alpha)^{|\SS_t(A_{j})|},\\
   & \pi^*_t(A_j,A_j) && = \quad  1-(1-\alpha)^{|\SS_t(A_{j})|},
\end{aligned}   
\end{equation*}
where $[p]\equiv\{1,\ldots,p\}$, and 
\begin{equation*}
    \beta_{j,j_1,\ldots,j_k}\equiv  \Pi_{s=1}^k\left[1-(1-\alpha)^{|\SS_t(A_{j_s})|}\right]\Pi_{l\not\in\{j,j_1,\ldots,j_k\}}(1-\alpha)^{|\SS_t(A_{l})|}.
\end{equation*}
\end{proposition}

Note that in Proposition \ref{thm:matchwithwho}, the probability function $ \pi^*_t$ is \emph{asymmetric} with respect to its covariates,  i.e., $\pi^*_t(A_{k},A_{k'})\neq \pi^*_t(A_{k'},A_{k})$, for any $k\neq k'\in\{0,1,\ldots,p\}$. This is because by definition of $\pi^*_t(A_{k},A_{k'})$, the covariate order matters. The first covariate, $A_{k}$, refers to the agent of interest, while the second covariate, $A_{k'}$, refers to the potential match for the agent of interest.
We also note that the probabilities in Proposition \ref{thm:matchwithwho} satisfy,
\begin{equation}
\label{eqn:a0incompat}
    (1-\alpha)^{|\SS_t|} = 1-\sum_{k=0}^p\pi_t^*(A_0,A_k),
\end{equation}
which can be explained as follows.
The compatibility of an $A_0$-type agent is captured by both sides of Equation \eqref{eqn:a0incompat}. On one hand, the probability that it is incompatible with any other agent in the network can be calculated as $(1-\alpha)^{|\SS_t|}$, as stated in Eq.~\eqref{eqn:connect}. On the other hand, the probability of compatibility with any other type of agent, $A_k$ where $k\geq 0$, is given by the sum of the probabilities of matching with each type, $\sum_{k=0}^p\pi_t^*(A_0,A_k)$. This latter quantity provides the right-hand-side of Equation \eqref{eqn:a0incompat}.

\section{Main Results}
\label{sec:mainresults}
In this section, we establish the asymptotic equivalence between the discrete and the continuous ODE models for dynamic matching. We demonstrate the existence of solutions to the ODEs, and using these solutions, we derive the long-term behavior of agents under both Greedy and Patient Algorithms. Our results demonstrate the power of ODEs as a framework for understanding and analyzing dynamic matching algorithms, and demonstrate that this conceptually simpler mathematical model can provide insights into the long-term behavior of agents in a market.
\subsection{Analysis of the Greedy Algorithm}

We now derive the ODE models for the Greedy Algorithm in Algorithm \ref{alg:greedyalg}. The goal is to analyze the size of agents in the market, $|\SS_t(A_k)|$, for each $k=0,1,\ldots,p$.
By taking a small step size in time $\Delta t>0$, we define the derivative of $|\SS_t(A_k)|$ as,
\begin{equation}
\label{eqn:defofder}
     \frac{d|\SS_t(A_k)|}{dt} \equiv \lim_{\Delta t \to 0_+} \frac{\E[|\SS_{t+\Delta t}(A_k)|]-|\SS_t(A_k)|}{\Delta t},\quad \forall k=0,1,\ldots,p.
\end{equation}
Here we consider the expected change of the set size, $\E[|\SS_{t+\Delta t}(A_j)|]-|\SS_t(A_j)|$, given the information at time $t$. This is different from the change of the set size, $|\SS_{t+\Delta t}(A_j)|-|\SS_t(A_j)|$ itself. The definition \eqref{eqn:defofder} would yield deterministic ODEs and facilitate the analysis, compared to obtaining stochastic ODEs otherwise. 
We derive the ODEs for $|\SS_t(A_0)|$, and $|\SS_t(A_j)|$ with $j\geq 1$, separately. 
\begin{theorem}
\label{thm:Greedy_limit}
Given the set of the agents' sizes, $\{|\SS_t(A_0)|,|\SS_t(A_1)|,\cdots,|\SS_t(A_p)|\}$ at time $t$, the Greedy Algorithm in Algorithm \ref{alg:greedyalg} yields that  for $A_0$-type agents,
\begin{equation*}
\begin{aligned}
     \frac{d|\SS_t(A_0)|}{dt}  & =   (1-p\lambda)m \left[1-\sum_{k=0}^p\pi_t^*(A_0,A_k)\right] \\
     & \quad \quad - (1-p\lambda)m \pi_t^*(A_0,A_0) - \lambda m\sum_{k=1}^{p} \pi_t^*(A_k,A_0) 
    - |\SS_t(A_0)|,
\end{aligned}
\end{equation*}
where $\lambda$ and $d|\SS_t(A_0)|/dt$ are defined by Eqs.~\eqref{eqn:occuragents} and \eqref{eqn:defofder}, respectively.
Moreover, the Greedy Algorithm yields that  for $A_j$-type agents,
\begin{equation*}
\begin{aligned}
    \frac{d|\SS_t(A_j)|}{dt} &  = 
    - (1-p\lambda)m\pi_t^*(A_0,A_j) -\lambda m\pi_t^*(A_j,A_j)\\
    & \quad \quad +\lambda m \left[1-\pi_t^*(A_j,A_0) - \pi_t^*(A_j,A_j)\right]   - |\SS_t(A_j)|,
\end{aligned}
\end{equation*}
where $j= 1,\ldots,p$.
\end{theorem}
\noindent
The ODEs in Theorem \ref{thm:Greedy_limit} describe the dynamic behavior of the number of agents of each type in the market over time. Specifically, the ODEs capture the influx of new agents into the market and the matching and departure of agents that become critical. The terms in the ODEs capture the fraction of agents of each type that are matched with agents of other types, the fraction that remain in the market without matching, and the fraction that leave the market due to becoming critical, which are explained as follows.
When $(1-p\lambda) m\Delta t$ of $A_0$-agents arrive at the market during $[t,t+\Delta t]$, a fraction $\pi_t^*(A_0,A_0)$ of these agents are matched with $A_0$-type agents, a fraction $\pi_t^*(A_0,A_j)$ are matched with $A_j$-type agents, and fraction $1-\sum_{k=0}^p\pi_t^*(A_0,A_k)$ agents remain in the market.
Similarly, when $\lambda m\Delta t$ of $A_j$-agents arrive at the market during $[t,t+\Delta t]$, a fraction $\pi_t^*(A_j,A_0)$ of these agents are matched with $A_0$-type agents, a fraction $\pi_t^*(A_j,A_j)$ are matched with $A_j$-type agents, and a fraction $1-\pi_t^*(A_j,A_0)-\pi_t^*(A_j,A_j)$ remian in the market. Additionally, a total of $|\SS_t(A_k)| \Delta t$  of $A_k$-type agents become critical and leave the market during  $[t,t+\Delta t]$, for any $k\geq 0$.

\subsubsection{Evaluation of the Greedy Algorithm}
It is of interest to study the existence of solutions to ODEs in Theorem  \ref{thm:Greedy_limit}. Furthermore, we  aim to assess the performance of the Greedy Algorithm  through the evaluation of the loss functions defined in Eqs.~\eqref{eqn:lossratio} and \eqref{eqn:lossforeachagent}.

\begin{theorem}
\label{main_result_greedy}
Given any initial values $(|\SS_0(A_0)|,|\SS_0(A_1)|,\ldots,|\SS_0(A_p)|)$ with $|\SS_0(A_k)|\geq 0$ and $k\geq 0$,
the ODEs in Theorem \ref{thm:Greedy_limit} have a unique solution. In addition,
\begin{itemize}
    \item if the initial values satisfy $|\SS_0(A_j)|=|\SS_0(A_{j'})|\geq 0$ for all $1\leq j\neq j'\leq p$, then
    $|\SS_t(A_j)|=|\SS_t(A_{j'})|$ for any $t> 0$;
    \item if  $|\SS_0(A_j)|=|\SS_0(A_{j'})|\geq 0$ for all $1\leq j\neq j'\leq p$ and $m\to\infty,\alpha\to 0$, then there exists a stationary solution to the ODEs in Theorem \ref{thm:Greedy_limit} when $T\to\infty$. Moreover, there exist constants $0<c\leq c'<\infty$ such that the loss functions in Eqs.~\eqref{eqn:lossratio} and \eqref{eqn:lossforeachagent}  satisfy,
\begin{equation*}
\begin{aligned}
& cd^{-1}\leq \quad \lim_{T \to \infty} L^{\text{greedy}}(A_k) && \leq c'd^{-1},\ \forall k\geq 0,\quad \text{and}\\
& cd^{-1}\leq \quad \lim_{T \to \infty} L^{\text{greedy}}
&& \leq c'd^{-1}.
\end{aligned}
\end{equation*}
Here $d$ is the density parameter in Eq.~\eqref{eqn:densityd}.
\end{itemize}
\end{theorem}
The proof of Theorem \ref{main_result_greedy} leverages the Poincaré–Bendixson Theorem \citep[e.g.,][]{ciesielski2012poincare} to demonstrate the convergence of the solution of the ODE to the stationary point in the limit as $T\to\infty$.
We make two remarks on Theorem \ref{main_result_greedy}. 
First, the case where $m\to\infty$ represents a large market scenario, characterized by an average of $m$ agents entering the market per unit time.
Second, our framework, characterized by a compatibility network with decreasing matching probabilities $\alpha$ that approach zero with increasing number of agents, aligns with the setting presented in \citet{akbarpour2020thickness}. Theorem \ref{main_result_greedy} demonstrates the validity of our results for any density parameter $d=\alpha m\in(0,\infty)$. This sparsity of the network, in contrast to the fixed constant matching probabilities assumed in \citet{ashlagi2018matching}, is a distinct feature of our framework.

\subsection{Analysis of the Patient Algorithm}

We derive the ODE models for the Patient Algorithm in Algorithm \ref{alg:patientalg}. 
\begin{theorem}
\label{thm:Patientlimit}
Given the set of the agents' sizes, $\{|\SS_t(A_0)|,|\SS_t(A_1)|,\cdots,|\SS_t(A_p)|\}$ at time $t$, the Patient Algorithm in Algorithm \ref{alg:patientalg} yields that  for $A_0$-type agents,
\begin{equation*}
\begin{aligned}
     \frac{d|\SS_t(A_0)|}{dt}  & =   (1-p\lambda)m - \sum_{k=0}^p|\SS_t(A_k)|\pi_t^*(A_k,A_0) - |\SS_t(A_0)|.
\end{aligned}
\end{equation*}
Moreover, the Patient Algorithm yields that  for $A_j$-type agents,
\begin{equation*}
\begin{aligned}
    \frac{d|\SS_t(A_j)|}{dt} &  = \lambda m  - |\SS_t(A_0)|\pi_t^*(A_0,A_j) - |\SS_t(A_j)| \pi_t^*(A_j,A_j) -|\SS_t(A_j)|,
\end{aligned}
\end{equation*}
where $j= 1,\ldots,p$.
\end{theorem}
The ODEs in Theorem \ref{thm:Patientlimit} model the dynamic evolution of the number of each agent type in the market as time progresses under the Patient Algorithm.
In particular, when $|\SS_t(A_0)|\Delta t$ of $A_0$-agents become critical during $[t,t+\Delta t]$, a fraction $\pi_t^*(A_0,A_0)$ of these agents are matched with $A_0$-type agents, a fraction $\pi_t^*(A_0,A_j)$ are matched with $A_j$-type agents ($j\geq 1$), and a fraction $1-\sum_{k=0}^p\pi_t^*(A_0,A_k)$ leave the market without matching. Similarly, when $|\SS_t(A_j)|\Delta t$ of $A_j$-agents ($j\geq 1$) become critical during $[t,t+\Delta t]$, a fraction $\pi_t^*(A_j,A_0)$ of these agents are matched with $A_0$-type agents, a fraction $\pi_t^*(A_j,A_j)$ are matched with $A_j$-type agents, and a fraction $1-\pi_t^*(A_j,A_0)-\pi_t^*(A_j,A_j)$ leave the market without matching. Additionally, there are a total of $(1-p\lambda)m \Delta t$ of $A_0$-type agents and a total of $\lambda m\Delta t$ of $A_j$-type agents entering the market during each interval $[t,t+\Delta t]$.  These results provide an intuitive explanation for the terms in the ODEs in Theorem \ref{thm:Patientlimit}.

\subsubsection{Evaluation of the Patient Algorithm}
We now establish the existence of solutions to the ODEs presented in Theorem \ref{thm:Patientlimit}. We also evaluate the performance of the Patient Algorithm using two metrics, defined in Eqs.~\eqref{eqn:lossratio} and \eqref{eqn:lossforeachagent}. 

\begin{theorem}
\label{main_result_patient}
Given any initial values $(|\SS_0(A_0)|,|\SS_0(A_1)|,\ldots,|\SS_0(A_p)|)$ with $|\SS_0(A_k)|\geq 0$ and $k\geq 0$,
the ODEs in Theorem \ref{thm:Patientlimit} have a unique solution. In addition,
\begin{itemize}
    \item if the initial values satisfy  $|\SS_0(A_j)|=|\SS_0(A_{j'})|\geq 0$ for all $1\leq j\neq j'\leq p$, then
     $|\SS_t(A_j)|=|\SS_t(A_{j'})|\geq 0$ for any $t> 0$;
    \item if $|\SS_0(A_j)|=|\SS_0(A_{j'})|\geq 0$ for all $1\leq j\neq j'\leq p$ and  $m\to\infty,\alpha\to 0$, then there exists a stationary solution to the ODEs in Theorem \ref{thm:Patientlimit} when $T\to\infty$. Moreover, the loss functions in Eq.~\eqref{eqn:lossforeachagent}  satisfy that as $d\to\infty$,
    \begin{equation*}
    \lim_{T\to\infty}L^{\text{patient}}(A_0) = e^{\left[-\frac{1}{2}+o(1)\right]d},
    \end{equation*}
    and for any $j=1,\ldots,p$,
     \begin{equation*}
    \lim_{T\to\infty}L^{\text{patient}}(A_j) = 
        \begin{cases}
         e^{-\left[1-\frac{1}{2p}-(p-1)\lambda+o(1)\right]d},   & \text{if }\lambda p> \frac{1}{2},\\
         e^{\left[-\frac{1}{2}+o(1)\right]d},   & \text{if }\lambda p\leq \frac{1}{2}.
        \end{cases}
    \end{equation*}
    Moreover, the total loss in Eq.~\eqref{eqn:lossratio} satisfies that as $d\to\infty$,
    \begin{equation*}
    \lim_{T\to\infty}L^{\text{patient}} = 
        \begin{cases}
         e^{-\left[1-\frac{1}{2p}-(p-1)\lambda+o(1)\right]d}   & \text{if }\lambda p> \frac{1}{2},\\
         e^{\left[-\frac{1}{2}+o(1)\right]d}   & \text{if }\lambda p\leq \frac{1}{2}.
        \end{cases}
    \end{equation*}
\end{itemize}
\end{theorem}
We make two remarks on this theorem. First, the proportion of easy-to-match agents $A_0$ and hard-to-match agents $A_j$ with $j\geq 1$ has a significant effect on the loss of hard-to-match agents when using the Patient Algorithm, while its effect on the loss of easy-to-match agents is relatively  \emph{negligible}. This is exemplified by the following result, which can be derived when $m \to \infty, \alpha\to 0$, and $d \to \infty$,
\begin{align*}
 \forall   0<\lambda_{1}< \lambda_{2}<\frac{1}{p}, \quad  \frac{\lim_{T\to\infty}L^{\text{patient}}(A_0)\text{ with $\lambda =\lambda_1$}}{\lim_{T\to\infty}L^{\text{patient}}(A_0)\text{ with $\lambda =\lambda_2$}} =\Theta(1).
\end{align*}
However, for agents $A_j$ with $j\geq 1$, and $m \to \infty, \alpha\to 0$, $d \to \infty$, we have
\begin{align*}
 \forall   0<\lambda_{1}< \frac{1}{2p}\leq \lambda_{2}<\frac{1}{p}, \quad  \frac{\lim_{T\to\infty}L^{\text{patient}}(A_j)\text{ with $\lambda =\lambda_1$}}{\lim_{T\to\infty}L^{\text{patient}}(A_j)\text{ with $\lambda =\lambda_2$}} \to 0.
\end{align*}

Second, Theorem \ref{main_result_patient} also shows a critical relationship between the parameter $\lambda$ and the performance of the Patient Algorithm, marked by a \emph{phase transition} phenomenon at $\lambda p = 1/2$. When $\lambda p \leq 1/2$, the loss function for hard-to-match agents is expected to vary gradually with changes in $\lambda$, remaining at $e^{\left[-\frac{1}{2}+o(1)\right]d}$. Conversely, when $\lambda p > 1/2$, the loss function decays at a faster rate, as described by $e^{-\left[1-\frac{1}{2p}-(p-1)\lambda+o(1)\right]d}$.
This latter function  is rapidly increasing with respect to $\lambda$, such that for any $\lambda_1 < \lambda_2$, then the relative loss,
\begin{equation*}
    \frac{e^{-\left[1-\frac{1}{2p}-(p-1)\lambda_2+o(1)\right]d}}{e^{-\left[1-\frac{1}{2p}-(p-1)\lambda_1+o(1)\right]d}}\to\infty,\quad\text{as }d\to\infty. 
\end{equation*}
As a result, even small changes in $\lambda$ can lead to significant changes in the loss when $\lambda p >1/2$. This phase transition phenomenon is also demonstrated in the simulation results.

\subsection{Comparison of the Greedy and Patient Algorithms}
\label{sec:compgreedpat}

By Theorems \ref{main_result_greedy} and \ref{main_result_patient},  the loss of the Greedy Algorithm exhibits a \emph{linear} decay rate  in $d$ with a constant $c$,
\begin{equation*}
   \lim_{T \to \infty} L^{\text{greedy}}\geq  \frac{c}{d}.
\end{equation*}
In contrast, the loss of the Patient Algorithm decays no slower than \emph{exponential} in $d$, 
\begin{equation*}
   \lim_{T \to \infty} L^{\text{patient}}= 
        \begin{cases}
         e^{-\left[1-\frac{1}{2p}-(p-1)\lambda+o(1)\right]d},   & \text{if }\lambda p> \frac{1}{2},\\
         e^{\left[-\frac{1}{2}+o(1)\right]d},   & \text{if }\lambda p\leq \frac{1}{2}.
        \end{cases}
\end{equation*}
This result shows that the Patient Algorithm outperforms the Greedy Algorithm in enhancing social welfare, as evidenced by a faster decay rate in the loss function. 

Next, we examine the average waiting time of agents under the two algorithms. 
\begin{proposition}
\label{prop:waittime}
Suppose the initial values $(|\SS_0(A_0)|,|\SS_0(A_1)|, \ldots,|\SS_0(A_p)|)$ satisfy $|\SS_0(A_j)|=|\SS_0(A_{j'})|\geq 0$   for all $1\leq j\neq j'\leq p$, and $m\to\infty,\alpha\to 0$. 
Then,
\begin{itemize}
    \item ODEs for the Greedy Algorithm in Theorem \ref{thm:Greedy_limit} suggest that when $T\to\infty$,  the average time that an
$A_k$-type agent  spends in the market is
$   \Theta\left(\frac{1}{d}\right)$, for any $k =0,\ldots,p$;
    \item  ODEs for the Patient Algorithm in Theorem \ref{thm:Patientlimit} suggest that when $T\to\infty$, 
 the average time that an $A_0$-type agent and an $A_j$-type agent ($j\geq 1$) spend in the market converge towards, respectively,
\begin{equation*}
\begin{cases}
    \Theta\left(1\right)  \text{ and } \ \Theta\left(1\right), & \text{ if }\lambda p>\frac{1}{2};\\
    \Theta\left(1\right)  \text{ and } \ \Theta\left(\frac{1}{d}\right), & \text{ if }\lambda p\leq \frac{1}{2}.
\end{cases}
\end{equation*}
\end{itemize}
\end{proposition}
\noindent
This proposition demonstrates the advantage of the Greedy Algorithm for an agent, particularly an $A_0$-type agent, in terms of reduced waiting time in the market when $d\to\infty$. 

The combination of Theorems \ref{main_result_greedy} and \ref{main_result_patient}, along with Proposition \ref{prop:waittime}, provides clear evidence that the Patient Algorithm offers an improvement in social welfare, as shown by the increased number of matched agents exiting the market. However, this improvement comes at the cost of prolonged waiting times for specific agents, particularly the $A_0$-type agents, compared to the performance of the Greedy Algorithm.

\section{Numerical Examples}
\label{sec:numericalexamples}
In this section, we provide simulation examples to support our theoretical findings, and further explore the implications of our results through a case study of real-world data on kidney exchanges.
\subsection{Discrete and Continuous Models}
\label{sec:discretevsconti}

\begin{figure}[ht!]
    \centering
    \includegraphics[width=0.8\textwidth]{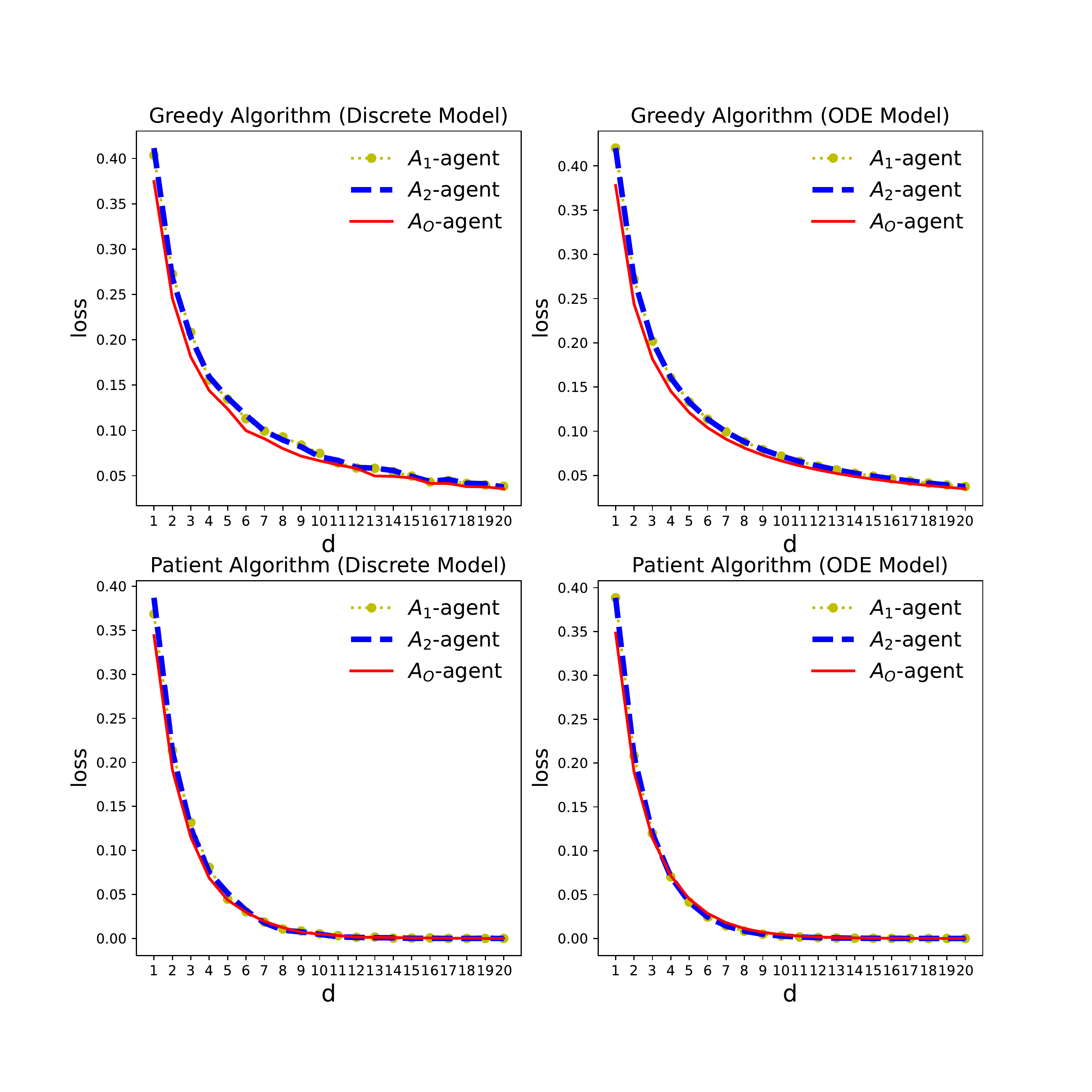}
    \vspace{-0.5in}
    \caption{The loss functions with $m=8,000,\lambda=0.2$.}
    \label{dchange}
\end{figure}

\noindent
We consider a dynamic market with  $p=2$, which implies that there are two distinct types of hard-to-match agents, denoted as $A_1$- and $A_2$-agents, and one type of easy-to-match agents, referred to as $A_0$-agents. These agents engage in market interactions, and the matching process among them is governed by the rules specified in Eq.~\eqref{eqn:connect}. By examining the behavior and performance of these agents under different algorithms, we aim to gain deeper understanding of the market dynamics and its effect on social welfare.

We evaluate the performance of the discrete models implemented in Algorithm \ref{alg:greedyalg} (Greedy Algorithm) and Algorithm \ref{alg:patientalg} (Patient Algorithm) through simulation, and compare it to the predictions made by the continuous models described in Theorem \ref{thm:Greedy_limit} (Greedy Algorithm) and Theorem \ref{thm:Patientlimit} (Patient Algorithm). The discrete model simulations record the state transitions of the discrete-state Markov chain, where the number of transitions approximates the time $T$. A length of $20,000$ transitions is considered, with the last $5,000$ being used to calculate the loss as they more accurately approximate the steady-state distribution. The continuous model simulations are solved using the numerical solver \texttt{scipy.integrate.odeint} in \texttt{Python}.
In order to quantify the performance of the algorithms, we utilize the loss function for individual agent types as defined in Eq.~\eqref{eqn:lossforeachagent}.  In this simulation, we fix the market size $m=8,000$ and the parameter $\lambda=0.2$.

Figure \ref{dchange} shows the comparison of the result, which demonstrates the efficacy of the continuous ODE models in accurately predicting the behavior of the discrete dynamic models. 
This corroborates the  asymptotic equivalence between the discrete and the continuous ODE models for dynamic matching in Theorems \ref{thm:Greedy_limit} and \ref{thm:Patientlimit}.
Moreover, it is seen that a higher density $d$ of compatible agents in the market results in a better performance in terms of the loss function.

\begin{figure}[ht!]
    \centering
    \includegraphics[width=0.8\textwidth]{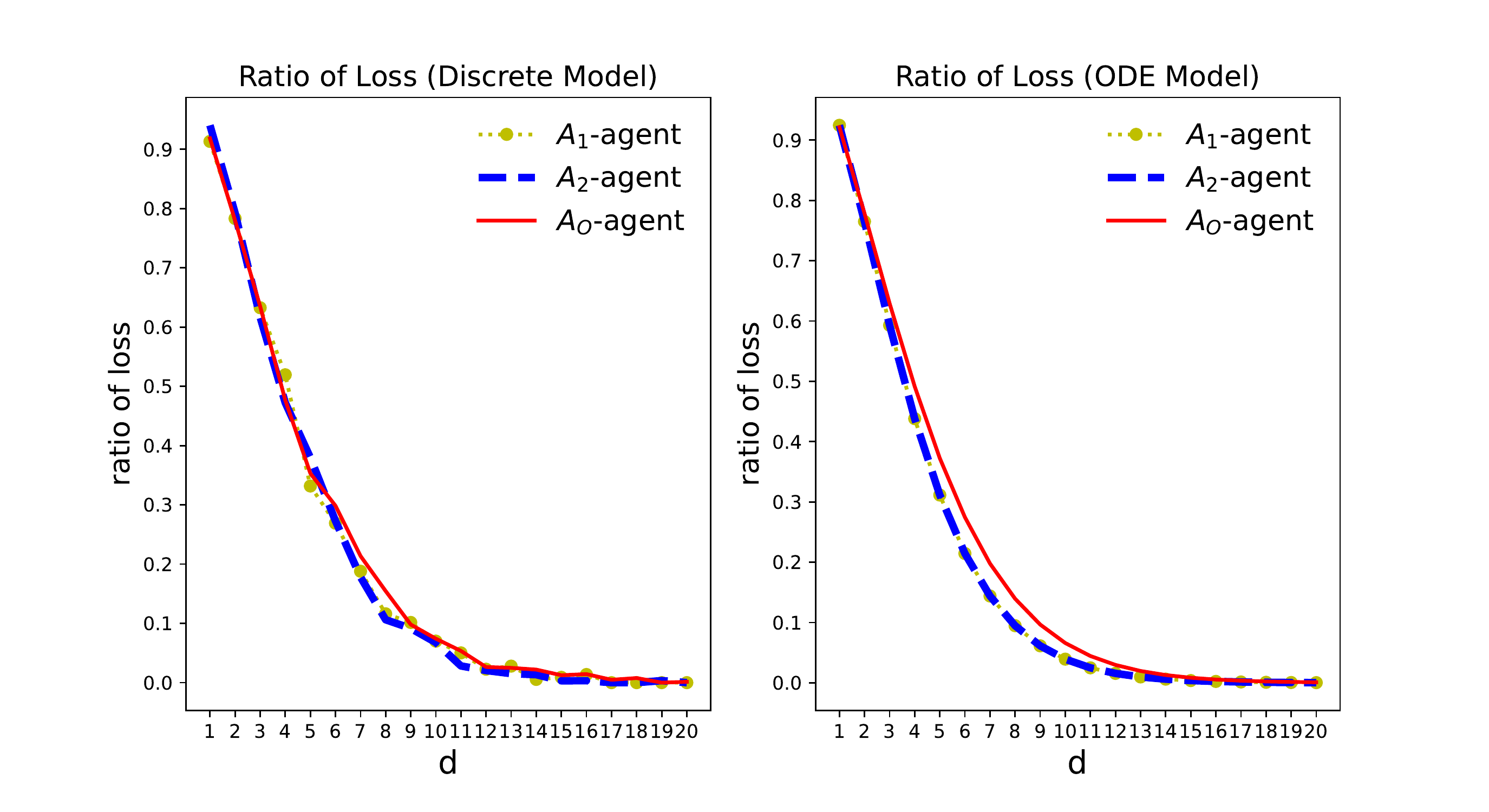}
    \vspace{-0.2in}
    \caption{The ratio of the loss of the Patient Algorithm to the loss of the Greedy Algorithm with $m=8,000,\lambda=0.2$.}
    \label{Ratio_of_Two_Loss}
\end{figure}

Figure \ref{Ratio_of_Two_Loss} shows a comparison of the performance of the Patient and Greedy Algorithms in terms of their relative loss. The figure clearly demonstrates a decrease in the ratio of the loss of the Patient Algorithm to the loss of the Greedy Algorithm as the density parameter $d$ increases.  This observation aligns with our theoretical predictions of Theorems \ref{main_result_greedy} and \ref{main_result_patient}, which indicate that the loss of the Patient Algorithm decays exponentially as $d$ increases, while the loss of the Greedy Algorithm decays linearly. The results of Figure \ref{Ratio_of_Two_Loss} thus confirm our theoretical conclusion that, for increasing values of $d$, the Patient Algorithm will outperform the Greedy Algorithm in terms of loss minimization.

\subsection{Sensitivity Analysis}

\begin{figure}[ht!]
    \centering
    \includegraphics[width=0.8\textwidth]{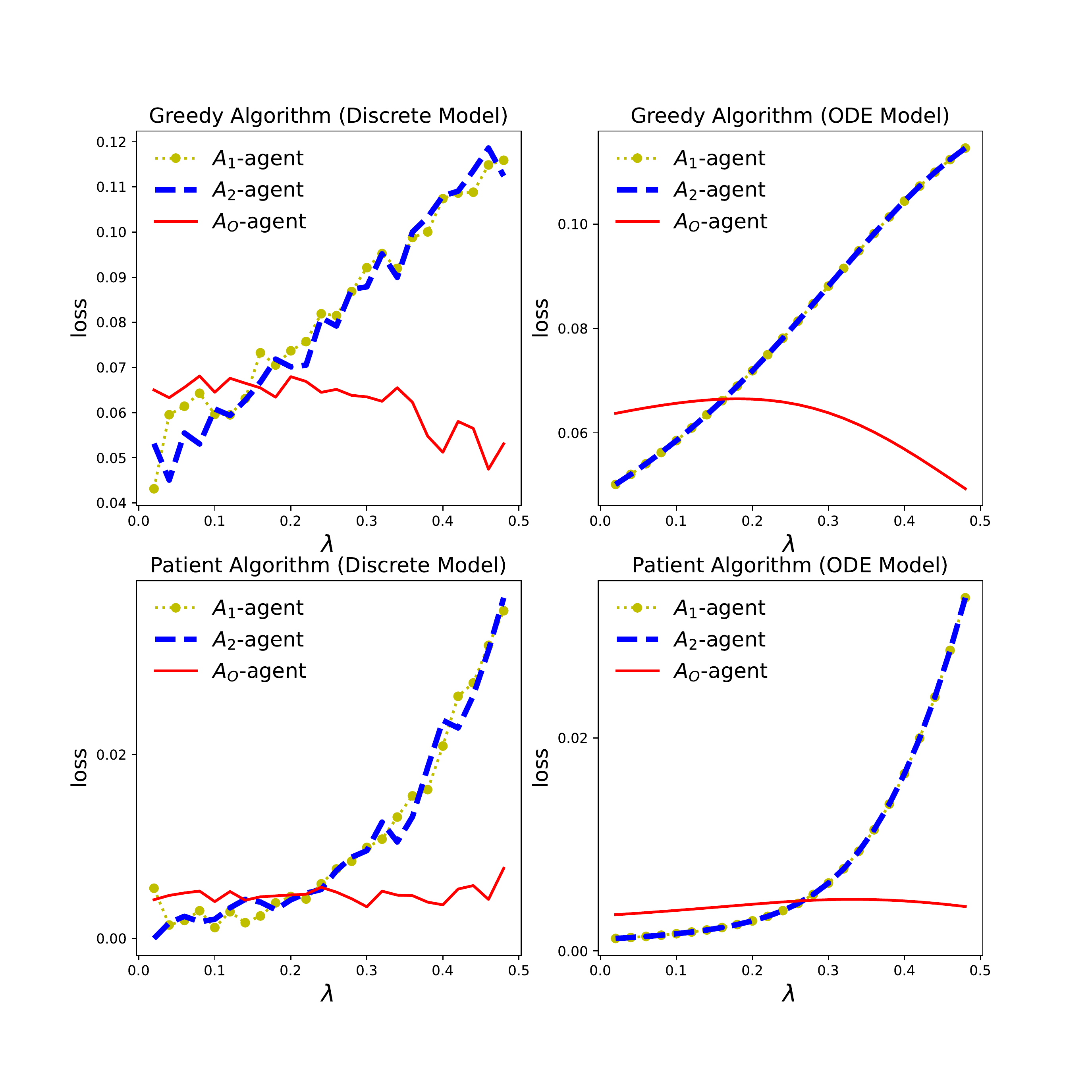}
        \vspace{-0.5in}
    \caption{The loss functions with $m=8,000,d=10$.}
    \label{lamchange}
\end{figure}

\noindent
To gain further understanding into the impact of the model parameters on our results, we perform a sensitivity analysis using the simulation setup outlined in Section \ref{sec:discretevsconti}. This analysis provides deeper insights into the interplay between different parameters and their effect on our results, offering a comprehensive view of the robustness and stability of our findings.

First, we carry out a sensitivity analysis of the impact of varying the arrival rate of different agent types, represented by $\lambda$ in Eq.~\eqref{eqn:occuragents}. Here we fix $m=8,000$ and $d=10$. The results in Figure \ref{lamchange} shows that an increase in $\lambda$ leads to a corresponding increase in the loss function for hard-to-match agents. This suggests that a higher arrival rate of these agents leads to increased difficulties in the dynamic matching process. It also demonstrates the accuracy of the continuous ODE models in providing predictions for the discrete dynamic models.

Second, we observe a critical point in the relationship between $\lambda$ and the performance of the algorithms. Figure \ref{lamchange} illustrates the loss function for both hard-to-match and easy-to-match agents. The results reveal that, when using the Patient Algorithm with $p=2$, the loss function for hard-to-match agents remains relatively small for $\lambda\leq 1/4$, but experiences a sharp increase when $\lambda>1/4$. In contrast, the loss function for easy-to-match agents remains relatively stable across different values of $\lambda$. 
This observation aligns with the theoretical phase transition phenomenon at $\lambda p = 1/2$ of the Patient Algorithm as established in Theorem \ref{main_result_patient}.

\begin{figure}[t!]
    \centering
    \includegraphics[width=0.8\textwidth]{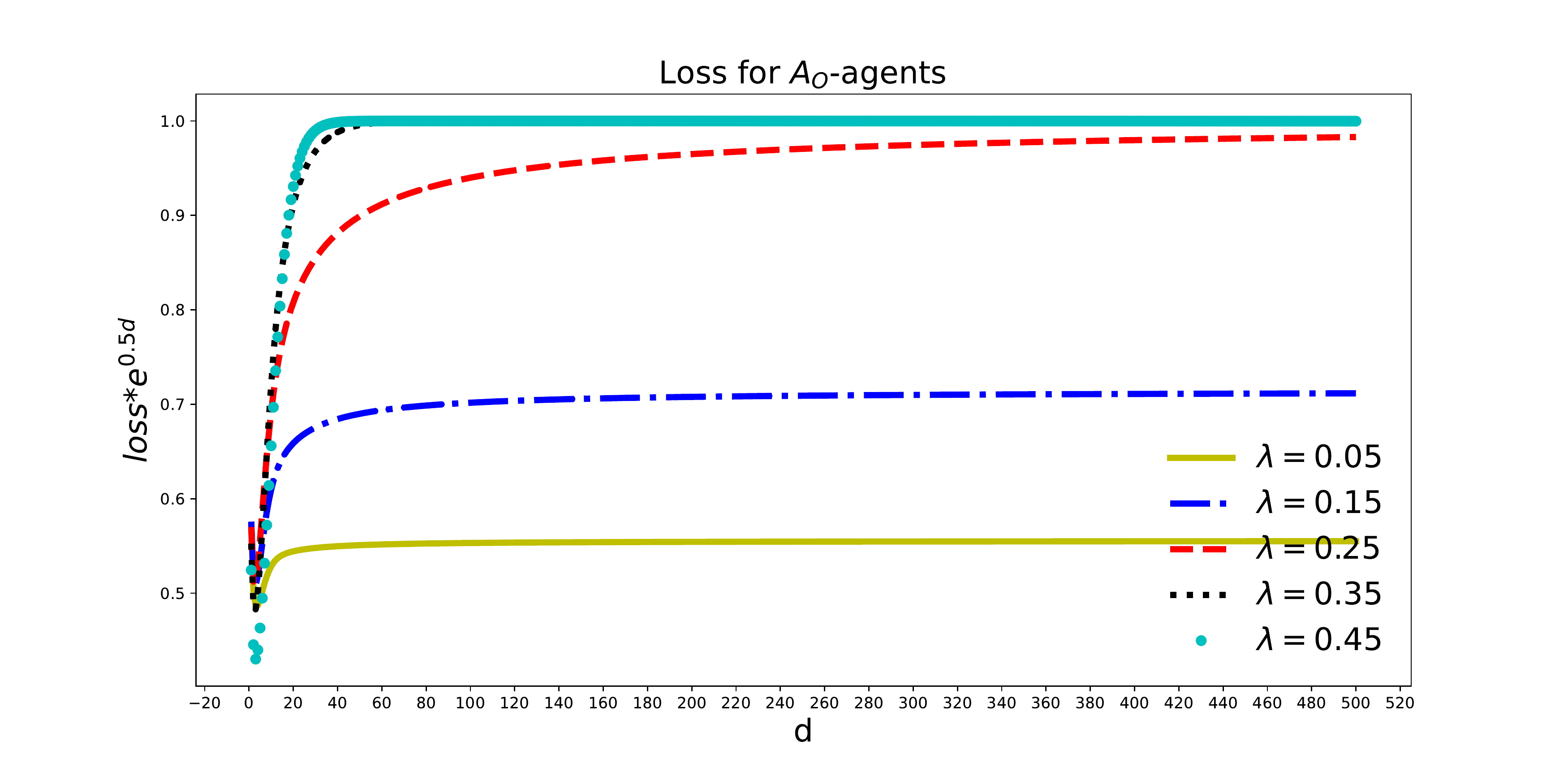}
     \vspace{-0.2in}
    \caption{The loss function of $A_0$-agent multiplied by $e^{d/2}$ with $m=8,000$.}
    \label{lossforO}
\end{figure}

Third,  we evaluate the impact of varying the density parameter $d$, as defined in Eq.\eqref{eqn:densityd}, on the matching performance of easy-to-match agents. Figure \ref{lossforO} demonstrate that, regardless of the value of $\lambda$, the ratio of $e^{d/2} \cdot L^{\text{patient}}(A_0)$ approaches a constant value as $d$ increases. This observation supports our theoretical finding in Theorem \ref{main_result_patient} that, in the asymptotic regime where $m$ approaches infinity, $T$ approaches infinity, and $d$ is not small, the loss for easy-to-match agents is proportional to $e^{\left[-\frac{1}{2}+o(1)\right]d}$, regardless of the value of $\lambda$.


\subsection{Real Data Analysis}
We conduct an empirical study using data from the Organ Procurement and Transplantation Network (OPTN) as of July, 2022 provided by the United Network for Organ Sharing (UNOS).  This dataset contains information on 1,097,058 patient-donor pairs seeking kidney transplants. We analyze the data using the Greedy and Patient Algorithms to gain insights into the success rates of kidney exchanges. Our aim is to provide a thorough examination of the real-world dynamics of these transplants, which can inform future research and policy-making in the field of organ transplantation.

In our study, we classify pairs into two categories: easy-to-match and hard-to-match agents. Pairs whose blood types are $X$-$O$  are considered easy-to-match agents, where $X$ is the patient's blood type,  $O$ is the donor's blood type, and $X \in \{A, B, AB, O\}$. Moreover,   Pairs whose blood types are $AB$-$Y$  are also considered easy-to-match agents, where $AB$ is the patient's blood type, $Y$ is the donor's blood type, and $Y \in \{A, B, AB, O\}$. This is because a patient with blood type $AB$ can receive an organ from any donor blood type, and a donor with blood type $O$ can donate to any patient blood type. The remaining types of agents are considered hard-to-match agents. For pairs to be matched, they must first satisfy blood type compatibility, and then need to satisfy that the donor's antigens must be the same as the patient's antigens. We preprocess the dataset and obtain samples with complete information on patient and donor blood types and antigens. Then we generate pairs according to a Poisson process with the rate $m=2,000$, and set a varying probability $\lambda\in(0,1)$ for them to be easy-to-match agents. The maximum waiting time for a pair to become critical is determined by an exponential distribution with a mean of $1$. 
We then run both the Greedy and Patient Algorithms in Algorithms \ref{alg:greedyalg} and \ref{alg:patientalg}, respectively, on the generated pairs and record the loss with different values of $\lambda$. 

\begin{figure}[t!]
    \centering
    \includegraphics[width=\textwidth]{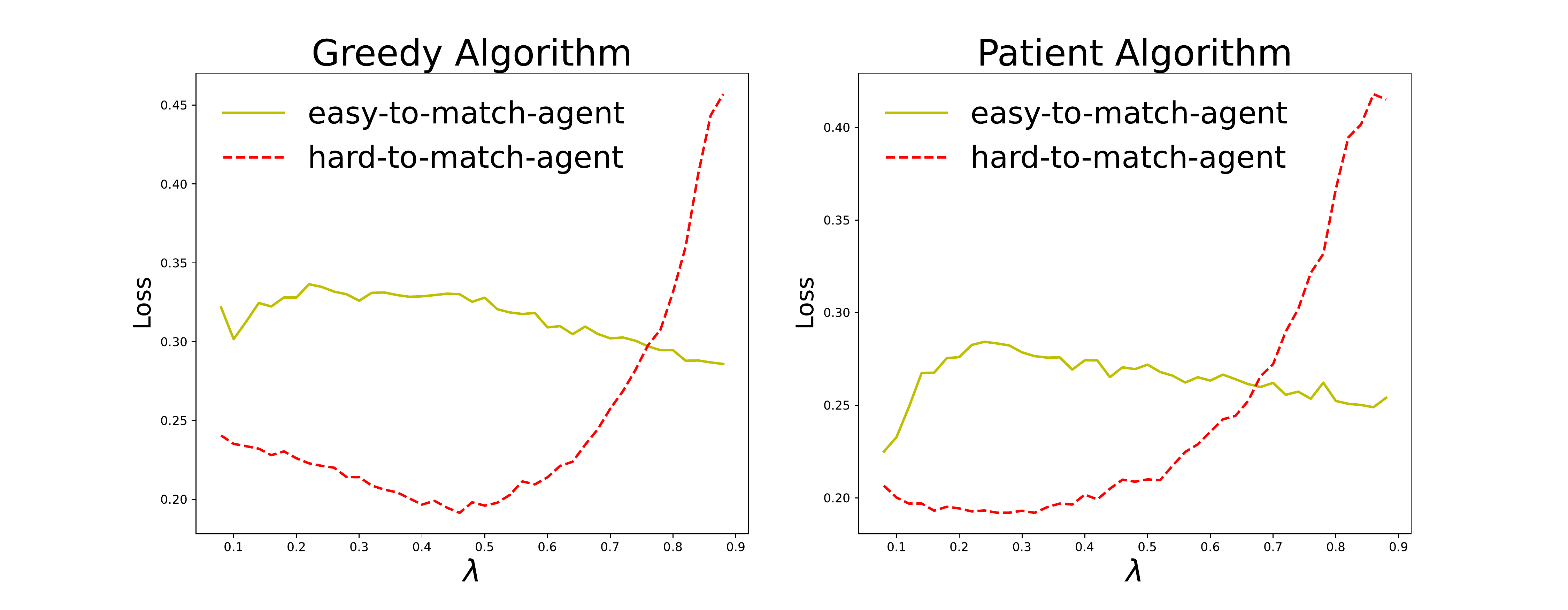}
    \vspace{-0.2in}
    \caption{The empirical loss with real-world UNOS data and $m=2,000$.}
    \label{lossforempirical}
\end{figure}

Figure~\ref{lossforempirical} indicates that varying the proportion of easy-to-match agents in the market, represented by $\lambda$, has little impact on the loss for easy-to-match agents, but causes a significant change in the loss for hard-to-match agents. This observation is consistent with Theorem \ref{main_result_patient}. The loss for easy-to-match agents remains at $e^{\left[-\frac{1}{2}+o(1)\right]d}$, regardless of the value of $\lambda$.
However, for hard-to-match agents, the situation is different. 
When $\lambda p \leq 1/2$ and $p=1$, the loss function is expected to change slowly with changes in $\lambda$, remaining at $e^{\left[-\frac{1}{2}+o(1)\right]d}$. On the other hand, when $\lambda p > 1/2$, the loss function for hard-to-match agents is a rapidly increasing function of $\lambda$, described by $e^{-\left[1-\frac{1}{2p}-(p-1)\lambda+o(1)\right]d}$.
Furthermore, it is observed that the Patient Algorithm yields a smaller loss compared to the Greedy Algorithm, further validating the theoretical predictions in Section \ref{sec:compgreedpat}.

\section{Related Works}
\label{sec:relatedworks}
In this section, we provide a brief review of recent advancements in the field of dynamic matching, encompassing the domains of kidney exchange, maximum matching, and multi-agent learning. Our aim is to contextualize this paper within the wider scope of the dynamic matching literature, highlighting the key contributions and results that are most pertinent to our approach.

\paragraph{Dynamic Matching} The dynamic matching model has a wide range of applications, including real-time ride sharing \citep{ozkan2020dynamic, feng2021we}, house allocation \citep{RePEc:jrp:jrpwrp:2009-075},  durable object assignment \citep{Blochoptimal}, online assortment optimization \citep{aouad2022online}, and market making \citep{LOERTSCHER2022105383}.
The recent literature has mainly focused on networked settings, where dynamic matching takes place between agents embedded in complex networks, and where the presence of heterogeneous agents and compatibility constraints pose new challenges for matching algorithms. 
Several mechanism designs have been proposed in the literature to address the challenges of dynamic matching \citep[see, e.g.,][]{baccara2020optimal, leshno2012dynamic}.
Studies have examined the differences between the Greedy and Patient Algorithms for dynamic matching in homogeneous random graphs \citep{akbarpour2020thickness}, prioritization in a dynamic, heterogeneous market with varying levels of hard-to-match agents and different matching technology \citep{ashlagi2019matching}, and 
the properties of greedy algorithms in dynamic networks with a constant arrival rate and fixed matching probabilities \citep{10.1145/3490486.3538323}. 
Our work differs in its definition of hard-to-match agents and matching probabilities and provides an innovative way to understand the trade-offs between the competing goals of matching agents quickly and optimally, while accounting for compatibility restrictions and stochastic arrival and departure times.

\paragraph{Kidney Exchange} The problem of kidney exchange, which involves the matching of donors and recipients for organ transplants, has received extensive attention in the literature and is a notable application area for the field of dynamic matching.
The seminar works in kidney exchange are by \citet{Roth2004,roth2005pairwise,roth2007efficient}, which proposed an algorithm for finding optimal matches in kidney exchange using integer programming. 
Subsequent work has focused on expanding the scope of kidney exchange programs. For example, \citet{rees2009nonsimultaneous} introduced an approach in which a single altruistic donor's kidney starts a chain of compatible matches involving multiple donor-recipient pairs. Another innovation is the use of non-directed donors (i.e., altruistic donors who do not have a specific recipient in mind) to initiate chains of exchanges \citep{montgomery2006domino}.
More recently, several works have focused on analyzing various models of kidney exchange, including the impact of agents entering and leaving the market stochastically and the role of waiting costs \citep[see, e.g.,][]{unver2010dynamic, ashlagi2013kidney}. 
Our work proposes a novel approach to modeling dynamic matching in kidney exchange programs by leveraging ordinary differential equation (ODE) models. This allows for a more comprehensive evaluation and comparison of different matching algorithms, specifically in the context of heterogeneous donor and recipient types. The use of ODE models provides a powerful framework that offers new insights into the complex dynamics of kidney exchange programs.

\paragraph{Maximum Matching}
Maximum matching is a fundamental problem in graph theory and combinatorial optimization, with the goal of fining a set of edges in a graph that do not share any vertices with each other. There have been many studies on this topic, both in terms of theoretical results and practical algorithms. In terms of theoretical results, the classic maximum matching problem for bipartite graphs is polynomial-time solvable by the Hungarian algorithm \citep{kuhn1955hungarian}. For general graphs, the problem is NP-hard \citep{edmonds1965paths}, but can be solved in polynomial time using the blossom algorithm \citep{edmonds1965maximum}. 
\citet{shapley1971assignment} formulated a model of two-sided matching with side payments, which is also called the maximum weighted bipartite matching \citep{karlin2017game}.
In terms of practical algorithms, there are several well-known heuristics for solving the maximum matching problem, such as the greedy algorithm \citep{galil1986efficient}, the Hopcroft-Karp algorithm \citep{hopcroft1973n}, and the push-relabel algorithm \citep{goldberg1988new}. These algorithms have been widely used in various applications.
Recently, there has been growing interest in the online version of maximum matching and its generalizations, due to the important new application domain of Internet advertising \citet{mehta2013online}. For example,
\cite{aouad2020dynamic} studied online matching problems on edge-weighted graphs for the goals of cost minimization and reward maximization.
Our problem involves finding the maximum matching while accounting for compatibility restrictions and dynamic, unpredictable changes in the availability of agents. This scenario poses significant challenges in the analysis of the matching process.

\paragraph{Matching While Learning} 
A recent thread of research focuses on the study of matching markets with preference learning, where participants in the market have preferences over each other, and agents learn optimal strategies from past experiences. For instance, \citet{dai2021learning, dai2021learningmultistage} have studied the problem of agents learning optimal strategies in decentralized matching. \citet{dai2022incentive} considers the learning of optimal strategies under  incentive constraints. 
\citet{li2023double} studies the learning and matching under complementary  preferences. 
The dynamic matching problem addressed in this paper is similar in nature to the problem of learning the optimal strategies 
of the Greedy or Patient algorithms, where preferences are represented by mutual compatibility. Moreover, our work differs from \cite{10.1145/3465456.3467588} which examines a one-sided matching problem of allocating objects to agents with private types. In contrast, our paper focuses on a two-sided matching problem where agents match with each other, and the compatibility between agents is characterized by their types.

\section{Conclusion and Discussion}
\label{sec:conclusion}

This paper presents a novel approach to dynamic matching in heterogeneous networks through the use of ordinary differential equation (ODE) models. The ODE models facilitate the analysis of the trade-off between agents' waiting times and the percentage of matched agents in these networks, where agents are subject to compatibility constraints and their arrival and departure times are uncertain. Our results show that while the Greedy Algorithm reduces waiting times, it may lead to a significantly worse loss of social welfare, while the Patient Algorithm, with its focus on maximizing compatibility, significantly extends waiting times. Our simulations, sensitivity analysis, and real data analysis all demonstrate the practical applicability of our theory and provide important insights for the design of real-world dynamic matching systems.

There are many interesting directions for future studies. First, it is  crucial to incorporate the cost of waiting into the loss function, considering its significant impact on both the patient's quality of life and the overall financial savings for society \citep{held2016cost}. 
Second, it is of interest to bridge the gap between the discrete dynamic model and the continuous ODE model in the finite-time regime. One approach to do so is by leveraging the Lyapunov function method, as used in previous works such as \cite{Anderson2017efficient} and \cite{ashlagi2018matching}. A major challenge in this method is constructing appropriate energy functions. A potential solution is to first establish energy functions for the ODEs in the continuous models and then use these as a foundation to construct energy functions for the discrete models.

\acks{X. D. was partially supported by UCLA CCPR under NIH grant NICHD P2C-HD041022. We would like to thank Mark S. Handcock for helpful discussions about parts of this paper.}

\vskip 0.2in
\bibliography{ODE-matching}

\begin{thebibliography}{40}
\providecommand{\natexlab}[1]{#1}
\providecommand{\url}[1]{\texttt{#1}}
\expandafter\ifx\csname urlstyle\endcsname\relax
  \providecommand{\doi}[1]{doi: #1}\else
  \providecommand{\doi}{doi: \begingroup \urlstyle{rm}\Url}\fi

\bibitem[Akbarpour et~al.(2020)Akbarpour, Li, and
  Gharan]{akbarpour2020thickness}
Mohammad Akbarpour, Shengwu Li, and Shayan~Oveis Gharan.
\newblock Thickness and information in dynamic matching markets.
\newblock \emph{Journal of Political Economy}, 128\penalty0 (3):\penalty0
  783--815, 2020.

\bibitem[Anderson et~al.(2017)Anderson, Ashlagi, Gamarnik, and
  Kanoria]{Anderson2017efficient}
Ross Anderson, Itai Ashlagi, David Gamarnik, and Yash Kanoria.
\newblock Efficient dynamic barter exchange.
\newblock \emph{Operations Research}, 65, 2017.

\bibitem[Andronov et~al.(2013)Andronov, Vitt, and Khaikin]{andronov2013theory}
Aleksandr~Aleksandrovich Andronov, Aleksandr~Adol'fovich Vitt, and
  Semen~Emmanuilovich Khaikin.
\newblock \emph{Theory of Oscillators}, volume~4.
\newblock Elsevier, 2013.

\bibitem[Aouad and Saban(2022)]{aouad2022online}
Ali Aouad and Daniela Saban.
\newblock Online assortment optimization for two-sided matching platforms.
\newblock \emph{Management Science}, 2022.

\bibitem[Aouad and Sarita{\c{c}}(2020)]{aouad2020dynamic}
Ali Aouad and {\"O}mer Sarita{\c{c}}.
\newblock Dynamic stochastic matching under limited time.
\newblock In \emph{Proceedings of the 21st ACM Conference on Economics and
  Computation}, 2020.

\bibitem[Ashlagi et~al.(2013)Ashlagi, Jaillet, and Manshadi]{ashlagi2013kidney}
Itai Ashlagi, Patrick Jaillet, and Vahideh~H Manshadi.
\newblock Kidney exchange in dynamic sparse heterogenous pools.
\newblock \emph{EC '13}, 2013.

\bibitem[Ashlagi et~al.(2019)Ashlagi, Burq, Jaillet, and
  Manshadi]{ashlagi2019matching}
Itai Ashlagi, Maximilien Burq, Patrick Jaillet, and Vahideh Manshadi.
\newblock On matching and thickness in heterogeneous dynamic markets.
\newblock \emph{Operations Research}, 67\penalty0 (4):\penalty0 927--949, 2019.

\bibitem[Ashlagi et~al.(2021)Ashlagi, Monachou, and
  Nikzad]{10.1145/3465456.3467588}
Itai Ashlagi, Faidra Monachou, and Afshin Nikzad.
\newblock Optimal dynamic allocation: Simplicity through information design.
\newblock In \emph{Proceedings of the 22nd ACM Conference on Economics and
  Computation}, EC '21, 2021.

\bibitem[Ashlagi et~al.(2022)Ashlagi, Nikzad, and Strack]{ashlagi2018matching}
Itai Ashlagi, Afshin Nikzad, and Philipp Strack.
\newblock Matching in dynamic imbalanced markets.
\newblock \emph{Review of Economic Studies}, 2022.

\bibitem[Baccara et~al.(2020)Baccara, Lee, and Yariv]{baccara2020optimal}
Mariagiovanna Baccara, SangMok Lee, and Leeat Yariv.
\newblock Optimal dynamic matching.
\newblock \emph{Theoretical Economics}, 15\penalty0 (3):\penalty0 1221--1278,
  2020.

\bibitem[Bloch and Houy(2009)]{Blochoptimal}
Francis Bloch and Nicolas Houy.
\newblock Optimal assignment of durable objects to successive agents.
\newblock \emph{Economic Theory}, 51, 11 2009.

\bibitem[Ciesielski(2012)]{ciesielski2012poincare}
Krzysztof Ciesielski.
\newblock The poincar{\'e}-bendixson theorem: from poincar{\'e} to the xxist
  century.
\newblock \emph{Central European Journal of Mathematics}, 10\penalty0
  (6):\penalty0 2110--2128, 2012.

\bibitem[Coddington and Levinson(1955)]{coddington1955theory}
Earl~A Coddington and Norman Levinson.
\newblock \emph{Theory of Ordinary Differential Equations}.
\newblock Tata McGraw-Hill Education, 1955.

\bibitem[Dai and Jordan(2021{\natexlab{a}})]{dai2021learning}
Xiaowu Dai and Michael~I Jordan.
\newblock Learning strategies in decentralized matching markets under uncertain
  preferences.
\newblock \emph{Journal of Machine Learning Research}, 22\penalty0
  (260):\penalty0 1--50, 2021{\natexlab{a}}.

\bibitem[Dai and Jordan(2021{\natexlab{b}})]{dai2021learningmultistage}
Xiaowu Dai and Michael~I Jordan.
\newblock Learning in multi-stage decentralized matching markets.
\newblock \emph{Advances in Neural Information Processing Systems},
  34:\penalty0 12798--12809, 2021{\natexlab{b}}.

\bibitem[Dai et~al.(2022)Dai, Qi, and Jordan]{dai2022incentive}
Xiaowu Dai, Yuan Qi, and Michael~I Jordan.
\newblock Incentive-aware recommender systems in two-sided markets.
\newblock \emph{arXiv preprint arXiv:2211.15381}, 2022.

\bibitem[Edmonds(1965{\natexlab{a}})]{edmonds1965maximum}
Jack Edmonds.
\newblock Maximum matching and a polyhedron with 0, 1-vertices.
\newblock \emph{Journal of Research of the National Bureau of Standards B},
  69\penalty0 (125-130):\penalty0 55--56, 1965{\natexlab{a}}.

\bibitem[Edmonds(1965{\natexlab{b}})]{edmonds1965paths}
Jack Edmonds.
\newblock Paths, trees, and flowers.
\newblock \emph{Canadian Journal of Mathematics}, 17:\penalty0 449--467,
  1965{\natexlab{b}}.

\bibitem[Feng et~al.(2021)Feng, Kong, and Wang]{feng2021we}
Guiyun Feng, Guangwen Kong, and Zizhuo Wang.
\newblock We are on the way: Analysis of on-demand ride-hailing systems.
\newblock \emph{Manufacturing \& Service Operations Management}, 23\penalty0
  (5):\penalty0 1237--1256, 2021.

\bibitem[Galil(1986)]{galil1986efficient}
Zvi Galil.
\newblock Efficient algorithms for finding maximum matching in graphs.
\newblock \emph{ACM Computing Surveys (CSUR)}, 18\penalty0 (1):\penalty0
  23--38, 1986.

\bibitem[Goldberg and Tarjan(1988)]{goldberg1988new}
Andrew~V Goldberg and Robert~E Tarjan.
\newblock A new approach to the maximum-flow problem.
\newblock \emph{Journal of the ACM (JACM)}, 35\penalty0 (4):\penalty0 921--940,
  1988.

\bibitem[Held et~al.(2016)Held, McCormick, Ojo, and Roberts]{held2016cost}
Philip~J Held, Frank McCormick, Akinlolu Ojo, and John~P Roberts.
\newblock A cost-benefit analysis of government compensation of kidney donors.
\newblock \emph{American Journal of Transplantation}, 16\penalty0 (3):\penalty0
  877--885, 2016.

\bibitem[Hopcroft and Karp(1973)]{hopcroft1973n}
John~E Hopcroft and Richard~M Karp.
\newblock An $n^{5/2}$ algorithm for maximum matchings in bipartite graphs.
\newblock \emph{SIAM Journal on computing}, 2\penalty0 (4):\penalty0 225--231,
  1973.

\bibitem[Karlin and Peres(2017)]{karlin2017game}
Anna~R Karlin and Yuval Peres.
\newblock \emph{Game Theory, Alive}, volume 101.
\newblock American Mathematical Society, 2017.

\bibitem[Kerimov et~al.(2022)Kerimov, Ashlagi, and
  Gurvich]{10.1145/3490486.3538323}
S\"{u}leyman Kerimov, Itai Ashlagi, and Itai Gurvich.
\newblock On the optimality of greedy policies in dynamic matching.
\newblock In \emph{Proceedings of the 23rd ACM Conference on Economics and
  Computation}, EC '22, 2022.

\bibitem[Kuhn(1955)]{kuhn1955hungarian}
Harold~W Kuhn.
\newblock The hungarian method for the assignment problem.
\newblock \emph{Naval Research Logistics Quarterly}, 2\penalty0 (1-2):\penalty0
  83--97, 1955.

\bibitem[Kurino(2009)]{RePEc:jrp:jrpwrp:2009-075}
Morimitsu Kurino.
\newblock House allocation with overlapping agents: A dynamic mechanism design
  approach.
\newblock Jena Economic Research Papers 2009-075, Friedrich-Schiller-University
  Jena, 2009.

\bibitem[Leon-Garcia(2008)]{leon2008probability}
Alberto Leon-Garcia.
\newblock \emph{Probability, Statistics, and Random Processes for Electrical
  Engineering}.
\newblock Prentice Hall, 2008.

\bibitem[Leshno(2022)]{leshno2012dynamic}
Jacob~D Leshno.
\newblock Dynamic matching in overloaded waiting lists.
\newblock \emph{American Economic Review}, 112\penalty0 (12):\penalty0
  3876--3910, 2022.

\bibitem[Li et~al.(2023)Li, Cheng, and Dai]{li2023double}
Yuantong Li, Guang Cheng, and Xiaowu Dai.
\newblock Double matching under complementary preferences.
\newblock \emph{arXiv preprint arXiv:2301.10230}, 2023.

\bibitem[Loertscher et~al.(2022)Loertscher, Muir, and
  Taylor]{LOERTSCHER2022105383}
Simon Loertscher, Ellen~V. Muir, and Peter~G. Taylor.
\newblock Optimal market thickness.
\newblock \emph{Journal of Economic Theory}, 200:\penalty0 105383, 2022.

\bibitem[Mehta(2013)]{mehta2013online}
Aranyak Mehta.
\newblock Online matching and ad allocation.
\newblock \emph{Foundations and Trends in Theoretical Computer Science},
  8\penalty0 (4):\penalty0 265--368, 2013.

\bibitem[Montgomery et~al.(2006)Montgomery, Gentry, Marks, Warren, Hiller,
  Houp, Zachary, Melancon, Maley, Rabb, et~al.]{montgomery2006domino}
Robert~A Montgomery, Sommer~E Gentry, William~H Marks, Daniel~S Warren, Janet
  Hiller, Julie Houp, Andrea~A Zachary, J~Keith Melancon, Warren~R Maley, Hamid
  Rabb, et~al.
\newblock Domino paired kidney donation: a strategy to make best use of live
  non-directed donation.
\newblock \emph{The Lancet}, 368\penalty0 (9533):\penalty0 419--421, 2006.

\bibitem[{\"O}zkan and Ward(2020)]{ozkan2020dynamic}
Erhun {\"O}zkan and Amy~R Ward.
\newblock Dynamic matching for real-time ride sharing.
\newblock \emph{Stochastic Systems}, 10\penalty0 (1):\penalty0 29--70, 2020.

\bibitem[Rees et~al.(2009)Rees, Kopke, Pelletier, Segev, Rutter, Fabrega,
  Rogers, Pankewycz, Hiller, Roth, et~al.]{rees2009nonsimultaneous}
Michael~A Rees, Jonathan~E Kopke, Ronald~P Pelletier, Dorry~L Segev, Matthew~E
  Rutter, Alfredo~J Fabrega, Jeffrey Rogers, Oleh~G Pankewycz, Janet Hiller,
  Alvin~E Roth, et~al.
\newblock A nonsimultaneous, extended, altruistic-donor chain.
\newblock \emph{New England Journal of Medicine}, 360\penalty0 (11):\penalty0
  1096--1101, 2009.

\bibitem[Roth et~al.(2004)Roth, Sönmez, Unver, and Ünver]{Roth2004}
Alvin Roth, Tayfun Sönmez, Utku Unver, and M.~Ünver.
\newblock Kidney exchange.
\newblock \emph{The Quarterly Journal of Economics}, 119:\penalty0 457--488,
  2004.

\bibitem[Roth et~al.(2005)Roth, S{\"o}nmez, and {\"U}nver]{roth2005pairwise}
Alvin~E Roth, Tayfun S{\"o}nmez, and M~Utku {\"U}nver.
\newblock Pairwise kidney exchange.
\newblock \emph{Journal of Economic theory}, 125\penalty0 (2):\penalty0
  151--188, 2005.

\bibitem[Roth et~al.(2007)Roth, S{\"o}nmez, and {\"U}nver]{roth2007efficient}
Alvin~E Roth, Tayfun S{\"o}nmez, and M~Utku {\"U}nver.
\newblock Efficient kidney exchange: Coincidence of wants in markets with
  compatibility-based preferences.
\newblock \emph{American Economic Review}, 97\penalty0 (3):\penalty0 828--851,
  2007.

\bibitem[Shapley and Shubik(1971)]{shapley1971assignment}
Lloyd~S Shapley and Martin Shubik.
\newblock The assignment game i: The core.
\newblock \emph{International Journal of Game Theory}, 1\penalty0 (1):\penalty0
  111--130, 1971.

\bibitem[{\"U}nver(2010)]{unver2010dynamic}
M.~Utku {\"U}nver.
\newblock Dynamic kidney exchange.
\newblock \emph{The Review of Economic Studies}, 77\penalty0 (1):\penalty0
  372--414, 2010.

\end{thebibliography}


\newpage

\appendix

\section{Proofs}
\subsection{Proof of Proposition \ref{thm:matchwithwho}}
\begin{proof}
We analyze each of the four terms in Proposition \ref{thm:matchwithwho} separately. 
First, we consider the term $\pi^*_t(A_0,A_0)$, where the planner is trying to match an $A_0$-type agent of interest with another $A_0$-type agent who is in the pool.
Since the hard-to-match agents are prioritized over easy-to-match agents for matching in both Greedy and Patient Algorithms, the matching of $\{A_0,A_0\}$ happens only when the existing hard-to-match agents are incompatible with the $A_0$-type agent of interest.
That is, $\pi_t^*(A_0,A_0) = \P(\BB_1\cap\BB_2)$, where the events $\BB_1$ and $\BB_2$ are defined by,
\begin{equation*}
\begin{aligned}
\BB_1 & = \Big\{\text{all $A_j$-type agents in the pool are incompatible with the agent of interest}, \forall j\geq 1\Big\},\\
\BB_2 & = \Big\{\text{at least one $A_0$-type agent in the pool is compatible with the agent of interest}\Big\}.
\end{aligned}
\end{equation*}
Note that, $\P(\BB_1) = (1-\alpha)^{\sum_{k=1}^{p} |\SS_t(A_{k})|}$ and $\P(\BB_2) = 1-(1-\alpha)^{|\SS_t(A_0)|}$.
By the independence of the events $\BB_1$ and $\BB_2$, we have
\begin{equation*}
    \pi_t^*(A_0,A_0) = \P(\BB_1)\P(\BB_2) =(1-\alpha)^{\sum_{k=1}^{p} |\SS_t(A_{k})|} \left[1-(1-\alpha)^{|\SS_t(A_0)|}\right].
\end{equation*}

Second, we consider the term $\pi^*_t(A_0,A_j)$, where the planner is trying to match an $A_0$-type agent of interest with an $A_j$-type agent who is in the pool, for $j\geq 1$. 
We can decompose this probability by,
\begin{equation*}
\begin{aligned}
 \pi^*_t(A_0,A_j) & = \sum_{k=0}^{p-1}\sum_{\{j_1,\ldots,j_k\}\subseteq [p]\setminus\{j\}}\P\left(\BB_3\cap\BB_4\right)\\
 & =  \sum_{k=0}^{p-1}\sum_{\{j_1,\ldots,j_k\}\subseteq [p]\setminus\{j\}}\P\left(\BB_3\right)\P\left(\BB_4|\BB_3\right),   
\end{aligned}
\end{equation*}
where the events $\BB_3$ and $\BB_4$ depend on $\{j,j_1,\ldots,j_k\}$ which are defined by,
\begin{equation}
\label{eqn:calpi0j}
\begin{aligned}
\BB_3 & = \Big\{\text{at least one agent in the pool whose type $\in\{A_j,A_{j_1},\ldots,A_{j_k}\}$} \\
&\quad\quad\quad\quad\quad\quad\quad\quad\quad\quad\quad\quad\quad \text{is compatible with the agent of interest}\Big\},\\
\BB_4 & = \Big\{\text{match the agent of interest with an $A_j$-type agent}\Big\}.
\end{aligned}
\end{equation}
We have that,
\begin{equation*}
\begin{aligned}
\P(\BB_3) & = \left[1-(1-\alpha)^{|\SS_t(A_j)|}\right]\Pi_{s=1}^k\left[1-(1-\alpha)^{|\SS_t(A_{j_s})|}\right]\Pi_{l\not\in\{j,j_1,\ldots,j_k\}}(1-\alpha)^{|\SS_t(A_{l})|},\\
\P(\BB_4) & = \frac{1}{k+1}.
\end{aligned}
\end{equation*}
Plugging these terms to Eq.~\eqref{eqn:calpi0j}, we obtain the desired result for $\pi^*_t(A_0,A_j)$.

Third, we consider the term $\pi^*_t(A_j,A_0)$. Similarly, we can decompose it as $\pi_t^*(A_0,A_0) = \P(\BB_5\cap\BB_6)$, where
\begin{equation*}
\begin{aligned}
\BB_5 & = \Big\{\text{all $A_k$-type agents in the pool are incompatible with the agent of interest}, \forall k\geq 1\Big\},\\
\BB_6 & = \Big\{\text{at least one $A_0$-type agent in the pool is compatible with the agent of interest}\Big\}.
\end{aligned}
\end{equation*}
By the compatibility model \eqref{eqn:connect}, an $A_j$-type agent is only compatible with an $A_j$-type agent or an $A_0$-type agent. Hence we have $\P(\BB_5) = (1-\alpha)^{|\SS_t(A_j)|}$, and $\P(\BB_6) = 1-(1-\alpha)^{|\SS_t(A_0)|}$. By the independence of the events $\BB_5$ and $\BB_6$, we have
\begin{equation*}
    \pi_t^*(A_j,A_0) = \P(\BB_5)\P(\BB_6) =(1-\alpha)^{|\SS_t(A_j)|}\left[1-(1-\alpha)^{|\SS_t(A_0)|}\right].
\end{equation*}

Finally, we consider the term $\pi^*_t(A_j,A_j)$. We have that $\pi^*_t(A_j,A_j) = \P(\BB_7)$, where
\begin{equation*}
    \BB_7 = \Big\{\text{at least one $A_j$-type agent in the pool is compatible with the agent of interest}\Big\}.
\end{equation*}
Since $\P(\BB_7) = (1-\alpha)^{|\SS_t(A_j)|}$, we conclude that $\pi^*_t(A_j,A_j) = (1-\alpha)^{|\SS_t(A_j)|}$. This completes the proof of the proposition. 
\end{proof}

\subsection{Proof of Theorem \ref{thm:Greedy_limit}}
We start with introducing some notations. Assume that at time interval $[t,t+\Delta t]$, there are $\CC$ agents that have already stayed in the market before time $t$ become critical, $\RR$ new agents arrive at the market. Then for any $A_k$-type agents, $k=0,1,\ldots,p$, we have,
\begin{equation}
\label{condition_probability}
\begin{aligned}
    &\quad  \E [|\SS_{t+\Delta t}(A_k)|]- |\SS_t(A_k)| \\
    & \quad = \ \E[\E[|\SS_{t+\Delta t}(A_k)|  ~|~ \CC,\RR]]-|\SS_t(A_k)|\\
    & \quad = \ \underbrace{\P(\CC=0,\RR=0)\cdot\left\{\E[|\SS_{t+\Delta t}(A_k)|  ~|~ \CC=0,\RR=0] -|\SS_t(A_k)|\right\}}_{:=~(a)}\\
    &\quad\  + \ \underbrace{\P(\CC=0,\RR=1)\cdot\left\{\E[|\SS_{t+\Delta t}(A_k)|~|~\CC=0,\RR=1]-|\SS_t(A_k)|\right\}}_{:=~(b)}\\
    & \quad\ + \  \underbrace{\P(\CC=1,\RR=0)\cdot\left\{\E[|\SS_{t+\Delta t}(A_k)|  ~|~ \CC=1,\RR=0]-|\SS_t(A_k)|\right\}}_{:=~(c)}\\
    &\quad\  + \ \underbrace{\sum_{u\geq 1,v \geq 1} \P(\CC=u,\RR=v) \cdot\left\{\E[|\SS_{t+\Delta t}(A_k)|  ~|~ \CC=u,\RR=v]-|\SS_t(A_k)|\right\}}_{:=~(d)}.
\end{aligned}
\end{equation}
In the following, we derive the ODE for  $A_j$-type agents with $j\geq 1$, and then use a similar argument to derive the ODE for $A_0$-type agents.

\subsubsection{ODE for hard-to-match agents}
\label{sec:pfodeaj1}
\begin{proof}
First, we consider $A_j$-type agents with $j\geq 1$, where
we analyze the terms $(a)$-$(d)$ in Eq.~\eqref{condition_probability} separately. 
\paragraph{The term (a):} 
This is a simple case with no agent becoming critical or arriving at the market during $[t,t+\Delta t]$. Hence
\begin{equation}
\label{eqn:k0r0}
\begin{aligned}
  \P(\CC=0,\RR=0)\cdot\left\{\E[|\SS_{t+\Delta t}(A_j)| ~|~ \CC=0,\RR=0]-|\SS_t(A_j)|\right\}   =0.
\end{aligned}
\end{equation}

\paragraph{The term (b):} In this case, there is no agent becoming critical and one agent arriving at the market during $[t,t+\Delta t]$. 
By definition, the Greedy Algorithm will immediately try to match the new agent with existing agents in the pool. We consider five scenarios of the new agent arriving at the market:
\begin{itemize}
    \item First, if the new agent is $A_k$-type where $k\neq j$, then by Eq.~\eqref{eqn:connect}, it is incompatible with any $A_j$-type agent, and so does not change the number of $A_j$-type agents in the pool. 
    \item Second, if the new agent is $A_j$-type, then with probability $(1-\alpha)^{|\SS_t(A_j)|+|\SS_{t}(A_0)|}$, it is incompatible with the existing $A_j$- and $A_0$-type agents, and so increases the number of $A_j$-type agents in the pool by $1$.
    \item Third, if the new agent is $A_j$-type,  with probability $(1-\alpha)^{|\SS_t(A_j)|}\left[1-(1-\alpha)^{|\SS_t(A_0)|}\right]$, it is incompatible with any existing $A_j$-type agent, but is compatible with an $A_0$-type agent in the pool. Hence it does not change the number of $A_j$-type agents in the pool.
    \item  Fourth, if the new agent is $A_j$-type, then with probability $1-(1-\alpha)^{|\SS_t(A_j)|}$, it is compatible with an existing $A_j$-type agent in the pool, and so decreases the number of $A_j$-type agents in the pool by $1$.
    \item  Fifth, if the new agent is $A_0$-type, then with probability $\pi^*_t(A_0,A_j)$, it is compatible with an existing $A_j$-type agent in the pool, and so decreases the number of $A_j$-type agents in the pool by $1$, where $\pi^*_t(A_0,A_j)$ is defined in Proposition \ref{thm:matchwithwho}.
\end{itemize}

Since we assume in Section \ref{sec:agentsinnetwork} that each agent becomes critical according to a Poisson process at rate 1. This implies that, if an agent $a$ is not critical at time $t$, then she becomes critical at some time between $t+X$, where $X$ is an exponential random variable with mean $1$, where 
\begin{equation}
\label{eqn:expdis1}
\P(X\leq \Delta t) = 1-e^{-\Delta t}.
\end{equation}
Moreover, if an agent $b$ is not critical at time $t$, then  the probability that $b$ is not critical during $[t,t+\Delta t]$ is, 
\begin{equation}
\label{eqn:expdis2}
\P(X> \Delta t) = e^{-\Delta t}.
\end{equation}

Putting together the above five scenarios, we have
\begin{equation}
\label{eqn:k0r1}
\begin{aligned}
    & \P(\CC=0,\RR=1)\cdot\left\{\E[|\SS_{t+\Delta t}(A_j)|~|~\CC=0,\RR=1]-|\SS_t(A_j)|\right\}\\
    &=e^{-|\SS_t|\Delta t}e^{-m \Delta t}m\Delta t
    \Biggl\{\sum_{1\leq k \leq p, k\neq j}\lambda \left[|\SS_t(A_j)|-|\SS_t(A_j)|\right] \\
    & \quad\quad\quad\quad\quad\quad\quad\quad\quad\quad  +\lambda\left[(1-\alpha)^{|\SS_t(A_j)|+|\SS_{t}(A_0)|}\right] \left[|\SS_t(A_j)|+1-|\SS_t(A_j)|\right] \\
    & \quad\quad\quad\quad\quad\quad\quad\quad\quad\quad  +  \lambda(1-\alpha)^{|\SS_t(A_j)|}\left[1-(1-\alpha)^{|\SS_t(A_0)|}\right]|\left[\SS_t(A_j)|-|\SS_t(A_j)|\right] \\
    & \quad\quad\quad\quad\quad\quad\quad\quad\quad\quad + \lambda \left[1-(1-\alpha)^{|\SS_t(A_j)|}\right]\left[|\SS_t(A_j)|-1-|\SS_t(A_j)|\right] \\
    & \quad\quad\quad\quad\quad\quad\quad\quad\quad\quad +(1-p\lambda) \pi^*_t(A_0,A_j)\left[|\SS_t(A_j)|-1-|\SS_t(A_j)|\right]\Biggl\},
\end{aligned}
\end{equation}
where $\lambda$ is defined by Eq.~\eqref{eqn:occuragents}. In the above derivation, we used the Poisson model of arrival in Section \ref{sec:agentsinnetwork} and the Poisson distribution in Eq.~\eqref{eqn:expdis2}.
\paragraph{The term (c):} 
In this case, there is one agent becoming critical but no agent arriving at the market during $[t,t+\Delta t]$. 
Recall that $|\SS_t|=\sum_{j=1}^{p}|\SS_t(A_j)|+|\SS_{t}(A_0)|$. 
We consider two scenarios of an agent becoming critical:
\begin{itemize}
\item First, if the critical agent is $A_k$-type with $k\neq j$ and $0\leq k\leq p$, then by the definition of Greedy Algorithm, it is incompatible with any $A_j$-type agent in the pool.
This is because otherwise, the critical agent has been matched with an $A_j$-type agent when either the critical $A_k$-type agent or an $A_j$-type agent has just arrived at the market. 
Thus, an $A_k$-type agent becoming critical with $k\neq j$ does not change the number of $A_j$-type agents in the pool. 
By Eqs.~\eqref{eqn:expdis1} and \eqref{eqn:expdis2}, this first scenario occurs with the probability,
\begin{equation*}
\begin{aligned}
    & \sum_{k=0,k\neq j}^p\P(\text{only one $A_k$-type agent becomes critical during $[t,t+\Delta t]$})\\
    & = \sum_{k=0,k\neq j}^p |\SS_t(A_k)| \left(1-e^{-\Delta t}\right) \left(e^{-\Delta t}\right)^{|\SS_t|-1}.
\end{aligned}
\end{equation*}
\item Second, if the critical agent is $A_j$-type, then it will decrease the number of $A_j$-type agents in the pool by $1$. 
Following the same argument above, this second scenario occurs with the probability
\begin{equation*}
\begin{aligned}
    & \P(\text{only one $A_j$-type agent becomes critical during $[t,t+\Delta t]$})\\
    & = |\SS_t(A_j)| \left(1-e^{-\Delta t}\right) \left(e^{-\Delta t}\right)^{|\SS_t|-1}.
\end{aligned}
\end{equation*}
\end{itemize}
Putting together the two scenarios, we have
\begin{equation}
\label{eqn:k1r0}
\begin{aligned}
& \P(\CC=1,\RR=0)\cdot\left\{\E[|\SS_{t+\Delta t}(A_j)|~|~\CC=1,\RR=0]-|\SS_t(A_j)|\right\}\\
& =\sum_{k=0,k\neq j}^{p}|\SS_t(A_k)|\left(1-e^{-\Delta t}\right)e^{-(|\SS_t|-1)\Delta t}\left[|\SS_{t}(A_j)|-|\SS_t(A_j)|\right]\\
&\quad\quad + |\SS_t(A_j)| \left(1-e^{-\Delta t}\right) \left(e^{-\Delta t}\right)^{|\SS_t|-1}\left[|\SS_{t}(A_j)|-1-|\SS_t(A_j)|\right]\\
& = -|\SS_t(A_j)| \left(1-e^{-\Delta t}\right) e^{-(|\SS_t|-1)\Delta t}.
\end{aligned}
\end{equation}
\paragraph{The term (d):} Using the result of Eq.~\eqref{eqn:expdis1}, we have,
\begin{align}
\label{eqn:k1r1}
\begin{split}
    &   \sum_{u\geq 1,v\geq 1}\P(\CC=u,\RR=v)\cdot\left\{\E[|\SS_{t+\Delta t}(A_j)|  ~|~ \CC=u,\RR=v]-|\SS_t(A_j)|\right\} \\ 
    &\leq \sum_{u\geq 1,v\geq 1} \P(\CC=u) \cdot \left[\frac{1}{v!}e^{-m \Delta t} (m\Delta t)^{v}\right] 
   \cdot v \\
    & \leq \sum_{1\leq u \leq |\SS_t|,v\geq 1} \binom{|\SS_t|}{u}
 (1-e^{-\Delta t})^{u} \cdot\left[\frac{1}{v!}e^{-m \Delta t} (m\Delta t)^{v}\right] \cdot v \\ 
    &=O( \Delta t^2 ),
\end{split}
\end{align}
where $\binom{|\SS_t|}{u}$ denotes the binomial coefficient of choosing $u$ agents from $|\SS_t|$ agents.
\paragraph{Putting terms (a)-(d) altogether:} 
Plugging Eqs.~\eqref{eqn:k0r0} and \eqref{eqn:k0r1}-\eqref{eqn:k1r1} into 
Eq.~\eqref{condition_probability}, and using the Taylor expansion, we obtain that,
\begin{equation*}
\begin{aligned}
 & \lim_{\Delta t\to 0+}\frac{\E[|\SS_{t+\Delta t}(A_j)|]-|\SS_t(A_j)|}{\Delta t}\\
& =-(1-p\lambda)m \pi^*_t(A_0,A_j) - \lambda m\left\{\left[1-(1-\alpha)^{|\SS_t(A_j)|}\right]-(1-\alpha)^{|\SS_t(A_j)|+|\SS_{t}(A_0)|}\right\} - |\SS_t(A_j)|. 
\end{aligned}
\end{equation*}
By Proposition \ref{thm:matchwithwho}, $\pi_t^*(A_j,A_0) = (1-\alpha)^{|\SS_t(A_{j})|}\left[1-(1-\alpha)^{|\SS_t(A_{0})|}\right]$, and $\pi^*_t(A_j,A_j) = 1-(1-\alpha)^{|\SS_t(A_{j})|}$. 
Then together with the definition in Eq.~\eqref{eqn:defofder}, we complete the proof for  $d|\SS_t(A_j)|/dt$ for $j\geq 1$ in Theorem \ref{thm:Greedy_limit}.
\end{proof}

\subsubsection{ODE for easy-to-match agents}
\begin{proof}
Next, we use a similar argument in Section \ref{sec:pfodeaj1} to derive ODE for the $A_0$-type agents. In particular, we also analyze the terms $(a)$-$(d)$ in Eq.~\eqref{condition_probability} separately. 

\paragraph{The term (a):} Same as Eq.~\eqref{eqn:k0r0}, we have
\begin{equation}
\label{eqn:k0r0a0}
\begin{aligned}
  \P(\CC=0,\RR=0)\cdot\left\{\E[|\SS_{t+\Delta t}(A_0)| ~|~ \CC=0,\RR=0]-|\SS_t(A_0)|\right\}   =0.
\end{aligned}
\end{equation}

\paragraph{The term (b):}  In this case, there is no agent becoming critical and one agent arriving at the market during $[t,t+\Delta t]$. 
The Greedy Algorithm will immediately try to match the new agent with existing agents in the pool. We consider five scenarios of the new agent arriving at the market:
\begin{itemize}
    \item First, if the new agent is $A_j$-type with $j\geq 1$, then by Proposition \ref{thm:matchwithwho}, it is compatible with an $A_0$-type agent with probability $\pi_t(A_j,A_0)$. If compatible, it will decrease the number of $A_0$-type agents in the pool by $1$.   
   \item Second, if the new agent is $A_j$-type with $j\geq 1$, then  it is incompatible with an $A_0$-type agent with probability $1-\pi_t(A_j,A_0)$. If incompatible, it does not change the number of $A_0$-type agents in the pool. 
   \item  Third, if the new agent is $A_0$-type, then with probability $\pi^*_t(A_0,A_j)$, it is compatible with an existing $A_j$-type agent in the pool with $j\geq 1$, and it does not change the number of $A_0$-type agents in the pool.
   \item  Fourth, if the new agent is $A_0$-type, then with probability $\pi^*_t(A_0,A_0)$, it is compatible with an existing $A_0$-type agent in the pool, and so decreases the number of $A_0$-type agents in the pool by $1$.
    \item Fifth, if the new agent is $A_0$-type, then with probability $(1-\alpha)^{|\SS_t|}$, it is incompatible with any agent in the pool, and so increases  the number of $A_0$-type agents in the pool by $1$.
\end{itemize}
Putting together the above five scenarios, we have
\begin{equation}
\label{eqn:k0r1a0}
\begin{aligned}
    & \P(\CC=0,\RR=1)\cdot\left\{\E[|\SS_{t+\Delta t}(A_0)|~|~\CC=0,\RR=1]-|\SS_t(A_0)|\right\}\\
    &=e^{-|\SS_t|\Delta t}e^{-m \Delta t}m\Delta t
    \Biggl\{\sum_{j=1}^{p}\lambda \pi_t^*(A_j,A_0) \left[|\SS_t(A_0)|-1-|\SS_t(A_0)|\right] \\
     & \quad\quad\quad\quad\quad\quad\quad\quad\quad\quad  + \sum_{j=1}^{p}\lambda (1-\pi_t^*(A_j,A_0)) \left[|\SS_t(A_0)|-|\SS_t(A_0)|\right] \\
    & \quad\quad\quad\quad\quad\quad\quad\quad\quad\quad +(1-p\lambda) \sum_{j=1}^p\pi^*_t(A_0,A_j)\left[|\SS_t(A_0)|-|\SS_t(A_0)|\right]\\
    & \quad\quad\quad\quad\quad\quad\quad\quad\quad\quad +(1-p\lambda) \pi^*_t(A_0,A_0)\left[|\SS_t(A_0)|-1-|\SS_t(A_0)|\right] \\
    & \quad\quad\quad\quad\quad\quad\quad\quad\quad\quad +(1-p\lambda) (1-\alpha)^{|\SS_t|}\left[|\SS_t(A_0)|+1-|\SS_t(A_0)|\right] \Biggl\},
\end{aligned}
\end{equation}
where we used the Poisson model of arrival in Section \ref{sec:agentsinnetwork} and the Poisson distribution in Eq.~\eqref{eqn:expdis2}.

\paragraph{The term (c):} In this case, there is one agent becoming critical but no agent arriving at the market during $[t,t+\Delta t]$. 
We consider two scenarios of an agent becoming critical:
\begin{itemize}
\item First, if the critical agent is $A_j$-type with $j\geq 1$, then by the definition of Greedy Algorithm, it is incompatible with any $A_0$-type agent in the pool.
This is because otherwise, the critical agent has been matched with an $A_0$-type agent when either the critical agent or an $A_0$-type agent has just arrived at the market. 
Thus, an $A_j$-type agent becoming critical with $j\geq 1$ does not change the number of $A_0$-type agents in the pool. 
Again, by Eqs.~\eqref{eqn:expdis1} and \eqref{eqn:expdis2}, this first scenario occurs with the probability,
\begin{equation*}
\begin{aligned}
    & \sum_{j=1}^p\P(\text{only one $A_j$-type agent becomes critical during $[t,t+\Delta t]$})\\
    & = \sum_{j=1}^p |\SS_t(A_j)| \left(1-e^{-\Delta t}\right) \left(e^{-\Delta t}\right)^{|\SS_t|-1}.
\end{aligned}
\end{equation*}
\item Second, if the critical agent is $A_0$-type, then it will decrease the number of $A_0$-type agents in the pool by $1$. 
This second scenario occurs with the probability
\begin{equation*}
\begin{aligned}
    & \P(\text{only one $A_0$-type agent becomes critical during $[t,t+\Delta t]$})\\
    & = |\SS_t(A_0)| \left(1-e^{-\Delta t}\right) \left(e^{-\Delta t}\right)^{|\SS_t|-1}.
\end{aligned}
\end{equation*}
\end{itemize}
Putting together the two scenarios, we have
\begin{equation}
\label{eqn:k1r0a0}
\begin{aligned}
& \P(\CC=1,\RR=0)\cdot\left\{\E[|\SS_{t+\Delta t}(A_0)|~|~\CC=1,\RR=0]-|\SS_t(A_0)|\right\}\\
& =\sum_{j=1}^{p}|\SS_t(A_j)|\left(1-e^{-\Delta t}\right)e^{-(|\SS_t|-1)\Delta t}\left[|\SS_{t}(A_0)|-|\SS_t(A_0)|\right]\\
&\quad\quad + |\SS_t(A_0)| \left(1-e^{-\Delta t}\right) \left(e^{-\Delta t}\right)^{|\SS_t|-1}\left[|\SS_{t}(A_0)|-1-|\SS_t(A_0)|\right]\\
& = -|\SS_t(A_0)| \left(1-e^{-\Delta t}\right) e^{-(|\SS_t|-1)\Delta t}.
\end{aligned}
\end{equation}
\paragraph{The term (d):} Same as Eq.~\eqref{eqn:k1r1}, we have
\begin{equation}
\label{eqn:k1r1a0}
\begin{aligned}
\sum_{u\geq 1,v\geq 1}\P(\CC=u,\RR=v)\cdot\left\{\E[|\SS_{t+\Delta t}(A_0)|  ~|~ \CC=u,\RR=v]-|\SS_t(A_0)|\right\} =O( \Delta t^2 ).
\end{aligned}
\end{equation}
\paragraph{Putting terms (a)-(d) altogether:} 
Plugging Eqs.~\eqref{eqn:k0r0a0}-\eqref{eqn:k1r1a0} into 
Eq.~\eqref{condition_probability}, and using the Taylor expansion, we obtain that,
\begin{equation*}
\begin{aligned}
 & \lim_{\Delta t\to 0+}\frac{\E[|\SS_{t+\Delta t}(A_0)|]-|\SS_t(A_0)|}{\Delta t}\\
 & = (1-p\lambda)m\left[(1-\alpha)^{|\SS_t|}-\pi_t^*(A_0,A_0)\right]-\lambda m\sum_{j=1}^p\pi_t^*(A_j,A_0)  - |\SS_t(A_0)|. 
\end{aligned}
\end{equation*}
By Eqs.~\eqref{eqn:a0incompat} and \eqref{eqn:defofder}, we complete the proof for  $d|\SS_t(A_0)|/dt$ in Theorem \ref{thm:Greedy_limit}.
\end{proof}

\subsection{Proof of Theorem \ref{main_result_greedy}}
\label{sec:pfthmgreedy}
\begin{proof}
By definition of the Greedy Algorithm, the expected number of agents perishing from the market during  $[t,t+dt]$ is $|\SS_t|dt$.  Hence the loss functions in Eqs.~\eqref{eqn:lossratio} and \eqref{eqn:lossforeachagent} can be equivalently written as,
\begin{equation}
\label{eqn:equivlossgreddy}
\begin{aligned}
    & L^{\text{greedy}}(A_0) =  \frac{\int_0^T|\SS_t(A_0)|dt}{(1-p\lambda)mT},\quad L^{\text{greedy}}(A_j) =  \frac{\int_0^T|\SS_t(A_j)|dt}{\lambda mT}, \quad \forall j\geq 1,\\
    &\text{and}\quad  
    L^{\text{greedy}} = \frac{\int_0^T|\SS_t|dt}{mT}.
\end{aligned}
\end{equation}
The proof of Theorem \ref{main_result_greedy} is completed in five parts: first, we establish that the ODEs in Theorem \ref{thm:Greedy_limit} have a unique solution; 
second, we prove the solution has a symmetric structure and simplifies the ODE system; 
third,  we prove that there exists a stationary solution to the ODE system;
fourth, we show that the ODE solution converges to a stationary solution using the Poincaré–Bendixson Theorem; finally, we analyze the loss of the Greedy Algorithm.
\paragraph{Part 1: Existence and uniqueness of the solution.} To prove the existence and uniqueness of the solution, we resort to a classical result of initial value problems in the literature.
\begin{lemma}[cf. Theorem 2.3 in \cite{coddington1955theory}]
\label{existence and uniqueness}
Let $D \subseteq \mathop R\times \mathop R^n$ be an open set with $(t_0,x_0)\in D$. If $F(t,x):D \to \mathop R^n$ is continuous in $t$ and Lipschitz continuous in $x$, then there exists some $\epsilon>0$, so that the initial value problem:
\begin{equation*}
\begin{aligned}
  &\frac{d x(t)}{dt} && =\quad F(t,x(t))\\
  &x(t_0) && =\quad x_0
 \end{aligned}
\end{equation*}
has a unique solution $x(t)$ on the interval $[t_0-\epsilon, t_0+\epsilon]$.
\end{lemma}
We go back to the ODEs in Theorem \ref{thm:Greedy_limit}. Note that the functions $\pi_t^*(A_j,A_k)$ for $j,k\geq 0$ that are defined in Proposition \ref{thm:matchwithwho} are continuous in $t$ and differentiable in $|\SS_t(A_k)|$. Hence given any initial value $(|\SS_0(A_0)|,|\SS_0(A_1)|,\ldots,|\SS_0(A_p)|)$ with $|\SS_0(A_k)|\geq 0$, Lemma \ref{existence and uniqueness} implies the existence and uniqueness of the solution to the ODEs in Theorem \ref{thm:Greedy_limit}. 

\paragraph{Part 2: A symmetric structure of the solution.} We prove that when the initial values satisfy $|\SS_0(A_1)|=\cdots=|\SS_0(A_p)|\geq 0$, then 
\begin{equation}
\label{eqn:equivamongaj}
|\SS_t(A_1)|=\cdots=|\SS_t(A_p)|,\quad\forall t>0.
\end{equation} 
This result can be proved by contradiction. Suppose there exists $t_0>0$ and $j_1\neq j_2\geq 1$, such that $|\SS_{t_0}(A_{j_1})|\neq |\SS_{t_0}(A_{j_2})|$. In this case,
we define $(|\hat{\SS}_t(A_1)|,\ldots,|\hat{\SS}_t(A_p)|)$ as follows:
\begin{equation*}
|\hat{\SS}_t(A_j)|=\begin{cases}
|\SS_t(A_{j_2})|, & \text{ if } j=j_1, \\ 
|\SS_t(A_{j_1})|, & \text{ if } j=j_2, \\ 
|\SS_t(A_j)|, & \text{ if } j\neq j_1,j_2.
\end{cases}
\end{equation*}
Then by symmetry, $(|\SS_t(A_0)|, |\hat{\SS}_t(A_1)|,\ldots,|\hat{\SS}_t(A_p)|)$ is a solution to the ODEs in Theorem \ref{thm:Greedy_limit}. However, 
\begin{equation*}
     (|\SS_{t_0}(A_0)|, |\hat{\SS}_{t_0}(A_1)|,\ldots,|\hat{\SS}_{t_0}(A_p)|) \neq  (|\SS_{t_0}(A_0)|,|\SS_{t_0}(A_1)|,\ldots,|\SS_{t_0}(A_p)|).
\end{equation*}
This contradicts with the uniqueness of the solution, as shown in Part 1.

A byproduct of the solution symmetry in Eq.~\eqref{eqn:equivamongaj} is that the ODEs in Theorem \ref{thm:Greedy_limit} can be reduced into a group of two-dimensional ODEs, 
\begin{equation}
\label{2d-greedy_ODE}
\begin{aligned}
\frac{d|\SS_t(A_0)|}{dt}  & =   (1-p\lambda)m (1-\alpha)^{|\SS_t(A_0)| + p|\SS_t(A_1)|} \\
& \quad \quad - (1-p\lambda)m  \left[1-(1-\alpha)^{|\SS_t(A_0)|}\right](1-\alpha)^{p|\SS_t(A_{1})|} \\
&\quad \quad - \lambda mp  (1-\alpha)^{|\SS_t(A_{1})|}\left[1-(1-\alpha)^{|\SS_t(A_{0})|}\right] - |\SS_t(A_0)|,\\
\frac{d|\SS_t(A_1)|}{dt} &  = 
 - (1-p\lambda)m\frac{\left[1-(1-\alpha)^{p|\SS_t(A_1)|}\right]}{p} -\lambda m\left[1-(1-\alpha)^{|\SS_t(A_{1})|}\right]\\
& \quad \quad +\lambda m (1-\alpha)^{|\SS_t(A_{0})|+|\SS_t(A_{1})|}   - |\SS_t(A_1)|.
\end{aligned}
\end{equation}
where we used Proposition \ref{thm:matchwithwho} to derive the explicit forms of $\pi_t^*(A_j,A_k)$ for $j,k\geq 0$.

\paragraph{Part 3: Existence of a stationary solution.}  We show that the ODE system in Eq.~\eqref{2d-greedy_ODE} has a stationary solution that doesn't depend on time. Suppose  $(|\SS(A_0)|,|\SS(A_1)|)$ is the stationary solution to Eq.~\eqref{2d-greedy_ODE}. Then
\begin{equation}
\label{2d-greedy-stable-2}
\begin{aligned}
0  & =   (1-p\lambda)m (1-\alpha)^{|\SS(A_0)| + p|\SS(A_1)|} \\
& \quad \quad - (1-p\lambda)m  \left[1-(1-\alpha)^{|\SS(A_0)|}\right](1-\alpha)^{p|\SS(A_{1})|} \\
&\quad \quad - \lambda mp  (1-\alpha)^{|\SS(A_{1})|}\left[1-(1-\alpha)^{|\SS(A_{0})|}\right] - |\SS(A_0)|,
\end{aligned}
\end{equation}
and 
\begin{equation}
\label{2d-greedy-stable-1}
\begin{aligned}
0 &  = 
 - (1-p\lambda)m\frac{\left[1-(1-\alpha)^{p|\SS(A_1)|}\right]}{p} -\lambda m\left[1-(1-\alpha)^{|\SS(A_{1})|}\right]\\
& \quad \quad +\lambda m (1-\alpha)^{|\SS(A_{0})|+|\SS(A_{1})|}   - |\SS(A_1)|.
\end{aligned}
\end{equation}
By Eq.~\eqref{2d-greedy-stable-1}, we can derive
\begin{equation}
\label{express1}
\begin{aligned}
& \lambda m (1-\alpha)^{|\SS(A_{0})|+|\SS(A_{1})|} \\
& = \lambda m \left[1-(1-\alpha)^{|\SS(A_{1})|}\right] +(1-p\lambda)m\frac{\left[1-(1-\alpha)^{p|\SS(A_1)|}\right]}{p}+|\SS(A_1)|.
\end{aligned}
\end{equation}
Plugging Eq.~\eqref{express1} into Eq.~\eqref{2d-greedy-stable-2}, we have,
\begin{equation}
\label{Z_O=greedy_f(Z_A)}
\begin{aligned}
|\SS(A_0)| & = -\lambda mp(1-\alpha)^{|\SS(A_1)|} + 1 +p|\SS(A_1)|-4(1-p\lambda)m(1-\alpha)^{p|\SS(A_1)|}\\
& \quad +2m\frac{(1-p\lambda)}{p\lambda }(1-\alpha)^{(p-1)|\SS(A_1)|} -2\frac{(1-p\lambda)^2m}{p\lambda}(1-\alpha)^{(2p-1)|\SS(A_1)|}\\
&\quad +2\frac{(1-p\lambda)}{\lambda}(1-\alpha)^{(p-1)|\SS(A_1)|}|\SS(A_1)|.
\end{aligned}
\end{equation}
Hence we can write 
\begin{equation*}
    |\SS(A_0)| = f(|\SS(A_1)|),
\end{equation*} 
where Eq.~\eqref{Z_O=greedy_f(Z_A)} implies that $f(\cdot)$ is a continuous function of $|\SS(A_1)|$. Moreover, 
\begin{equation}
\label{eqn:limitfbeh}
\lim_{m\to\infty}\lim_{|\SS(A_1)| \to 0} f(|\SS(A_1)|)=-\infty,\ \text{ and } \ \lim_{|\SS(A_1)| \to \infty} f(|\SS(A_1)|)=\infty.
\end{equation}

On the other hand,
Eq.~\eqref{2d-greedy-stable-1} yields that,
\begin{equation}
\label{Z_O=greedy_g(Z_A)}
\begin{aligned}
 |\SS(A_0)| =\frac{1}{\ln{(1-\alpha)}}& \ln\Bigg( (1-\alpha)^{-|\SS(A_1)|}-1    + \frac{|\SS(A_1)| (1-\alpha)^{-|\SS(A_1)|} }{\lambda m} \\
 & \quad\quad \quad  +\left(\frac{1}{\lambda p}-1\right) \left[(1-\alpha)^{-|\SS(A_1)|} -(1-\alpha)^{(p-1)|\SS(A_1)| }\right] \Bigg).
 \end{aligned}
\end{equation}
Hence we can write
\begin{equation*}
    |\SS(A_0)| = g(|\SS(A_1)|),
\end{equation*} 
where Eq.~\eqref{Z_O=greedy_g(Z_A)} implies that $g$ is a continuous and decreasing function of $\SS(A_1)$. Moreover,
\begin{equation}
\label{eqn:limitgbeh}
 \lim_{m\to\infty}\lim_{|\SS(A_1)| \to 0} g(|\SS(A_1)|)=\infty, \ \text{ and }\ \lim_{|\SS(A_1)| \to \infty }g(|\SS(A_1)|)=-\infty.
\end{equation}
Combining Eqs.~\eqref{eqn:limitfbeh} and \eqref{eqn:limitgbeh}, we know that functions $f$ and $g$ have at least one intersection in $\{ |\SS(A_1)|>0 \}$. Moreover, since Eq.~\eqref{2d-greedy-stable-2} requires the solution $|\SS(A_0)|>0$, we have that the stationary solution $(|\SS(A_0)|,|\SS(A_1)|)$ to Eqs.~\eqref{2d-greedy-stable-2}-\eqref{2d-greedy-stable-1}
exists and lies in $(0,\infty)\times (0,\infty)$.

\paragraph{Part 4: Convergence to a stationary solution.}
We show that the solution to the ODEs in Eq.~\eqref{2d-greedy_ODE} will converge to a stationary solution. To prove this result, we use two important results in dynamic systems.
\begin{lemma}[\emph{Bendixson Criterion}; cf. \citet{andronov2013theory}]
\label{Bendixson Criterion}
Consider a system:
\begin{align*}
&\frac{dx(t)}{dt}=F(x,y),\\
&\frac{dy(t)}{dt}=G(x,y).
\end{align*}
If $\frac{\partial F}{ \partial x}+\frac{\partial G}{\partial y}$ has the same sign $({\displaystyle \neq 0})$ in a simply connected region $G$ of the two-dimensional plane, then the system has no  periodic trajectories in the domain $G$.
\end{lemma}
\begin{lemma}[\emph{Poincaré–Bendixson Theorem}; cf. \citet{ciesielski2012poincare}]
\label{lem:poincarebendixson}
Consider a system:
\begin{align*}
&\frac{dx(t)}{dt}=F(x,y),\\
&\frac{dy(t)}{dt}=G(x,y).
\end{align*}
If the solution to this system is unique and defined for all $t\geq0$, then it is either a periodic trajectory or  a fixed point.
\end{lemma}
We go back to the ODES in Eq.~\eqref{2d-greedy_ODE}, where for any initial values, $|\SS^*(A_0)|, |\SS^*(A_1)|>0$,
\begin{equation*}
\begin{aligned}
    & \left.\frac{d|\SS_t(A_0)|}{dt}\right|_{|\SS_0(A_0)|=0,|\SS_0(A_1)|=0}>0,\ && \text{ and } \quad \left.\frac{d|\SS_t(A_1)|}{dt}\right|_{|\SS_0(A_0)|=0,|\SS_0(A_1)|=0}>0,\\
    &  \left.\frac{d|\SS_t(A_0)|}{dt}\right|_{|\SS_0(A_0)|=0,|\SS_0(A_1)|=|\SS^*(A_1)|}>0,\  && \text{ and } \quad \left.\frac{d|\SS_t(A_1)|}{dt}\right|_{|\SS_0(A_0)|=|\SS^*(A_0)|,|\SS_0(A_1)|=0}>0.
\end{aligned}
\end{equation*}
Moreover, when $|\SS_t(A_0)|, |\SS_t(A_1)|>0$, we have,
\begin{equation*}
    \frac{d|\SS_t(A_0)|}{dt}<(1-p\lambda)m-|\SS_t(A_0)|, \ \text{ and }\
    \frac{d|\SS_t(A_1)|}{dt}<\lambda m-|\SS_t(A_1)|.
\end{equation*}
Hence the trajectory of solution $(|\SS_t(A_0)|, |\SS_t(A_1)|)$ will be in a bounded area in $[0,\infty)\times [0,\infty)$.  On the other hand, by Lemma \ref{Bendixson Criterion}, the solution to ODES in Eq.~\eqref{2d-greedy_ODE} is not a periodic trajectory. Then by Lemma \ref{lem:poincarebendixson}, the solution to ODES in Eq.~\eqref{2d-greedy_ODE} converges to a fixed point that is a stationary solution in $(0,\infty)\times (0,\infty)$.  

\paragraph{Part 5: Property of the convergent stationary solution.}
We now analyze the properties of the stationary solution of Eq.~\eqref{2d-greedy_ODE}.
First, we show that when  $m\to \infty$ and $\alpha\to 0$, there exists a constant $C_1$ such that 
\begin{equation}
\label{eqn:alphassa1c}
    \alpha|\SS(A_1)|\leq C_1.
\end{equation}
We prove \eqref{eqn:alphassa1c} by contradiction. 
\begin{equation}
\label{eqn:ineqalambdam}
\begin{aligned}
&\lambda m_k \left[1-(1-\alpha_k)^{|\SS(A_{1})|}\right] +(1-p\lambda)m_k\frac{\left[1-(1-\alpha_k)^{p|\SS(A_1)|}\right]}{p}+|\SS(A_1)| \\
& \geq \lambda m_k \left[1-(1-\alpha_k)^{ \frac{k}{\alpha_k}}\right]+(1-p\lambda)m_k 
\frac{[1-(1-\alpha_k)^{p \frac{k}{\alpha_k}}]}{p}+|\SS(A_1)| \\ &\geq
\lambda m_k (1-\epsilon_k)+(1-p\lambda )m_k \frac{1-\epsilon_k}{p} \\
& \geq m_k \left[\lambda + \frac{1-p\lambda}{p} -\lambda \epsilon_k - \frac{1-p\lambda}{p}\epsilon_k\right].
\end{aligned}
\end{equation}
Note that $\epsilon_k\to e^{-k}$ when $\alpha_k\to 0$. Hence for large enough $k$, we have
\begin{equation*}
    \epsilon_k < \frac{1-p\lambda}{2},
\end{equation*}
which implies that 
\begin{equation}
\label{eqn:ineqlep}
    -\lambda \epsilon_k - \frac{1-p\lambda}{p}\epsilon_k  > -\frac{1-p\lambda}{2p}.
\end{equation}
Combining Eqs.~\eqref{eqn:ineqalambdam} and \eqref{eqn:ineqlep}, we have
\begin{equation}
\label{eqn:ineqlamdabnms}
\begin{aligned}
&\lambda m_k \left[1-(1-\alpha_k)^{|\SS(A_{1})|}\right] +(1-p\lambda)m_k\frac{\left[1-(1-\alpha_k)^{p|\SS(A_1)|}\right]}{p}+|\SS(A_1)| \\
& > m_k \left[\lambda + \frac{1-p\lambda}{2p}\right]  \\
& > \lambda m_k.
\end{aligned}
\end{equation}
On the other hand, when $k\to\infty$, we have
\begin{equation}
\label{eqn:lambmaso}
    \lambda m (1-\alpha)^{|\SS(A_{0})|+|\SS(A_{1})|} \leq \lambda m_k
\end{equation}
Combining Eqs.~\eqref{eqn:ineqlamdabnms} and \eqref{eqn:lambmaso}, we have
\begin{equation*}
\begin{aligned}
&\lambda m_k \left[1-(1-\alpha_k)^{|\SS(A_{1})|}\right] +(1-p\lambda)m_k\frac{\left[1-(1-\alpha_k)^{p|\SS(A_1)|}\right]}{p}+|\SS(A_1)| \\
& > \lambda m (1-\alpha)^{|\SS(A_{0})|+|\SS(A_{1})|}. 
\end{aligned}
\end{equation*}
This contradicts Eq.~\eqref{express1}. Therefore, the result in Eq.~\eqref{eqn:alphassa1c} holds.

Next, we show that when $m\to\infty$ and $\alpha\to 0$, there exists a constant $C_1$ such that,
\begin{equation}
\label{eqn:alphassa1c2}
    \alpha|\SS(A_1)|\geq  C_1.
\end{equation}
We also prove \eqref{eqn:alphassa1c2} by contradiction. 
Suppose that there exist $\{m_k,\alpha_k,\epsilon_k'\}$ such that 
$m_k\to\infty$, $\alpha_k\to 0, \epsilon_k'\to 0$, and 
\begin{equation*}
\alpha_k|\SS(A_1)|\leq \epsilon_k', \quad \text{as }k\to\infty.
\end{equation*}
In this case, by Eq.~\eqref{Z_O=greedy_f(Z_A)}, we can show that 
\begin{equation*}
|\SS(A_0)|\leq -\frac{p\lambda m}{2}, \quad \text{as }k\to\infty,
\end{equation*}
which contradicts with $|\SS(A_0)| >0$. 
Therefore, Eqs.~\eqref{eqn:alphassa1c} and \eqref{eqn:alphassa1c2} suggest that there exist constants $C_1$ and $C'_1$ such that $ C_1\leq \alpha|\SS(A_1)|\leq C'_1$, which implies that for any $t\in[0,T]$,
\begin{equation}
\label{eqn:bdonssa1alpha}
   C_1\alpha^{-1}\leq |\SS(A_1)| \leq C'_1\alpha^{-1}.
\end{equation}
Moreover, by Eq.~\eqref{express1}, we have that 
\begin{equation*}
 (1-\alpha)^{|\SS(A_0)|}=\frac{\lambda m \left[1-(1-\alpha)^{|\SS(A_1)|}\right]+(1-p\lambda)m \frac{[1-(1-\alpha)^{p |\SS(A_1)|}]}{p}+|\SS(A_1)|}{\lambda m}. 
\end{equation*}
Hence when $m\to\infty$ and $\alpha\to 0$, there exist constant $0<C_\alpha\leq C'_\alpha<\infty$ such that,  
\begin{equation*}
    C_\alpha\leq (1-\alpha)^{|\SS(A_0)|}\leq C'_\alpha.
\end{equation*}
Thus, there exists constants $0<C_0<C'_0<\infty$ such that for any $t\in[0,T]$, 
\begin{equation}
\label{eqn:bdonssa1alpha0}
    C_0\alpha^{-1}\leq |\SS(A_0)|\leq C'_0\alpha^{-1}.
\end{equation}

By the above Part 4, the solution to the ODEs in Eq.~\eqref{2d-greedy_ODE} will converge to a stationary solution. Hence the loss functions defined in Eq.~\eqref{eqn:equivlossgreddy} satisfy
\begin{equation*}
     \lim_{T \to \infty} L^{\text{greedy}}(A_1) = \frac{\lim_{T\to\infty}|\SS(A_1)|}{\lambda m}.
\end{equation*}
By Eq.~\eqref{eqn:bdonssa1alpha}, there exist constants $0<c_1\leq c'_1<\infty$ such that,
\begin{equation*}
     c_1d^{-1}\leq \lim_{T \to \infty} L^{\text{greedy}}(A_1) \leq c'_1d^{-1}.
\end{equation*}
Similarly, the above Part 4 and Eq.~\eqref{eqn:bdonssa1alpha0} imply that there exist constants $0<c_2\leq c'_2<\infty$ such that,
\begin{equation*}
     c_2d^{-1}\leq \lim_{T \to \infty} L^{\text{greedy}}(A_0) \leq c'_2d^{-1}.
\end{equation*}
Together, we also have constants $0<c_3\leq c'_3<\infty$ such that,
\begin{equation*}
     c_3d^{-1}\leq \lim_{T \to \infty} L^{\text{greedy}} \leq c'_3d^{-1}.
\end{equation*}
Letting $c = \min\{c_1,c_2,c_3\}$ and $c' = \max\{c'_1,c'_2,c'_3\}$, we complete the proof of Theorem \ref{main_result_greedy}. 
\end{proof}

\subsection{Proof of Theorem \ref{thm:Patientlimit}}
Similar to the proof in Section \ref{thm:Greedy_limit}, we
assume that at time interval $[t,t+\Delta t]$, there are $\CC$ agents that have already stayed in the market before time $t$ become critical, $\RR$ new agents arrive at the market. Then for any $A_k$-type agents, $k=0,1,\ldots,p$, we have the decomposition in Eq.~\eqref{condition_probability}.
In the following, we derive the ODE for  $A_j$-type agents with $j\geq 1$, and then use a similar argument to derive the ODE for $A_0$-type agents.

\subsubsection{ODE for hard-to-match agents}
\label{sec:pfodeaj1_patient}
\begin{proof}
First, we consider $A_j$-type agents with $j\geq 1$, where
we analyze the terms $(a)$-$(d)$ in Eq.~\eqref{condition_probability} separately. 
\paragraph{The term (a):}  Same as Eq.~\eqref{eqn:k0r0}, we have
\begin{equation}
\label{eqn:patient_00}
\begin{aligned}
  \P(\CC=0,\RR=0)\cdot\left\{\E[|\SS_{t+\Delta t}(A_j)| ~|~ \CC=0,\RR=0]-|\SS_t(A_j)|\right\}   =0.
\end{aligned}
\end{equation}
\paragraph{The term (b):}  In this case, there is no agent becoming critical and one agent arriving at the market during $[t,t+\Delta t]$. 
By definition, the Patient Algorithm will not immediately act on the new agent. We consider two scenarios of the new agent arriving at the market:
\begin{itemize}
    \item First, if the new agent is $A_k$-type where $k\neq j$ and $0\leq k\leq p$, then it does not change the number of $A_j$-type agents in the pool. 
    \item Second, if the new agent is $A_j$-type, then it increases the number of $A_j$-type agents in the pool by $1$.
\end{itemize}
Putting together the above two scenarios, we have
\begin{equation}
\label{patient_01}
\begin{aligned}
    & \P(\CC=0,\RR=1)\cdot\left\{\E[|\SS_{t+\Delta t}(A_j)|~|~\CC=0,\RR=1]-|\SS_t(A_j)|\right\}\\
    &=e^{-|\SS_t|\Delta t}e^{-m \Delta t}m\Delta t
    \Biggl\{\sum_{1\leq k \leq p, k\neq j}\lambda \left[|\SS_t(A_j)|-|\SS_t(A_j)|\right] \\
     & \quad\quad\quad\quad\quad\quad\quad\quad\quad\quad\quad\quad  +(1-p\lambda) \left[|\SS_t(A_j)|-|\SS_t(A_j)|\right] \\
    & \quad\quad\quad\quad\quad\quad\quad\quad\quad\quad\quad\quad  +\lambda\left[|\SS_t(A_j)|+1-|\SS_t(A_j)|\right] \Biggl\},
\end{aligned}
\end{equation}
\paragraph{The term (c):} In this case, there is one agent becoming critical but no agent arriving at the market during $[t,t+\Delta t]$. 
We consider five scenarios of an agent becoming critical:
\begin{itemize}
\item First, if the critical agent is $A_k$-type with $k\neq j$ and $1\leq k\leq p$, then it is incompatible with any $A_j$-type agent in the pool and does not change the number of $A_j$-type agents in the pool. 
By Eqs.~\eqref{eqn:expdis1} and \eqref{eqn:expdis2}, this scenario occurs with the probability,
\begin{equation*}
\begin{aligned}
    & \sum_{k=1,k\neq j}^p\P(\text{only one $A_k$-type agent becomes critical during $[t,t+\Delta t]$})\\
    & = \sum_{k=1,k\neq j}^p |\SS_t(A_k)| \left(1-e^{-\Delta t}\right) \left(e^{-\Delta t}\right)^{|\SS_t|-1}.
\end{aligned}
\end{equation*}
\item Second, if the critical agent is $A_0$-type with, then the planner would match the critical agent with an $A_j$-type agent in the pool with probability $\pi_t^*(A_0,A_j)$. It will decrease the number of $A_j$-type agents in the pool by $1$. 
This scenario occurs with the probability,
\begin{equation*}
\begin{aligned}
    & \P(\text{only one $A_0$-type agent becomes critical and is matched to an $A_j$-type agent})\\
    & = |\SS_t(A_0)| \left(1-e^{-\Delta t}\right) \left(e^{-\Delta t}\right)^{|\SS_t|-1}\pi_t^*(A_0,A_j).
\end{aligned}
\end{equation*}
\item Third, if the critical agent is $A_0$-type with, then the planner would \emph{not} match the critical agent with an $A_j$-type agent in the pool with probability $1-\pi_t^*(A_0,A_j)$. It does not change the number of $A_j$-type agents in the pool. 
This scenario occurs with the probability,
\begin{equation*}
\begin{aligned}
    & \P(\text{only one $A_0$-type agent becomes critical and is not matched to an $A_j$-type agent})\\
    & = |\SS_t(A_0)| \left(1-e^{-\Delta t}\right) \left(e^{-\Delta t}\right)^{|\SS_t|-1}\left[1-\pi_t^*(A_0,A_j)\right].
\end{aligned}
\end{equation*}
\item Fourth, if the critical agent is $A_j$-type, then the planner would match the critical agent with an $A_j$-type agent in the pool with probability $\pi_t^*(A_j,A_j)$. It will decrease the number of $A_j$-type agents in the pool by $2$. 
This scenario occurs with the probability,
\begin{equation*}
\begin{aligned}
    & \P(\text{only one $A_j$-type agent becomes critical and is matched to an $A_j$-type agent})\\
    & = |\SS_t(A_j)| \left(1-e^{-\Delta t}\right) \left(e^{-\Delta t}\right)^{|\SS_t|-1}\pi_t^*(A_j,A_j).
\end{aligned}
\end{equation*}
\item Fifth, if the critical agent is $A_j$-type, then the planner would \emph{not} match the critical agent with an $A_j$-type agent in the pool with probability $1-\pi_t^*(A_j,A_j)$. It will decrease the number of $A_j$-type agents in the pool by $1$. 
This scenario occurs with the probability,
\begin{equation*}
\begin{aligned}
    & \P(\text{only one $A_j$-type agent becomes critical and is not matched to an $A_j$-type agent})\\
    & = |\SS_t(A_j)| \left(1-e^{-\Delta t}\right) \left(e^{-\Delta t}\right)^{|\SS_t|-1}\left[1-\pi_t^*(A_j,A_j)\right].
\end{aligned}
\end{equation*}
\end{itemize}
Putting together the five scenarios, we have
\begin{equation}
\label{patient_10_aj}
\begin{aligned}
& \P(\CC=1,\RR=0)\cdot\left\{\E[|\SS_{t+\Delta t}(A_j)|~|~\CC=1,\RR=0]-|\SS_t(A_j)|\right\}\\
& =\left(1-e^{-\Delta t}\right)e^{-(|\SS_t|-1)\Delta t}\Bigg\{\sum_{k=1,k\neq j}^{p}|\SS_t(A_k)|\left[|\SS_{t}(A_j)|-|\SS_t(A_j)|\right]\\
&\quad\quad\quad\quad\quad\quad + |\SS_t(A_0)| \pi_t^*(A_0,A_j)\left[|\SS_{t}(A_j)|-1-|\SS_t(A_j)|\right]\\
&\quad\quad\quad\quad\quad\quad + |\SS_t(A_0)| \left[1-\pi_t^*(A_0,A_j)\right]\left[|\SS_{t}(A_j)|-|\SS_t(A_j)|\right]\\
&\quad\quad\quad\quad\quad\quad + |\SS_t(A_j)| \pi_t^*(A_j,A_j)\left[|\SS_{t}(A_j)|-2-|\SS_t(A_j)|\right]\\
&\quad\quad\quad\quad\quad\quad + |\SS_t(A_j)| \left[1-\pi_t^*(A_j,A_j)\right]\left[|\SS_{t}(A_j)|-1-|\SS_t(A_j)|\right]\Bigg\}\\
& = \left(1-e^{-\Delta t}\right) e^{-(|\SS_t|-1)\Delta t}\Bigg\{-|\SS_t(A_0)| \pi_t^*(A_0,A_j) -2|\SS_t(A_j)| \pi_t^*(A_j,A_j)\\
&\quad\quad\quad\quad\quad\quad -|\SS_t(A_j)|\left[1-\pi_t^*(A_j,A_j)\right]\Bigg\}.
\end{aligned}
\end{equation}
\paragraph{The term (d):} Same as Eq.~\eqref{eqn:k1r1}, we have
\begin{equation}
\label{eqn:k1r1a0_patient}
\begin{aligned}
\sum_{u\geq 1,v\geq 1}\P(\CC=u,\RR=v)\cdot\left\{\E[|\SS_{t+\Delta t}(A_j)|  ~|~ \CC=u,\RR=v]-|\SS_t(A_j)|\right\} =O( \Delta t^2 ).
\end{aligned}
\end{equation}
\paragraph{Putting terms (a)-(d) altogether:} 
Plugging Eqs.~\eqref{eqn:patient_00}-\eqref{eqn:k1r1a0_patient} into 
Eq.~\eqref{condition_probability}, and using the Taylor expansion, we obtain that,
\begin{equation*}
\begin{aligned}
 & \lim_{\Delta t\to 0+}\frac{\E[|\SS_{t+\Delta t}(A_j)|]-|\SS_t(A_j)|}{\Delta t}\\
& =\lambda m-|\SS_t(A_0)| \pi_t^*(A_0,A_j) -|\SS_t(A_j)| \pi_t^*(A_j,A_j)-|\SS_t(A_j)|,\quad\forall j\geq 1.
\end{aligned}
\end{equation*}
By the definition in Eq.~\eqref{eqn:defofder}, we complete the proof for  $d|\SS_t(A_j)|/dt$ for $j\geq 1$.
\end{proof}

\subsubsection{ODE for easy-to-match agents}
\begin{proof}
Next, we use a similar argument in Section \ref{sec:pfodeaj1_patient} to derive ODE for the $A_0$-type agents. In particular, we also analyze the terms $(a)$-$(d)$ in Eq.~\eqref{condition_probability} separately. 

\paragraph{The term (a):}  Same as Eq.~\eqref{eqn:k0r0}, we have
\begin{equation}
\label{eqn:patient_00a0}
\begin{aligned}
  \P(\CC=0,\RR=0)\cdot\left\{\E[|\SS_{t+\Delta t}(A_0)| ~|~ \CC=0,\RR=0]-|\SS_t(A_0)|\right\}   =0.
\end{aligned}
\end{equation}
\paragraph{The term (b):} In this case, there is no agent becoming critical and one agent arriving at the market during $[t,t+\Delta t]$. We consider two scenarios of the new agent arriving at the market:
\begin{itemize}
    \item First, if the new agent is $A_j$-type where $j\geq 1$, then it does not change the number of $A_0$-type agents in the pool. 
    \item Second, if the new agent is $A_0$-type, then it increases the number of $A_0$-type agents in the pool by $1$.
\end{itemize}
Putting together the above two scenarios, we have
\begin{equation}
\label{patient_01_agent0}
\begin{aligned}
    & \P(\CC=0,\RR=1)\cdot\left\{\E[|\SS_{t+\Delta t}(A_0)|~|~\CC=0,\RR=1]-|\SS_t(A_0)|\right\}\\
    &=e^{-|\SS_t|\Delta t}e^{-m \Delta t}m\Delta t
    \Biggl\{\sum_{j=1}^p\lambda \left[|\SS_t(A_j)|-|\SS_t(A_j)|\right] \\
     & \quad\quad\quad\quad\quad\quad\quad\quad\quad\quad\quad\quad  +(1-p\lambda) \left[|\SS_t(A_j)|+1-|\SS_t(A_j)|\right] \Biggl\},
\end{aligned}
\end{equation}
\paragraph{The term (c):}  In this case, there is one agent becoming critical but no agent arriving at the market during $[t,t+\Delta t]$. 
We consider four scenarios of an agent becoming critical:
\begin{itemize}
\item First, if the critical agent is $A_j$-type with $j\geq 1$, then the planner would match the critical agent with an $A_0$-type agent in the pool with probability $\pi_t^*(A_j,A_0)$. It will decrease the number of $A_0$-type agents in the pool by $1$. 
This scenario occurs with the probability,
\begin{equation*}
\begin{aligned}
    & \P(\text{only one $A_j$-type agent becomes critical and is matched to an $A_0$-type agent})\\
    & = |\SS_t(A_j)| \left(1-e^{-\Delta t}\right) \left(e^{-\Delta t}\right)^{|\SS_t|-1}\pi_t^*(A_j,A_0).
\end{aligned}
\end{equation*}
\item Second, if the critical agent is $A_j$-type with $j\geq 1$, then the planner would \emph{not} match the critical agent with an $A_0$-type agent in the pool with probability $1-\pi_t^*(A_j,A_0)$. It does not change the number of $A_0$-type agents in the pool. 
This scenario occurs with the probability,
\begin{equation*}
\begin{aligned}
    & \P(\text{only one $A_j$-type agent becomes critical and is not matched to an $A_0$-type agent})\\
    & = |\SS_t(A_j)| \left(1-e^{-\Delta t}\right) \left(e^{-\Delta t}\right)^{|\SS_t|-1}\left[1-\pi_t^*(A_j,A_0)\right].
\end{aligned}
\end{equation*}
\item Third, if the critical agent is $A_0$-type with, then the planner would match the critical agent with an $A_0$-type agent in the pool with probability $\pi_t^*(A_0,A_0)$. It will decrease the number of $A_0$-type agents in the pool by $2$.
This scenario occurs with the probability,
\begin{equation*}
\begin{aligned}
    & \P(\text{only one $A_0$-type agent becomes critical and is matched to an $A_0$-type agent})\\
    & = |\SS_t(A_0)| \left(1-e^{-\Delta t}\right) \left(e^{-\Delta t}\right)^{|\SS_t|-1}\pi_t^*(A_0,A_0).
\end{aligned}
\end{equation*}
\item Fourth, if the critical agent is $A_0$-type, then the planner would \emph{not} match the critical agent with an $A_0$-type agent in the pool with probability $1-\pi_t^*(A_0,A_0)$. It will decrease the number of $A_0$-type agents in the pool by $1$. 
This scenario occurs with the probability,
\begin{equation*}
\begin{aligned}
    & \P(\text{only one $A_0$-type agent becomes critical and is not matched to an $A_0$-type agent})\\
    & = |\SS_t(A_0)| \left(1-e^{-\Delta t}\right) \left(e^{-\Delta t}\right)^{|\SS_t|-1}\left[1-\pi_t^*(A_0,A_0)\right].
\end{aligned}
\end{equation*}
\end{itemize}
Putting together the four scenarios, we have
\begin{equation}
\label{patient_10_a0}
\begin{aligned}
& \P(\CC=1,\RR=0)\cdot\left\{\E[|\SS_{t+\Delta t}(A_j)|~|~\CC=1,\RR=0]-|\SS_t(A_j)|\right\}\\
& =\left(1-e^{-\Delta t}\right)e^{-(|\SS_t|-1)\Delta t}\Bigg\{\sum_{j=1}^{p}|\SS_t(A_j)| \pi_t^*(A_j,A_0) \left[|\SS_{t}(A_0)|-1-|\SS_t(A_0)|\right]\\
&\quad\quad\quad\quad\quad\quad + \sum_{j=1}^{p}|\SS_t(A_j)| \left[1-\pi_t^*(A_j,A_0)\right] \left[|\SS_{t}(A_0)|-|\SS_t(A_0)|\right]\\
&\quad\quad\quad\quad\quad\quad + |\SS_t(A_0)| \pi_t^*(A_0,A_0)\left[|\SS_{t}(A_0)|-2-|\SS_t(A_0)|\right]\\
&\quad\quad\quad\quad\quad\quad + |\SS_t(A_0)| \left[1-\pi_t^*(A_0,A_0)\right]\left[|\SS_{t}(A_0)|-1-|\SS_t(A_0)|\right]\Bigg\}\\
& = \left(1-e^{-\Delta t}\right) e^{-(|\SS_t|-1)\Delta t}\Bigg\{-\sum_{j=1}^p|\SS_t(A_j)| \pi_t^*(A_j,A_0) -2|\SS_t(A_0)| \pi_t^*(A_0,A_0)\\
&\quad\quad\quad\quad\quad\quad -|\SS_t(A_0)|\left[1-\pi_t^*(A_0,A_0)\right]\Bigg\}.
\end{aligned}
\end{equation}

\paragraph{The term (d):} 
Same as Eq.~\eqref{eqn:k1r1}, we have
\begin{equation}
\label{eqn:k1r1a0_patienta0}
\begin{aligned}
\sum_{u\geq 1,v\geq 1}\P(\CC=u,\RR=v)\cdot\left\{\E[|\SS_{t+\Delta t}(A_0)|  ~|~ \CC=u,\RR=v]-|\SS_t(A_0)|\right\} =O( \Delta t^2 ).
\end{aligned}
\end{equation}
\paragraph{Putting terms (a)-(d) altogether:} 
Plugging Eqs.~\eqref{eqn:patient_00a0}-\eqref{eqn:k1r1a0_patienta0} into 
Eq.~\eqref{condition_probability}, and using the Taylor expansion, we obtain that,
\begin{equation*}
\begin{aligned}
 & \lim_{\Delta t\to 0+}\frac{\E[|\SS_{t+\Delta t}(A_0)|]-|\SS_t(A_0)|}{\Delta t}\\
& = (1-p\lambda) m-\sum_{j=1}^p|\SS_t(A_j)| \pi_t^*(A_j,A_0) -|\SS_t(A_0)| \pi_t^*(A_0,A_0) -|\SS_t(A_0)|.
\end{aligned}
\end{equation*}
This completes the proof for  $d|\SS_t(A_0)|/dt$.
\end{proof}

\subsection{Proof of Theorem \ref{main_result_patient}}
\label{sec:pfpatloss}
\begin{proof}
By definition of the Patient Algorithm,  the expected number of $A_0$-type agents becoming critical during $[t,t+dt]$ is $|\SS_t(A_0)|dt$. Among them, a proportion of $(1-\alpha)^{\sum_{k=0}^p|\SS_t(A_k)|}$ agents will perish from the market. 
Similarly, the expected number of $A_j$-type agents becoming critical  during $[t,t+d t]$ is $|\SS_t(A_j)|dt$. Among them, a proportion of $(1-\alpha)^{|\SS_t(A_0)|+|\SS_{t}(A_j)|}$ agents will perish from the market. 
The loss functions in Eqs.~\eqref{eqn:lossratio} and \eqref{eqn:lossforeachagent} can be equivalently written as,
\begin{equation}
\label{eqn:equivlosspatient}
\begin{aligned}
     L^{\text{patient}}(A_0) & =  \frac{\int_0^T|\SS_t(A_0)|(1-\alpha)^{\sum_{k=0}^p|\SS_t(A_k)|}dt}{(1-p\lambda)mT},\\
     L^{\text{patient}}(A_j) & =  \frac{\int_0^T|\SS_t(A_j)|(1-\alpha)^{|\SS_t(A_0)|+|\SS_t(A_j)|}dt}{\lambda mT}, \quad \forall j\geq 1,\\
    \text{and}\quad\quad   L^{\text{patient}} & = \frac{\int_0^T|\SS_t(A_0)|(1-\alpha)^{\sum_{k=0}^p|\SS_t(A_k)|}dt}{mT}\\
    &\quad + \frac{\sum_{j=1}^p\int_0^T|\SS_t(A_j)|(1-\alpha)^{|\SS_t(A_0)|+|\SS_t(A_j)|}dt}{mT}.
\end{aligned}
\end{equation}
The rest of the proof is completed in five parts: first, we establish that the ODEs in Theorem \ref{thm:Patientlimit} have a unique solution; second, we prove the solution has a symmetric structure and simplifies the ODE system; third,  we prove that there exists a stationary solution to the ODE system; 
fourth, we show that the ODE solution converges to a stationary solution using the Poincaré–Bendixson Theorem; finally, we analyze the loss of the Patient Algorithm.
\paragraph{Part 1: Existence and uniqueness of the solution.} The proof is the same as Part 1 in Section \ref{sec:pfthmgreedy}, and we omit the details.

\paragraph{Part 2: A symmetric structure of the solution.} We prove that when the initial values satisfy $|\SS_0(A_1)|=\cdots=|\SS_0(A_p)|\geq 0$, then 
\begin{equation}
\label{eqn:equivamongajpat}
|\SS_t(A_1)|=\cdots=|\SS_t(A_p)|,\quad\forall t>0.
\end{equation} 
The proof of Eq.~\eqref{eqn:equivamongajpat} is the same as Part 2 in Section \ref{sec:pfthmgreedy}, and we omit the details.
A byproduct of this result is that the ODEs in Theorem \ref{thm:Patientlimit} can be reduced into a group of two-dimensional ODEs, 
\begin{equation}
\label{2d-patient-ODE}
\begin{aligned}
\frac{d|\SS_t(A_0)|}{dt}  & =   (1-p\lambda)m  - |\SS_t(A_0)|\left[1-(1-\alpha)^{|\SS_t(A_0)|}\right](1-\alpha)^{p|\SS_t(A_1)|} \\
& \quad \quad -p|\SS_t(A_1)|\left[1-(1-\alpha)^{|\SS_t(A_0)|}\right](1-\alpha)^{|\SS_t(A_1)|} - |\SS_t(A_0)|,\\
\frac{d|\SS_t(A_1)|}{dt} &  = 
 \lambda m -\frac{1}{p}|\SS_t(A_0)|\left[1-(1-\alpha)^{p|\SS_t(A_1)|}\right]\\
& \quad \quad -|\SS_t(A_1)|\left[1-(1-\alpha)^{|\SS_t(A_1)|}\right]- |\SS_t(A_1)|,
\end{aligned}
\end{equation}
where we used Proposition \ref{thm:matchwithwho} and Eq.~\eqref{eqn:a0incompat}.

\paragraph{Part 3: Existence of a stationary solution.}  We show that the ODE system in Eq.~\eqref{2d-patient-ODE} has a stationary solution that doesn't depend on time. Suppose  $(|\SS(A_0)|,|\SS(A_1)|)$ is the stationary solution to Eq.~\eqref{2d-patient-ODE}. Then
\begin{equation}
\label{2d-stable-2}
\begin{aligned}
0  & =   (1-p\lambda)m  - |\SS(A_0)|\left[1-(1-\alpha)^{|\SS(A_0)|}\right](1-\alpha)^{p|\SS(A_1)|} \\
& \quad \quad -p|\SS(A_1)|\left[1-(1-\alpha)^{|\SS(A_0)|}\right](1-\alpha)^{|\SS(A_1)|} - |\SS(A_0)|,
\end{aligned}
\end{equation}
and 
\begin{equation}
\label{2d-stable-1}
\begin{aligned}
0 &  = 
 \lambda m -\frac{1}{p}|\SS(A_0)|\left[1-(1-\alpha)^{p|\SS(A_1)|}\right]\\
& \quad \quad -|\SS(A_1)|\left[1-(1-\alpha)^{|\SS(A_1)|}\right]- |\SS(A_1)|.
\end{aligned}
\end{equation}
By Eq.~\eqref{2d-stable-1}, we can derive
\begin{equation}
\label{rewritten_stable_1}
\begin{aligned}
|\SS(A_0)| = \frac{p\left\{\lambda m-|\SS(A_1)|\left[2-(1-\alpha)^{|\SS(A_1)|}\right]\right\}}{1-(1-\alpha)^{p|\SS(A_1)|}}\equiv f(|\SS(A_1)|).
\end{aligned}
\end{equation}
Plugging Eq.~\eqref{rewritten_stable_1} to Eq.~\eqref{2d-stable-2}, we have
\begin{equation}
\label{another_q}
\begin{aligned}
&|\SS(A_0)| = \frac{1}{\ln(1-\alpha)}\\
&\quad\cdot\ln\left\{1+\frac{\frac{p\{\lambda m - |\SS(A_1)|[2-(1-\alpha)^{|\SS_t(A_1)|}]\}}{1-(1-\alpha)^{p|\SS(A_1)|}}-(1-p\lambda)m}{n|\SS(A_1)|(1-\alpha)^{|\SS(A_1)|}+\frac{p\{\lambda m-|\SS(A_1)|[2-(1-\alpha)^{|\SS(A_1)|}]\}}{1-(1-\alpha)^{p|\SS(A_1)|}}(1-\alpha)^{p|\SS(A_1)|}}\right\}\\
&\quad\quad\quad\ \equiv g(|\SS(A_1)|).
\end{aligned}
\end{equation}
Let $s^*$ be the solution of 
\begin{equation*}
p\left\{\lambda m - \alpha^*\left[2-(1-\alpha)^{s^*}\right]\right\}=0,
\end{equation*} 
where the uniqueness of the solution $s^*$ is due to that the function 
\begin{equation*}
    p\left\{\lambda m-|\SS(A_1)|\left[2-(1-\alpha)^{|\SS(A_1)|}\right]\right\}
\end{equation*} is decreasing in $|\SS(A_1)|\in[0,\infty)$. Besides, $s^*>0$ holds because that 
\begin{equation*}
\lim_{|\SS(A_1)| \to 0} p\left\{\lambda m-|\SS(A_1)|\left[2-(1-\alpha)^{|\SS(A_1)|}\right]\right\} =p\lambda m >0.
\end{equation*}
Define \begin{align*}
\begin{split}
s^{**}& = \sup \Big\{ x \ \Big| \ \forall |\SS(A_1)| \in [0,x), g(|\SS(A_1)|)\text{ is well-defined, i.e.,} \\ & \ 1+\frac{\frac{p\{\lambda m - |\SS(A_1)|[2-(1-\alpha)^{|\SS(A_1)|}]\}}{1-(1-\alpha)^{p|\SS(A_1)|}}-(1-p\lambda)m}{n|\SS(A_1)|(1-\alpha)^{|\SS(A_1)|}+\frac{p\{\lambda m-|\SS(A_1)|[2-(1-\alpha)^{|\SS(A_1)|}]\}}{1-(1-\alpha)^{p|\SS(A_1)|}}(1-\alpha)^{p|\SS(A_1)|}}>0     \Big\}.
\end{split}
\end{align*}
Similarly, we can check $s^{**}>0$.
Let $\hat s= \min \{ s^{*}, s^{**} \}$.
Then we have that the functions $f$ and $q$ in Eqs.~\eqref{rewritten_stable_1}-\eqref{another_q}, respectively, are well-defined in $[0,\hat s)$. 
Moreover, note that, $\lim_{|\SS(A_1)| \to 0_+} f(|\SS(A_1)|)=+\infty$, $\lim_{|\SS(A_1)| \to 0_+} g(|\SS(A_1)|)=\frac{ \ln 2 }{ \ln (1-\alpha)} <0 $, and 
\begin{align*}
\lim_{|\SS(A_1)| \to \hat s} f(|\SS(A_1)|)=\begin{cases}  =0,  & \text{if} \quad \hat s=s^*<s^{**}, \\  < \infty & \text{if} \quad \hat s=s^{**},
\end{cases}
\end{align*}
and
\begin{align*}
\lim_{|\SS(A_1)| \to \hat s} g(|\SS(A_1)|)=  \begin{cases}  \frac{1}{ \ln(1-\alpha)}\ln \left(1-\frac{(1-p\lambda) m}{p s^* (1-\alpha)^{s^*} }\right)>0, & \text{if} \quad \hat s=s^*<s^{**}, \\   
+\infty & \text{if} \quad \hat s=s^{**}.
\end{cases}
\end{align*}
Therefore, the functions $f$ and $g$ have at least one intersection point in the interval $(0,\hat s)$, and in the intersection point, $|\SS(A_0)|=f(|\SS(A_1)|)>0$. 
Thus, Eqs.~\eqref{2d-stable-2} and \eqref{2d-stable-1} have at least one solution $(|\SS(A_0)|,|\SS(A_1)|)\in (0,\infty)\times (0,\infty)$, which is the stationary solution to  Eq.~\eqref{2d-patient-ODE}.

\paragraph{Part 4: Convergence to a stationary solution.}
We show that the solution to the ODEs in Eq.~\eqref{2d-patient-ODE} will converge to a stationary solution. The proof is the same as Part 4 in Section \ref{sec:pfthmgreedy}, and we omit the details.

\paragraph{Part 5: Property of the convergent stationary solution.}
We now analyze the properties of the stationary solution of Eq.~\eqref{2d-patient-ODE}. 
In the following, we consider three cases $\lambda p>1/2$, $\lambda p<1/2$, and $\lambda p=1/2$, separately. 

\textbf{Case 1.}  We start with the case that $\lambda p>1/2$. First, we show that,
\begin{equation}
\label{eqn:patzacc1}
    |\SS(A_1)|< \lambda m, \text{ and } |\SS(A_0)|< (1-p\lambda)m.
\end{equation}
To prove this result, we note that Eq.~\eqref{2d-stable-1} yields,
\begin{equation*}
\begin{aligned}
    \lambda m >|\SS(A_1)|\left[2-(1-\alpha)^{|\SS(A_1)|}\right] >|\SS(A_1)|.
\end{aligned}
\end{equation*}
Similarly, we note that Eq.~\eqref{2d-stable-2} gives,
\begin{equation*}
\begin{aligned}
    (1-p\lambda) m >|\SS(A_0)|\left[1+(1-\alpha)^{p|\SS(A_1)|} - (1-\alpha)^{|\SS(A_0)|+p|\SS(A_1)|}\right] >|\SS(A_0)|.
\end{aligned}
\end{equation*}
Hence Eq.~\eqref{eqn:patzacc1} holds.

We also show that when $m\to\infty,\alpha\to 0$, there exist constants $C_1,C_2>0$ such that,
\begin{equation}
\label{eqn:patzacc12}
    |\SS(A_1)|> C_1m, \text{ and } |\SS(A_0)|> C_2m.
\end{equation}
We use the proof by contradiction to prove Eq.~\eqref{eqn:patzacc12}. Suppose that there exist a sequence  $\{m_k,\alpha_k,\epsilon_k\}$ where $m_k\to\infty,\alpha_k\to 0, \epsilon_k\to 0$ as $k\to\infty$, and $|\SS(A_1)|\leq \epsilon_k m$ for any $k\geq 1$. Then by Eq.~\eqref{2d-stable-1}, 
\begin{equation*}
\begin{aligned}
    |\SS(A_0)| & = \frac{p\left\{\lambda m_k-|\SS(A_1)|\left[2-(1-\alpha_k)^{|\SS(A_1)|}\right]\right\}}{1-(1-\alpha_k)^{p|\SS(A_1)|}}\\
    & >p(\lambda-\epsilon_k)m_k,
\end{aligned}
\end{equation*}
which contradicts Eq.~\eqref{eqn:patzacc1} when $\lambda p>1/2$. Hence there exists a constant $C_1>0$ such that $|\SS(A_1)|> C_1m$.
Similarly, suppose that there exist a sequence  $\{m_k,\alpha_k,\epsilon_k\}$ where $m_k\to\infty,\alpha_k\to 0,\epsilon_k\to 0,$ as $k\to\infty$, and $|\SS(A_0)|\leq \epsilon_k m$ for any $k\geq 1$. Then by Eq.\eqref{2d-stable-2},
\begin{equation}
\label{control_Z_O}
\begin{aligned}
(1-p\lambda)m_k  & =      |\SS(A_0)|\left[1-(1-\alpha_k)^{|\SS(A_0)|}\right](1-\alpha_k)^{p|\SS(A_1)|} \\
& \quad \quad +p|\SS(A_1)|\left[1-(1-\alpha_k)^{|\SS(A_0)|}\right](1-\alpha_k)^{|\SS(A_1)|} + |\SS(A_0)|,\\
& \leq p|\SS(A_1)|(1-\alpha_k)^{|\SS(A_1)|}+\epsilon_km_k. 
\end{aligned}
\end{equation}
However, since $|\SS(A_1)|> C_1m_k$, 
we have $(1-\alpha_k)^{|\SS(A_1)|}<(1-\alpha_k)^{C_1 m_k}=e^{[-C_1+o(1)]d_k} \to 0$. Therefore, by Eq.~\eqref{control_Z_O}, we have  that as $k \to \infty$,
\begin{equation*}
    (1-p\lambda)\leq \frac{1}{m_k}p |\SS(A_1)| (1-\alpha_k)^{|\SS(A_1)|} +\epsilon_k < p \lambda (1-\alpha_k)^{|\SS(A_1)|}  +\epsilon_k\to 0.
\end{equation*}
Here in the second inequality, we used Eq.~\eqref{eqn:patzacc1}. This result contradicts the assumption that $(1-p\lambda)>0$ in Eq.~\eqref{eqn:occuragents}.
Hence Eq.~\eqref{eqn:patzacc12} holds.

By Eqs.~\eqref{eqn:patzacc1} and \eqref{eqn:patzacc12},  we let $|\SS(A_1)| =C_3 m$, $|\SS(A_0)|=C_4 m$, where $C_1<C_3<\lambda$ and $C_2<C_4<(1-p\lambda)$. Again, by Eqs.~\eqref{2d-stable-2} and \eqref{2d-stable-1}, we have,
\begin{equation}
\label{eqn:rewrten1pam}
\begin{aligned}
& (1-p\lambda) - C_4\left[1-(1-\alpha)^{C_4m}\right](1-\alpha)^{pC_3m} \\
&\quad\quad\quad\quad\quad\quad -pC_3\left[1-(1-\alpha)^{C_4m}\right](1-\alpha)^{C_3m} - C_4 =0,\\
&\lambda -\frac{1}{p}C_4\left[1-(1-\alpha)^{pC_3m}\right] -C_3\left[1-(1-\alpha)^{C_3m}\right]-C_3=0.
\end{aligned}
\end{equation}
Since when $m\to\infty,\alpha\to 0$, we have $(1-\alpha)^m = \left[(1-\alpha)^{1/\alpha}\right]^{d}= e^{[-1+o(1)]d}$. 
Due to $C_3>C_1>0$, we have $(1-\alpha)^{C_3m} = e^{-[C_3+o(C_3)]d}\leq e^{-[C_1+o(1)]d} = o(e^{-C_1d/2})$. Similarly, $(1-\alpha)^{C_4m} = o(e^{-C_2d/2})$. As a result, when $m\to\infty,\alpha\to 0$, Eq.~\eqref{eqn:rewrten1pam} can be written as:
\begin{equation*}
\begin{aligned}
&(1-p\lambda) - C_4\left[1-o\left(e^{-C_2d/2}\right)\right]o(e^{-pC_1d/2}) \\
&\quad\quad\quad\quad\quad\quad -pC_3\left[1-o\left(e^{-C_2d/2}\right)\right]o\left(e^{-C_1d/2}\right)-C_4=0,\\
&\lambda -\frac{1}{p}C_4\left[1-o\left(e^{-pC_1d/2}\right)\right]  -C_3\left[1-o\left(e^{-C_1d/2}\right)\right]-C_3=0.
\end{aligned}
\end{equation*}
Thus, $C_3=\lambda-\frac{1}{2p}+o(e^{-\min \{C_1,C_2\}d/2}), C_4=1-p\lambda+o(e^{-C_2d/2}) $. As a result, when  $m\to \infty$ and $\alpha\to 0$, there exist constants $c_1,c_2>0$ that only depend on
$\lambda$ and $p$ such that,
\begin{equation}
\label{eqn:pateintlpg}
\begin{aligned}
|\SS(A_0)| & = \left[1-p\lambda+o\left(e^{-c_1 d}\right)\right]m,\\
|\SS(A_1)| & = \left[\lambda-\frac{1}{2p}+o\left(e^{-c_2 d}\right)\right]m,
\end{aligned}\quad \text{if }\lambda p>\frac{1}{2}.
\end{equation} 

Based on Eq.~\eqref{eqn:pateintlpg}, we can analyze the loss of the Patient Algorithm when $\lambda p>1/2$. By definition of the loss functions in Eq.~\eqref{eqn:equivlosspatient}, then for any $j\geq 1$, when $T\to\infty$ and $d\to\infty$,
\begin{equation*}
\label{cal_big_A} 
\begin{aligned}
L^{\text{patient}}(A_j)&=\frac{|\SS(A_1)|}{\lambda m} (1-\alpha)^{ |\SS(A_1)|+ |\SS(A_0)|} \\ 
& = \frac{ \lambda -\frac{1}{2p}+o(e^{-c_2 d}) }{\lambda } (1-\alpha)^{ [ \lambda -\frac{1}{2p}+ (1-p\lambda)+o(e^{-c_1 d})+o (e^{- c_2 d})  ]m }\\ 
& = \left(1-\frac{1}{2p\lambda}\right)e^{-\left[\lambda -\frac{1}{2p}+1-p\lambda +o(1) \right]d   }\\
& = e^{-\left[1 -\frac{1}{2p}-(p-1)\lambda +o(1) \right]d   },
\end{aligned}
\end{equation*}
where we use the fact that $O(1)$ can be written as $e^{ o(1)d}$.
Moreover, consider $A_0$ when $T\to\infty$ and $d\to\infty$, then
\begin{equation*}
\label{cal_big_O} 
\begin{aligned}
L^{\text{patient}}(A_0)&=\frac{|\SS(A_0)|}{ (1-p\lambda) m} (1-\alpha)^{ p|\SS(A_1)|+ |\SS(A_0)|}
\\ 
& = \frac{ 1-p\lambda+o(e^{-c_2 d}) }{1-p\lambda } (1-\alpha)^{ \left[ p\lambda - \frac{1}{2}+ 1-p\lambda+o(e^{-c_1 d})+o (e^{- c_2 d})  \right]m }\\ 
&=\frac{ 1-p\lambda+o(e^{-c_2 d})  }{1-p\lambda }  e^{ [-\alpha+O(\alpha^2) ] \left[ \frac{1}{2}+o(e^{-c_1 d} )+o(e^{- c_2 d})  \right] m    }\\ 
&=e^ {\left[-\frac{1}{2} +o (1) \right]d }.
\end{aligned}
\end{equation*}

\textbf{Case 2.}
Next, we consider the case that $\lambda p<1/2$.
We first want to show when $m\to\infty,\alpha\to 0$, there exists a constant $C_5>0$ such that,
\begin{equation}
\label{eqn:patzacc1g}
    |\SS(A_1)|< C_5 \alpha^{-1}.
\end{equation}
We use the proof by contradiction to show Eq.~\eqref{eqn:patzacc1g}. Suppose that there exist a sequence  $\{m_k,\alpha_k,M_k\}$ where $m_k\to\infty,\alpha_k\to 0, M_k\to\infty,$ as $k\to\infty$, and $|\SS(A_1)|> M_k \alpha_k^{-1}$ for any $k\geq 1$. 
Then $(1-\alpha_k)^{|\SS(A_1)|}<(1-\alpha_k)^{M_k/\alpha_k} = e^{[-1+o(1)]M_k}\to 0$, which together with Eq.~\eqref{eqn:patzacc1} imply that
\begin{equation*}
\begin{aligned}
0 & < p|\SS(A_1)|(1-\alpha_k)^{|\SS(A_1)|}\left[1-(1-\alpha_k)^{|\SS(A_0)|}\right] \\
& < p \lambda m_k (1-\alpha_k)^{|\SS(A_1)|} =o(m_k).
\end{aligned}
\end{equation*}
Thus, by Eq.~\eqref{2d-stable-2}, we have,
\begin{equation}
\label{Z_O_approximate_eq}
\begin{aligned}
|\SS(A_0)|  & =  \frac{(1-p\lambda)m_k-p|\SS(A_1)|(1-\alpha_k)^{|\SS(A_1)|}[1-(1-\alpha_k)^{|\SS(A_0)|}]}{1+[1-(1-\alpha_k)^{|\SS(A_0)|}](1-\alpha_k)^{p|\SS(A_1)|}}\\
& = \frac{(1-p\lambda)m_k-o(m_k)}{1+o(1)} = [1-p\lambda +o(1)]m_k.
\end{aligned}
\end{equation}
Moreover, by Eq.~\eqref{2d-stable-1}, we have,
\begin{equation}
\label{Z_O_appro_upper}
\begin{aligned}
|\SS(A_0)|&= p\cdot\frac{ \lambda m_k -|\SS(A_1)|(1-\alpha_k)^{|\SS(A_1)|} - 2|\SS(A_1)|\left[1-(1-\alpha_k)^{|\SS(A_1)|} \right]}{1-(1-\alpha_k)^{p |\SS(A_1)|} }\\
&<p \cdot\frac{\lambda m_k }{1-o(1)}=[1+o(1)]p\lambda m_k.
\end{aligned}
\end{equation}
Combining Eqs.~\eqref{Z_O_approximate_eq} and \eqref{Z_O_appro_upper}, we have that
\begin{equation*}
    [1-p\lambda+o(1)]m_k <[1+o(1)]p\lambda m_k,
\end{equation*}
which implies $1-p\lambda \leq p\lambda$ and contradicts with $\lambda p <\frac{1}{2}$. Thus, we proved Eq.~\eqref{eqn:patzacc1g}.

We also show that when $m\to\infty,\alpha\to 0$, there exist a constant $C_6>0$ such that,
\begin{equation}
\label{eqn:patzacc1g2}
|\SS(A_1)|>C_6\alpha^{-1}.
\end{equation}
Otherwise, there exist a sequence  $\{m_k,\alpha_k,\epsilon_k\}$ where $m_k\to\infty,\alpha_k\to 0, \epsilon_k\to0,$ as $k\to\infty$, and $|\SS(A_1)|< \epsilon_k \alpha_k^{-1}$ for any $k\geq 1$. By Eq.~\eqref{2d-stable-1}, we have
\begin{equation}
\label{Z_O_infty}
\begin{aligned}
|\SS(A_0)|&=p \cdot \frac{ \lambda m_k- |\SS(A_1)|(1-\alpha_k)^{|\SS(A_1)|} -2 |\SS(A_1)|[1-(1-\alpha_k)^{|\SS(A_1)|} ] }{1-(1-\alpha_k)^{p |\SS(A_1)|} }\\
& > p\cdot \frac{ \lambda m_k -3 \epsilon_k \alpha^{-1}_k }{ 1- (1-\alpha_k)^ {p \epsilon_k \alpha^{-1}_k } }.
\end{aligned}
\end{equation}
Note that $1-(1-\alpha_k)^{p\epsilon_k \alpha_k^{-1} }=1-e^{[-1+o(1)](p\epsilon_k)}=p\epsilon_k+o(\epsilon_k)$. By Eq.~\eqref{Z_O_infty}, we have that
\begin{equation*}
\frac{|\SS(A_0)|}{m_k}\geq p\cdot \frac{ \lambda m_k -3 \epsilon_k \alpha^{-1}_k }{ p\epsilon_km_k}  = \lambda \epsilon_k^{-1} - 3\alpha_k^{-1}m_k^{-1}.
\end{equation*}
Let $\epsilon_k = \min\{k^{-1},\lambda\alpha_km_k/4\}$ for $k\geq 1$, then $|\SS(A_0)|/m_k\to\infty$. This contradicts with Eq.~\eqref{eqn:patzacc1} where $|\SS(A_0)|<(1-p\lambda)m_k$. Thus, we proved Eq.~\eqref{eqn:patzacc1g2}. Moreover, we also note that 
by Eq.~\eqref{2d-stable-1},
\begin{equation}
\label{eqn:lbdonssa0}
\begin{aligned}
|\SS(A_0)|& =p\cdot \frac{ \lambda m- |\SS(A_1)|(1-\alpha)^{|\SS(A_1)|} -2 |\SS(A_1)|[1-(1-\alpha)^{|\SS(A_1)|} ] }{1-(1-\alpha)^{p|\SS(A_1)|} } \geq p\lambda m.
\end{aligned}
\end{equation}

Next, by Eqs.~\eqref{eqn:patzacc1g} and \eqref{eqn:patzacc1g2}, we let 
$|\SS(A_1)|=C_7\alpha^{-1}$, and by Eqs. \eqref{eqn:patzacc1} and \eqref{eqn:lbdonssa0}, we let $|\SS(A_0)|=C_8m$, where the constant $C_6< C_7<C_5 $ and $p\lambda\leq C_8<1-p\lambda $.
Again, by Eqs.~\eqref{2d-stable-2} and \eqref{2d-stable-1}, we have,
\begin{equation}
\label{lambdansmall2}
\begin{aligned}
&(1-p\lambda)m- C_8 m \left[1-(1-\alpha)^{C_8 m} \right] (1-\alpha)^{p C_7 \alpha^{-1}}\\ 
&\quad\quad\quad -pC_7\alpha^{-1}(1-\alpha)^{C_7\alpha^{-1}}\left[1-(1-\alpha)^{C_8 m} \right] -C_8 m  =0,
\end{aligned}
\end{equation}
and
\begin{equation}
\label{lambdansmall1}
\lambda m -\frac{1}{p} C_8 m \left[1-(1-\alpha)^ {p C_7\alpha^{-1}}\right] - C_7\alpha^{-1}[1-(1-\alpha)^{C_7\alpha^{-1}}]-C_7\alpha^{-1}=0.
\end{equation}
By rewriting Eq.~\eqref{lambdansmall1}, we obtain that,
\begin{equation}
\label{solve_O_1}
C_8=\frac{p \left\{ \lambda -C_7\frac{1}{\alpha m}- C_7 \frac{1}{\alpha m}[1- (1-\alpha)^{ C_7\alpha^{-1}}  ]    \right\}    }{  1-(1-\alpha)^{pC_7\alpha^{-1} }   }=\frac{ p\left[\lambda+O(\frac{1}{\alpha m})\right]  }{1-(1-\alpha)^{pC_7\alpha^{-1}} }
\end{equation}
By rewriting Eq.~\eqref{lambdansmall2}, we obtain that, 
\begin{equation}
\label{solve_O_2}
\begin{aligned}
C_8 & =\frac{ (1-p\lambda) -pC_7\frac{1}{\alpha m}(1-\alpha)^{C_7 \alpha^{-1} }\left[1-(1-\alpha)^{C_8m}\right]}{1+ [1-(1-\alpha)^{C_8 m} ](1-\alpha)^{p C_7 \alpha^{-1}}  }\\
& =\frac{(1-p\lambda)+O(\frac{1}{\alpha m}) }{1+[1-o(e^{-{C_9}{\alpha m}}  )](1-\alpha)^{p C_7 \alpha^{-1}}  }
\end{aligned}
\end{equation}
Here $C_9$ is a positive constant derived as follows: we have  a constant $0<C_{10}<C_8$ such that $(1-\alpha)^{C_8m}<e^{-(C_{10}+o(1))\alpha m   }=o(e^{-C_{10}\alpha m/2})$. Let $C_9=C_{10}/2$, then we  have $(1-\alpha)^{C_8m}=o(e^{-C_9 \alpha m} ) $.
By Eqs.~\eqref{solve_O_1} and \eqref{solve_O_2}, we have that,
\begin{equation*}
\frac{ p\left[\lambda+O(\frac{1}{\alpha m})\right]  }{1-(1-\alpha)^{pC_7\alpha^{-1}} }=\frac{(1-p\lambda)+O(\frac{1}{\alpha m}) }{1+[1-o(e^{-C_9 \alpha m}  )](1-\alpha)^{p C_7 \alpha^{-1} }  }
\end{equation*}
Solving this equation gives,
\begin{equation}
\label{O_char}
(1-\alpha)^{p C_7\alpha^{-1}}=1-2p\lambda+O\left((\alpha m)^{-1}\right).
\end{equation}
On the other hand, we have:
\begin{equation*}
(1-\alpha)^{p C_7 \alpha^{-1} }
=e^{ p C_7 \alpha^{-1}  \ln (1-\alpha)  }  =e^{[-\alpha-\frac{1}{2}\alpha^2+O(\alpha^3)]p C_7 \alpha^{-1}  }=e^{-p C_7-\frac{1}{2} p C_7 \alpha+O(\alpha^2) }.
\end{equation*}
Plugging this into Eq.~\eqref{O_char}, we have that
\begin{equation*}
e^{-pC_7-\frac{1}{2}pC_7 \alpha+O(\alpha^2) }=1-2p\lambda+O\left((\alpha m)^{-1}\right).
\end{equation*}
Solving this equation gives,
\begin{equation*}
\begin{aligned}
C_7 & =\frac{\ln (\frac{1}{1-2p \lambda}) +O(\frac{1}{\alpha m})+O(\alpha^2)  }{p(1+\alpha/2)}\\
& =\frac{1}{p}\left[ 
\ln \left(\frac{1}{1-2p\lambda}\right)+O\left(\frac{1}{\alpha m}\right)+O(\alpha^2)\right]\left[1-\frac{\alpha}{2}+O(\alpha^2)\right].
\end{aligned}
\end{equation*}
As a result, 
\begin{equation*}
\begin{aligned}
|\SS(A_1)|& =C_7\alpha^{-1}=\frac{1}{p}\left[ 
\ln \left(\frac{1}{1-2p\lambda}\right)+O\left(\frac{1}{\alpha m}\right)+O(\alpha^2)\right] \left[ \alpha^{-1}-\frac{1}{2}+O(\alpha)  \right]\\
& =\frac{ \ln \left(\frac{1}{1-2p\lambda}\right)+O(\frac{1}{\alpha m}) }{p}\alpha^{-1}+\frac{\ln (1-2p\lambda)}{2p}+o(1).
\end{aligned}
\end{equation*}
Then we plug Eq.~\eqref{O_char} into Eq.~\eqref{solve_O_1}, we have
\begin{equation*}
C_8=\frac{ p\left[\lambda+O(\frac{1}{\alpha m})\right] }{2p\lambda + O(\frac{1}{\alpha m}) }=\frac{1}{2}+O\left(\frac{1}{\alpha m}\right).
\end{equation*}
Thus,
\begin{equation*}
|\SS(A_0)|=C_8m=\left[\frac{1}{2}+O\left(\frac{1}{\alpha m}\right) \right]m.
\end{equation*}
Therefore, when  $m\to \infty$ and $\alpha\to 0$, we have
\begin{equation}
\label{eqn:ssa01les}
\begin{aligned}
|\SS(A_0)| & = \left[\frac{1}{2}+O\left(\frac{1}{d}\right)\right]m,\\
|\SS(A_1)| & = \left[\ln\left(\frac{1}{1-2\lambda p}\right) + O\left(\frac{1}{d}\right)\right]\frac{m}{pd} + \frac{1}{2p}\ln(1-2\lambda p) + o(1),
\end{aligned}\quad \text{if }\lambda p<\frac{1}{2}.
\end{equation} 

Based on Eq.~\eqref{eqn:ssa01les}, we can analyze the loss of the Patient Algorithm when $\lambda p<1/2$. 
By definition of the loss functions in Eq.~\eqref{eqn:equivlosspatient}, then for any $j\geq 1$, when $T\to \infty$ and $d\to\infty$,
\begin{align*}
\begin{split}
L^{\text{patient}}(A_j)&=\frac{|\SS(A_1)|}{\lambda m} (1-\alpha)^{ |\SS(A_1)|+ |\SS(A_0)|}=O(d^{-1}) (1-\alpha)^{O(\alpha^{-1})} (1-\alpha)^{\frac{m}{2}} \\
&=O(d^{-1}) O(1) e^{ \frac{m}{2}\ln(1-\alpha)   } =O(d^{-1})e^{ ( -\frac{1}{2} + o(1) )d) }\\ 
& =e^{ \left[-\frac{1}{2}+o(1)\right] d},
\end{split}
\end{align*}
where the last equation is because that $O(d^{-1})$  can be written as $e^{ o(1)d}$.
Moreover, consider $A_0$ when $T\to\infty$ and $d\to\infty$, then
\begin{align*}
\label{cal_middle_O_rewritten}
\begin{split}
L^{\text{patient}}(A_0)&=\frac{|\SS(A_0)|}{ (1-p\lambda) m} (1-\alpha)^{ p|\SS(A_1)|+ |\SS(A_0)|} \\ 
&= \frac{1}{ 2(1-p \lambda)} (1-\alpha)^{ \left[\ln\left(\frac{1}{1-2p\lambda}\right) + o(1)   \right] \alpha^{-1} } (1-\alpha)^{\frac{m}{2}+ O(\alpha^{-1})  }
\\ & = \frac{\frac{1}{2}-p\lambda  }{1-p\lambda}
e^{\left[-\frac{1}{2}+o(1)\right]d }
\\ & =e^ {\left[-\frac{1}{2}+o(1)\right]d },
\end{split}
\end{align*}
where we use the fact that $O(1)$ can be written as $e^{ o(1)d}$.
In the third equation above, we used the following facts,
\begin{equation*}
\begin{aligned}
& (1-\alpha)^ { \left[\ln(\frac{1}{1-2p\lambda}) + o(1)\right]\alpha^{-1} }= e^{\left[\ln(\frac{1}{1-2p\lambda}) + o(1)\right]\alpha^{-1} \ln (1-\alpha)   } \\
&=e^{ \ln (1-2p\lambda) + o(1) }=1-2p\lambda+o(1),
\end{aligned}
\end{equation*}
and 
\begin{equation*}
\begin{aligned}
& (1-\alpha) ^{ \frac{m}{2}+O(\alpha^{-1}) }
 = e^ {\left[ \frac{m}{2}+ O(\alpha^{-1})  \right] \ln (1-\alpha)  }\\
& =e^  {\left[ \frac{m}{2}+ O(\alpha^{-1})  \right]\left[-\alpha + O(\alpha^2)\right]  }=e^{\left[-\frac{1}{2}+o(1)\right]d }.
\end{aligned}
\end{equation*}

\textbf{Case 3.}
Finally, we consider the case that $\lambda p=1/2$. 
We first want to show that when $m \to \infty, \alpha\to 0$,
\begin{equation}
\label{eqn:1alphassa10}
    (1-\alpha)^{|\SS(A_1)|} \to 0.
\end{equation}
Otherwise, there exists a sequence $\{m_k, \alpha_k \}$ and a constant $C_{12}>0$ where $m_k \to \infty$, $\alpha_k\to 0$, as $k\to\infty$, and $(1-\alpha_k)^{|\SS(A_1)|} >C_{12}$. By Eq.~\eqref{2d-stable-1}, we have
\begin{equation}
\label{eqn:1/2}
\begin{aligned}
|\SS(A_0)| & =p\cdot\frac{   \lambda m_k-|\SS(A_1)|\left[ 2- (1-\alpha_k)^{ |\SS(A_1)|} \right]     }{ 1- (1-\alpha_k)^{n |\SS(A_1)|} }=p\cdot\frac{[\lambda+o(1)]m_k }{1-(1-\alpha_k)^{n |\SS(A_1)|}  } \\
& >p[\lambda+o(1) ]m_k \frac{1}{1-C_{12}^n}=\left(\frac{1}{2}+o(1)\right)m_k \frac{1}{1-C_{12}^n}\geq \frac{1}{2}m_k.
\end{aligned}
\end{equation}
On the other hand, by Eq.~\eqref{2d-stable-2}, we have:
\begin{equation*}
\begin{aligned}
|\SS(A_0)|& = \frac{ (1-p\lambda) m_k  -p |\SS(A_1)| (1-\alpha_k)^{|\SS(A_1)|}\left[1-(1-\alpha_k)^{|\SS(A_0)|} \right]  }{ 1+   \left[ 1-(1-\alpha)^{|\SS(A_0)|} \right](1-\alpha)^{ p |\SS(A_1)|} }\\
& <(1-p\lambda)m_k =\frac{1}{2}m_k,
\end{aligned}
\end{equation*}
which contradicts with Eq.~\eqref{eqn:1/2}. Thus, we proved Eq.~\eqref{eqn:1alphassa10}. As a result, by Eq.~\eqref{2d-stable-2}, we have that,
\begin{equation}
\label{eqn:bdonssa01/2}
\begin{aligned}
|\SS(A_0)|& =\frac{ (1-p\lambda) m  -p |\SS(A_1)| (1-\alpha)^{|\SS(A_1)|}[1-(1-\alpha)^{|\SS(A_0)|} ]  }{ 1+   [ 1-(1-\alpha)^{|\SS(A_0)|} ](1-\alpha)^{ p |\SS(A_1)|} }\\
& =\frac{ (1-n\lambda+ o(1) )m}  {  1+ o(1)   }=\left(\frac{1}{2}+ o(1) \right)m.
\end{aligned}
\end{equation}
This gives the desired result for $|\SS(A_0)|$ when $m\to\infty,\alpha\to 0$.

We also show that when $m\to\infty,\alpha\to 0$,
\begin{equation}
\label{eqn:ssa1om1}
    |\SS(A_1)|/m\to 0. 
\end{equation}
Otherwise, there exist a sequence  $\{m_k,\alpha_k\}$ and a constant $C_{11}>0$ where $m_k\to\infty,\alpha_k\to 0,$ as $k\to\infty$, and $|\SS(A_1)|> C_{11}m_k$ for any $k\geq 1$. By Eq.~\eqref{2d-stable-1}, we have
\begin{equation}
\label{eqn:ss10eq1}
\begin{aligned}
|\SS(A_0)|& =p\cdot\frac{   \lambda m_k-|\SS(A_1)|\left[ 2- (1-\alpha_k)^{ |\SS(A_1)|} \right] }{ 1- (1-\alpha_k)^{p |\SS(A_1)|} }< p\cdot\frac{ (\lambda-C_{11}) m_k    }{1- (1-\alpha_k)^{ p |\SS(A_1)| } }\\
& \leq p\cdot\frac{  (\lambda -C_{11}) m_k   }{ 1-e^{ -(pC_{11}+o(1)) \alpha_km_k} }
= \left[ p\lambda - pC_{11} +o(1)  \right] m_k
\end{aligned}
\end{equation}
On the other hand, by Eq.~\eqref{2d-stable-2}, we have:
\begin{equation}
\label{eqn:ss10eq2}
\begin{aligned}
|\SS(A_0)| & =\frac{ (1-p\lambda) m_k  -p |\SS(A_1)| (1-\alpha_k)^{|\SS(A_1)|}\left[1-(1-\alpha_k)^{|\SS(A_0)|} \right]  }{ 1+   \left[ 1-(1-\alpha_k)^{|\SS(A_0)|} \right](1-\alpha_k)^{ p |\SS(A_1)|} }\\&= \frac{(1-p\lambda)m_k -p |\SS(A_1)| O(e^{ [-C_{11}+o(1)]\alpha_km_k }  )\left[1-(1-\alpha_k)^{|\SS(A_0)|} \right]}{1 + [1- (1-\alpha_k)^{|\SS(A_0)|}  ]O(e^{-[pC_{11}+o(1)]\alpha_km_k }) }\\ & = \frac{ [1-p\lambda + o (1)  ] m_k }{1+o (1)} \\ & 
=[ 1-p\lambda + o(1)] m_k.
\end{aligned}
\end{equation}
By Eqs.~\eqref{eqn:ss10eq1} and \eqref{eqn:ss10eq2}, we have that $ [ p\lambda - pC_{11} + o(1)]m_k =[1-p\lambda +o(1) ]m_k$, which implies that $p\lambda>1/2$, and contradicts with $p\lambda =1/2$. Thus we proved Eq.~\eqref{eqn:ssa1om1}. 

Next, we want to prove that when $m\to\infty,\alpha\to 0$, \begin{equation}
\label{key_lambdan0.5}
\frac{ -2p\left[1-(1-\alpha)^{|\SS(A_1)|} \right] |\SS(A_1)|  }{ \left[-2+O\left((1-\alpha)^{p |\SS(A_1)|}\right)\right] (1-\alpha)^{p |\SS(A_1)|} }=\left[\frac{1}{2}+O\left(\frac{|\SS(A_1)|}{m}(1-\alpha)^{|\SS(A_1)|} \right) \right]m.
\end{equation}
This result can be proved as follows.
By Eqs.~\eqref{2d-stable-2} and \eqref{2d-stable-1}, we obtain that
\begin{equation*}
\begin{aligned}
|\SS(A_0)| & =p\cdot\frac{   \lambda m-|\SS(A_1)|\left[ 2- (1-\alpha)^{ |\SS(A_1)|} \right]     }{ 1- (1-\alpha)^{p |\SS(A_1)|} }\\
& = \frac{ (1-p\lambda) m  -p |\SS(A_1)| (1-\alpha)^{|\SS(A_1)|}\left[1-(1-\alpha)^{|\SS(A_0)|} \right]  }{ 1+   [ 1-(1-\alpha)^{|\SS(A_0)|} ](1-\alpha)^{ p |\SS(A_1)|} }.
\end{aligned}
\end{equation*}
Thus, by Taylor expansion, we have
\begin{equation*}
\label{formula_Z_A}
\begin{aligned}
&  p\left\{\lambda m - |\SS(A_1)| \left[ 2 - (1-\alpha)^{|\SS(A_1)|}  \right]\right\}  \\  
= & \left\{(1-p\lambda)m-p|\SS(A_1)| (1-\alpha)^{|\SS(A_1)|}\left[1-(1-\alpha)^{|\SS(A_0)|} \right] \right\} \\
& \quad \cdot \frac{1-(1-\alpha)^{p |\SS(A_1)| }   }{1+ (1-\alpha)^{p |\SS(A_1)|} -(1-\alpha)^{|\SS(A_0)|+p|\SS(A_1)|} }  \\ 
& = \left\{(1-p\lambda)m-p|\SS(A_1)| (1-\alpha)^{|\SS(A_1)|}\left[1-(1-\alpha)^{|\SS(A_0)|} \right] \right\}\\
&\quad\cdot \left[ 1-2 (1-\alpha)^{ p |\SS(A_1)| } + O\left( (1-\alpha)^{2p |\SS(A_1)|}  \right) \right]+ O\left(e^{ \left(-\frac{1}{2}+o(1) \right) \alpha m }\right).
\end{aligned}
\end{equation*}
The above equation yields that,
\begin{equation}
\label{ori_key_lambdan0.5}  
\begin{aligned}
& \frac{ -2p |\SS(A_1)| + 2p |\SS(A_1)| (1-\alpha)^{|\SS(A_1)|}     }{  -2(1-\alpha)^{ p |\SS(A_1)|} + O\left((1-\alpha)^{ 2 p |\SS(A_1)| } \right)}\\
& =\frac{1}{2}m-p|\SS(A_1)| (1-\alpha)^{|\SS(A_1)|}
+\frac{ O \left(e^ { \left(-\frac{1}{2}+o(1) \right)\alpha m } \right)   }{ -2(1-\alpha)^{ p |\SS(A_1)|} + O\left((1-\alpha)^{ 2 p |\SS(A_1)| }\right)}.
\end{aligned}
\end{equation}
Note that  by Eq.~\eqref{eqn:ssa1om1}, $|\SS(A_1)|/m \to 0$,  then if $m\to \infty$, $\alpha \to 0$,  we have
\begin{equation*}
    (1-\alpha)^{p|\SS(A_1)|} \geq (1-\alpha)^{ p \epsilon m} =e^{(-p\epsilon + o(1))\alpha m}, \quad \forall \epsilon >0.
\end{equation*}
Thus, 
\begin{equation*}
    \frac{ O \left(e^ { \left(-\frac{1}{2}+o(1) \right)\alpha m }  \right)   }{ -2(1-\alpha)^{ p|\SS(A_1)|} + O\left((1-\alpha)^{2 p|\SS(A_1)|}\right) } = o\left(e^{- \left(\frac{1}{2}+\epsilon\right) \alpha m } \right), \quad \forall \epsilon > 0.
\end{equation*}
Again since $|\SS(A_1)|/m \to 0$, we have 
\begin{equation*}
e^{-\delta \alpha m} =o\left( (1-\alpha)^{|\SS(A_1)|}  \right), \quad \forall \delta >0.
\end{equation*}
By Eq.~\eqref{eqn:1alphassa10}, we have $(1-\alpha)^{|\SS(A_1)|} \to 0 $. Hence  as $m \to \infty$, $\alpha \to 0$,
\begin{equation*}
    |\SS(A_1)|>\alpha^{-1}.
\end{equation*}
Thus, as $m \to \infty, \alpha \to 0 $, there exists a constant $C_{13}>0$ such that
\begin{equation*}
    \frac{|\SS(A_1)|} {m} (1-\alpha)^{|\SS(A_1)|} > \frac{1}{\alpha m} e^{-\frac{1}{4} \alpha m}>C_{13}e^{- \left(\frac{1}{2}+\epsilon\right) \alpha m }.
\end{equation*}
As a result,  
\begin{equation}
\label{eqn:smossa1}
    \frac{ O \left(e^ { \left(-\frac{1}{2}+o(1) \right)\alpha m }  \right)   }{ -2(1-\alpha)^{ p|\SS(A_1)|} + O\left((1-\alpha)^{2 p|\SS(A_1)|}\right) } = o\left(\frac{|\SS(A_1)|}{m} (1-\alpha)^{|\SS(A_1)|} \right).
\end{equation}
Thus, by Eqs.~\eqref{ori_key_lambdan0.5} and \eqref{eqn:smossa1}, we complete the proof of Eq.~\eqref{key_lambdan0.5}.

Next, we want to prove that when $m\to\infty,\alpha\to 0$, then
\begin{equation}
\label{eqn:ssa1eq2}
    \frac{|\SS(A_1)|}{(1-\alpha)^{p|\SS(A_1)|}} = \left(\frac{1}{2p}+o(1)\right)m.
\end{equation}
To prove Eq.~\eqref{eqn:ssa1eq2}, we note that by Eqs.~\eqref{eqn:1alphassa10}, \eqref{eqn:ssa1om1} and \eqref{key_lambdan0.5}, then
\begin{align}
\label{sim_key_lambdan0.5}
\frac{ [-2p+o(1)] |\SS(A_1)|}{  [-2+o(1)](1-\alpha)^{p|\SS(A_1)|}  }=\left(\frac{1}{2}+o(1)\right)m.
\end{align}
Thus, we proved Eq.~\eqref{eqn:ssa1eq2}.

Finally, we can prove that when $m\to\infty,\alpha\to 0$, then
\begin{equation}
\label{eqn:ssa1eq3}
    (1-\alpha)^{p|\SS(A_1)| +|\SS(A_0)|  }
 =(1-\alpha)^{\frac{m}{2}}(1-\alpha)^{O\left( (1-\alpha)^{(p+1)|\SS(A_1)| } \right)  m}.
\end{equation}
Note that by Eq.~\eqref{2d-stable-2}, we have:
\begin{align*}
\begin{split}
|\SS(A_0)|&=\frac{ (1-p\lambda m) - p|\SS(A_1)|(1-\alpha)^{|\SS(A_1)|} [1-(1-\alpha)^{|\SS(A_0)|} ]  }{ 1+ [1-(1-\alpha)^{|\SS(A_0)|} ] (1-\alpha)^{ p |\SS(A_1)| } }\\
& =\frac{ \frac{1}{2} m - (1-\alpha)^{(p+1 ) |\SS(A_1)| }\left[1+o(1) \right]m  }{ 1+ (1-\alpha)^{p |\SS(A_1)| } \left[1+ O \left(e^{(-\frac{1}{2}+o(1))d } \right)\right] } \\ 
&= m \left[ \frac{1}{2}-O \left((1-\alpha)^{(n+1) |\SS(A_1)| }\right)  \right]\left[1-(1-\alpha)^{p |\SS(A_1)|}+ O \left((1-\alpha)^{2 p |\SS(A_1)|}\right)\right]
\\ &
=\frac{m}{2}-\frac{m}{2} (1-\alpha)^{p |\SS(A_1)|}+ O\left( (1-\alpha)^{(p+1)|\SS(A_1)|} \right).
\end{split}
\end{align*}
Then we have:
\begin{align}
\label{eqn:spfj1}
\begin{split}
(1-\alpha)^{p|\SS(A_1)| +|\SS(A_0)|  } =(1-\alpha)^{\frac{m}{2}+O\left((1-\alpha)^{(p+1)|\SS(A_1)| }\right) }(1-\alpha)^{-\frac{m}{2}(1-\alpha)^{p |\SS(A_1)|} +p|\SS(A_1)|} 
\end{split}
\end{align}
On the other hand, note that by Eq.~\eqref{key_lambdan0.5},  we have
\begin{align*}
\begin{split}
&p |\SS(A_1)|-p(1-\alpha)^{|\SS(A_1)|} |\SS(A_1)|\\
& =\frac{m}{2}(1-\alpha)^{p |\SS(A_1)|}+m \left\{     O\left((1-\alpha)^{(p+1)|\SS(A_1)|} \right)+O\left(\frac{|\SS(A_1)|}{m}(1-\alpha)^{|\SS(A_1)|}        \right)\right\},
\end{split}
\end{align*}
which implies that
\begin{align}
\label{eqn:skas1}
\begin{split}
& -\frac{m}{2}(1-\alpha)^{p |\SS(A_1)|}+p |\SS(A_1)|\\
& =m\left\{ O\left((1-\alpha)^{(p+1)|\SS(A_1)|}\right) +O\left(\frac{|\SS(A_1)|}{m}(1-\alpha)^{|\SS(A_1)|} \right) \right\}.
\end{split}
\end{align}
Moreover, by Eq.~\eqref{sim_key_lambdan0.5}, we have that  $|\SS(A_1)|/m=O\left((1-\alpha)^{p |\SS(A_1)|}\right)$, which implies that  
\begin{equation}
\label{eqn:skas2}
    \frac{|\SS(A_1)|}{m}(1-\alpha)^{|\SS(A_1)|}=O\left((1-\alpha)^{ (p+1)|\SS(A_1)|}\right).
\end{equation}
Combining Eqs.~\eqref{eqn:skas1} and \eqref{eqn:skas2}, we obtain that
\begin{equation}
\label{eqn:spfj2}
     -\frac{m}{2}(1-\alpha)^{p |\SS(A_1)|}+p |\SS(A_1)|=
m\left\{ O \left((1-\alpha)^{(p+1)|\SS(A_1)|}\right) \right\}.
\end{equation}
Therefore, combining  Eqs.~\eqref{eqn:spfj1} and \eqref{eqn:spfj2}, we completed the proof of Eqs.~\eqref{eqn:ssa1eq3}. 

Putting together Eqs.~\eqref{eqn:bdonssa01/2}, \eqref{eqn:ssa1eq2}, and \eqref{eqn:ssa1eq3}, we prove that when $m\to\infty,\alpha\to 0$, and $\lambda p=1/2$, then
\begin{equation}
\label{eqn:final1/2}
\begin{aligned}
|\SS(A_0)|& =\left[\frac{1}{2}+ o(1) \right]m,\\
|\SS(A_1)| & = \left[\frac{1}{2p}+o(1)\right](1-\alpha)^{p|\SS(A_1)|}m,\\
(1-\alpha)^{p|\SS(A_1)| +|\SS(A_0)|  }
 & =(1-\alpha)^{m/2}(1-\alpha)^{O\left( (1-\alpha)^{(p+1)|\SS(A_1)| } \right)  m}.
\end{aligned}
\end{equation}

Based on Eq.~\eqref{eqn:final1/2}, we can analyze the loss of the Patient Algorithm when $\lambda p=1/2$. By definition of the loss function in Eq.~\eqref{eqn:equivlosspatient}, we have that  when $T\to\infty$ and for any $j\geq 1$,
\begin{align}
\label{cal_middle_A}
\begin{split}
L^{\text{patient}}(A_j)&=\frac{|\SS(A_1)|}{\lambda m} (1-\alpha)^{ |\SS(A_1)|+ |\SS(A_0)|}
\\ & = \frac{ \left(\frac{1}{2p}+o(1)\right)(1-\alpha)^{p |\SS(A_1)|}m  }{\lambda m} (1-\alpha)^{ |\SS(A_1)|+ |\SS(A_0)|}\\
& =\frac{ \frac{1}{2p} + o (1) }{\lambda }(1-\alpha)^{ (p+1) |\SS(A_1)|+ |\SS(A_0)| }  \\ 
& = \left(\frac{1}{2p\lambda}+o(1)\right) (1-\alpha)^{\frac{m}{2}}(1-\alpha)^{|\SS(A_1)|} (1-\alpha)^{ O\left( (1-\alpha)^{(p+1)|\SS(A_1)|} m \right)}.
\end{split}
\end{align}
We want to show that
\begin{equation}
\label{eqn:bdona1ofd}
|\SS(A_1)| = O\left(\alpha^{-1}\ln(d)\right).
\end{equation}
This can be proved by contradiction. Suppose a sequence
$\{m_k,d_k,M_k\}$, where $m_k\to\infty, d_k\to\infty, M_k\to\infty,$ and $\alpha_k = d_k/m_k\to 0$, as $k\to\infty$,  and 
\begin{equation*}
|\SS(A_1)|>M_k\alpha_k^{-1}\ln(d_k).
\end{equation*}
Then
\begin{equation*}
\begin{aligned}
& \frac{|\SS(A_1)|}{(1-\alpha_k)^{p |\SS(A_1)|} }
>\frac{ M_k \alpha_k^{-1} \ln (d_k)}{  (1-\alpha_k)^{p M_k \alpha_k^{-1} \ln (d_k)} }\\
& = \frac{ M_k \alpha_k^{-1} \ln (d_k)  }{ e^{-(o(1)+pM_k \ln (d_k))} }=M_k \alpha_k^{-1}\ln (d_k) d_k^{p M_k}e^{o(1)}.
\end{aligned}
\end{equation*}
Thus,
\begin{equation*}
\frac{|\SS(A_1)|}{(1-\alpha_k)^{p |\SS(A_1)|} m_k}
=M_k \ln(d_k)d_k^{p M_k-1}e^{o(1)}\to \infty,
\end{equation*}
which contradicts Eq.~\eqref{eqn:final1/2}.
As a result, we proved Eq.~\eqref{eqn:bdona1ofd}. 
Combining Eqs.~\eqref{eqn:final1/2} and \eqref{eqn:bdona1ofd}, we obtain that, as $d\to\infty$
\begin{align*}
(1-\alpha)^{p |\SS(A_1)|}=(2p+o(1))\frac{|\SS(A_1)|}{m}=O\left(\frac{\ln d}{d}\right).
\end{align*}
Therefore, 
$(1-\alpha)^{|\SS(A_1)|}=O\left( \left(\frac{\ln (d)} {d}\right)^{\frac{1}{p}} \right),$ and $ (1-\alpha)^{(p+1)|\SS(A_1)|}=O\left( \left(\frac{\ln (d)} {d}\right)^{1+\frac{1}{p}} \right)$.
As a result, 
\begin{equation}
\label{n+1m}
\begin{aligned}
& (1-\alpha)^{ O\left( (1-\alpha)^{(p+1)|\SS(A_1)|} m \right)} = e^{O( (\frac{\ln (d) }{d})^{1+\frac{1}{p} })m \ln (1-\alpha) }\\
& = e^{ O(\alpha) O( (\frac{ \ln (d)}{d} )^{1+\frac{1}{p}} ) m   } = e^{ O (d^{-\frac{1}{p}}\ln (d)^{1+\frac{1}{p} }) } = e^{o(1)}=1+o(1).
\end{aligned}
\end{equation}
Therefore, when $d\to\infty$, $T\to\infty$, and $p\lambda=1/2$, Eq.~\eqref{cal_middle_A} can be written as,
\begin{align*}
\begin{split}
L^{\text{patient}}(A_j)&=\left(\frac{1}{2p\lambda}+o(1)\right) (1-\alpha)^{\frac{m}{2}+|\SS(A_1)|} (1-\alpha)^{ O\left( (1-\alpha)^{(p+1)|\SS(A_1)|} m \right)} \\ &
=\left(\frac{1}{2p\lambda}+o(1)\right) (1-\alpha)^{\frac{m}{2}}(1-\alpha)^{|\SS(A_1)|}(1+o(1))
\\& =\left(\frac{1}{2p\lambda} +o(1) \right)e^{ \ln (1-\alpha) \frac{m}{2} }(1-\alpha)^{|\SS(A_1)|}(1+o(1) )
\\ & = e^{\left[-\frac{1}{2}+o(1)\right]d}.
\end{split}
\end{align*}
In the last equation, we use that $(1-\alpha)^{|\SS(A_1)|}=e^{o(1)d}$, which can be proven as follows.
By Eq.~\eqref{eqn:bdona1ofd}, $|\SS(A_1)| = O\left(\alpha^{-1}\ln(d)\right)$. Hence,
$(1-\alpha)^{|\SS(A_1)|}=(1-\alpha)^{ O\left(\alpha^{-1}\ln(d)\right)} =e^{ o(1) d}$. 
Similarly, by definition of the loss function in Eq.~\eqref{eqn:equivlosspatient}, we have that  when $T\to\infty$ and $p\lambda=1/2$,
\begin{align}
\label{cal_middle_O}
\begin{split}
L^{\text{patient}}(A_0)&=\frac{|\SS(A_0|}{(1-p\lambda ) m }(1-\alpha)^{ p|\SS(A_1)| +|\SS(A_0)| }
= \left(1+o(1)\right)(1-\alpha)^{  p|\SS(A_1)| +|\SS(A_0)| } \\ 
&=\left(1+o(1)\right)(1-\alpha)^{\frac{m}{2} }(1+o(1) )=  (1+o(1))e^{ \frac{m}{2}\ln (1-\alpha) }  \\  &  = (1+ o(1) )e^{ \frac{m}{2} (-\alpha+ O(\alpha^2) ) } \\
& =e^{ \left[-\frac{1}{2}+ o(1)\right]d }.
\end{split}
\end{align}
The second and third equations are due to  Eqs.~\eqref{eqn:final1/2} and \eqref{n+1m}. Combining the results in the three cases, we finish the proof of $\lim_{T\to\infty}L^{\text{patient}}(A_k)$ for $k\geq 0$ in Theorem \ref{main_result_patient}. 
Together, by definition of the loss function in Eq.~\eqref{eqn:equivlosspatient}, we have that when $T\to\infty$,
 \begin{equation*}
    \lim_{T\to\infty}L^{\text{patient}} = 
        \begin{cases}
         e^{-\left[1-\frac{1}{2p}-(p-1)\lambda+o(1)\right]d}   & \text{if }\lambda p> \frac{1}{2},\\
         e^{\left[-\frac{1}{2}+o(1)\right]d}   & \text{if }\lambda p\leq \frac{1}{2}.
        \end{cases}
    \end{equation*}
This completes the proof of Theorem \ref{main_result_greedy}. 
\end{proof}

\subsection{Proof of Proposition \ref{prop:waittime}}
\begin{proof}
We consider the Greedy Algorithm and the Patient Algorithm separately. 
First, under the Greedy Algorithm, Eqs.~\eqref{eqn:bdonssa1alpha} and \eqref{eqn:bdonssa1alpha0} in Section \ref{sec:pfthmgreedy} show that there exists a constant $\tilde{c} = \min\{C_0,C_1\}$ such that,
\begin{equation*}
   |\SS(A_k)| \geq \tilde{c}\frac{m}{d}.  
\end{equation*}
Then by Little's Law \citep[e.g.,][]{leon2008probability}, the average waiting time of $A_0$-type agents and $A_j$-type agents ($j\geq 1$) in the market can be calculated as, respectively,
\begin{equation*}
    \frac{\tilde{c}}{1-p\lambda}\cdot\frac{1}{d}\quad \text{ and }\quad \frac{\tilde{c}}{\lambda}\cdot\frac{1}{d}.
\end{equation*}

Second, under the Patient Algorithm, Eq.~\eqref{eqn:pateintlpg} in Section \ref{sec:pfpatloss} shows that 
\begin{equation*}
\begin{aligned}
|\SS(A_0)| & = \left[1-p\lambda+o\left(e^{-c_1 d}\right)\right]m,\\
|\SS(A_1)| & = \left[\lambda-\frac{1}{2p}+o\left(e^{-c_2 d}\right)\right]m,
\end{aligned}\quad \text{if }\lambda p>\frac{1}{2}.
\end{equation*} 
By Little's Law, the average waiting time of $A_0$-type agents and $A_j$-type agents ($j\geq 1$) in the market can be calculated as, respectively,
\begin{equation*}
    \frac{1-p\lambda+o\left(e^{-c_1 d}\right)}{1-p\lambda}\quad \text{ and } \quad  \frac{\lambda-\frac{1}{2p}+o\left(e^{-c_2 d}\right)}{\lambda},\quad \text{if }\lambda p>\frac{1}{2}.
\end{equation*}
Similarly, Eq.~\eqref{eqn:ssa01les} in Section \ref{sec:pfpatloss} shows that 
\begin{equation*}
\begin{aligned}
|\SS(A_0)| & = \left[\frac{1}{2}+O\left(\frac{1}{d}\right)\right]m,\\
|\SS(A_1)| & = \left[\ln\left(\frac{1}{1-2\lambda p}\right) + O\left(\frac{1}{d}\right)\right]\frac{1}{p\alpha} + \frac{1}{2p}\ln(1-2\lambda p) + o(1),
\end{aligned}\quad \text{if }\lambda p<\frac{1}{2}.
\end{equation*} 
By Little's Law  \citep[e.g.,][]{leon2008probability}, the average waiting time of $A_0$-type agents and $A_j$-type agents ($j\geq 1$) in the market can be calculated as, respectively,
\begin{equation*}
    \frac{1/2+o(1)}{1-p\lambda}\quad \text{ and } \quad  \frac{\ln\left(\frac{1}{1-2\lambda p}\right) + o(1)}{\lambda p d},\quad \text{if }\lambda p<\frac{1}{2}.
\end{equation*}
Finally, Eqs.~\eqref{eqn:final1/2} and \eqref{eqn:bdona1ofd} in Section \ref{sec:pfpatloss} show that 
\begin{equation*}
\begin{aligned}
|\SS(A_0)|& =\left[\frac{1}{2}+ o(1) \right]m,\\
|\SS(A_1)| & = \left[\frac{1}{2p}+o(1)\right](1-\alpha)^{p \cdot O\left(\alpha^{-1}\ln(d)\right)}m,
\end{aligned}\quad \text{if }\lambda p=\frac{1}{2}.
\end{equation*}
By Little's Law, the average waiting time of $A_0$-type agents and $A_j$-type agents ($j\geq 1$) in the market can be calculated as, respectively,
\begin{equation*}
    \frac{1/2+o(1)}{1-p\lambda}\quad \text{ and } \quad  \Theta\left(\frac{1}{\lambda d}\right),\quad \text{if }\lambda p=\frac{1}{2}.
\end{equation*}
This completes the proof of Proposition \ref{prop:waittime}.
\end{proof}
\end{document}